\documentclass[lettersize,journal]{IEEEtran}

\usepackage{amsmath,amsfonts,amssymb}
\usepackage{algorithmic}
\usepackage{algorithm}
\usepackage{array}
\usepackage{textcomp}
\usepackage{stfloats}
\usepackage{url}
\usepackage{verbatim}
\usepackage{graphicx}
\usepackage{makecell}
\usepackage{ltablex} 
\usepackage[table]{xcolor}
\usepackage{cite}
\usepackage{enumitem}
\usepackage{longtable}
\usepackage{array}
\usepackage{multirow}
\usepackage{rotating}  
\usepackage{lscape}    
\usepackage{ltablex}         
\keepXColumns               

\newcolumntype{L}[1]{>{\raggedright\arraybackslash}p{#1}}
\newcolumntype{C}[1]{>{\centering\arraybackslash}p{#1}}
\newcolumntype{R}[1]{>{\raggedleft\arraybackslash}p{#1}}

\renewcommand\cellgape{\Gape[4pt]}
\usepackage{supertabular}
\usepackage{hyperref} 
\hypersetup{
    colorlinks=true,
    linkcolor=blue,
    citecolor=cyan,
}
\usepackage{bbding}
\usepackage{longtable}
\usepackage{booktabs}
\usepackage{stfloats}
\usepackage{color}
\usepackage[dvipsnames]{xcolor}
\usepackage{multirow}
\usepackage{caption}
\usepackage{subcaption}
\usepackage{pifont}
\usepackage{calc}  
\usepackage{adjustbox}
\usepackage{xspace}

\newcommand{\eg}{\textit{e.g.}\xspace}
\newcommand{\ie}{\textit{i.e.}\xspace}

\newcommand{\etc}{\textit{etc.}\xspace}

\newcommand{\etal}{\textit{et al.}\xspace}

\usepackage{relsize}
\usepackage{tabularx}

\usepackage{makecell}
\usepackage{cuted} 
\renewcommand\makecell[1]{\shortstack[l]{#1}}

\renewcommand\cellgape[1]{}
\setcellgapes{0pt}

\usepackage{tcolorbox}
\tcbuselibrary{skins,breakable}
\usepackage{xcolor}

\usepackage{wrapfig}

\definecolor{boxblue}{HTML}{254EDB}   
\definecolor{boxfill}{HTML}{E8F0FF}   
\definecolor{titlewhite}{HTML}{FFFFFF} 

\newtcolorbox{textbox}[1][]{%
  breakable,
  enhanced,
  colback=boxfill,
  colframe=boxblue,
  boxrule=0.8pt,
  arc=6pt,
  left=8pt,right=8pt,top=8pt,bottom=8pt,
  coltitle=titlewhite,
  fonttitle=\bfseries,
  colbacktitle=boxblue,
  title={#1}
}

\usepackage{chemfig}
\usepackage{mhchem}
\usepackage{tikz}
\usepackage{pgfplots}
\usetikzlibrary{shapes.geometric, arrows.meta, positioning, calc}

\usepackage{epigraph}

\usepackage{tikz}
\usepackage{eso-pic}
\usepackage{fancyhdr}
\usepackage{lastpage}

\newlength{\headerLogoTopMargin}      
\newlength{\headerLinePosition}       

\fancypagestyle{IEEEtitlepagestyle}{%
    \fancyhf{}
    
    \fancyfoot[C]{%
        \vspace*{-1em}%
        \begin{tikzpicture}[remember picture]
            \draw[line width=0.4pt] (-0.5\textwidth,0) -- (0.5\textwidth,0);
            \node[anchor=west] at (0.5\textwidth,0) {\hspace{0.5em}\thepage};
        \end{tikzpicture}%
    }
}

\AddToShipoutPicture*{
    \AtPageUpperLeft{%
        \begin{tikzpicture}[remember picture,overlay]
            \node[anchor=north west,inner sep=0pt] at (1cm,-\headerLogoTopMargin+5pt) {
                \includegraphics[width=120pt]{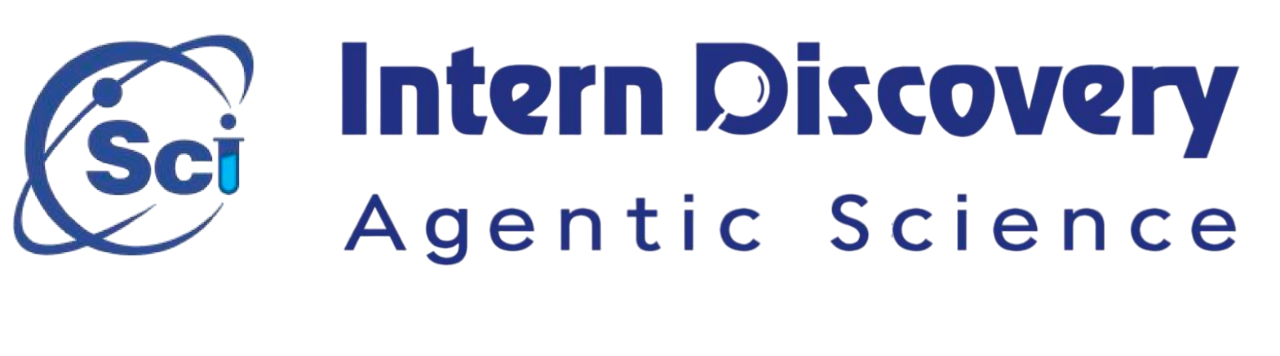}
            };
            
            \node[anchor=north east,inner sep=0pt] at (\paperwidth-1.1cm,-\headerLogoTopMargin+2pt) {
                \includegraphics[width=180pt]{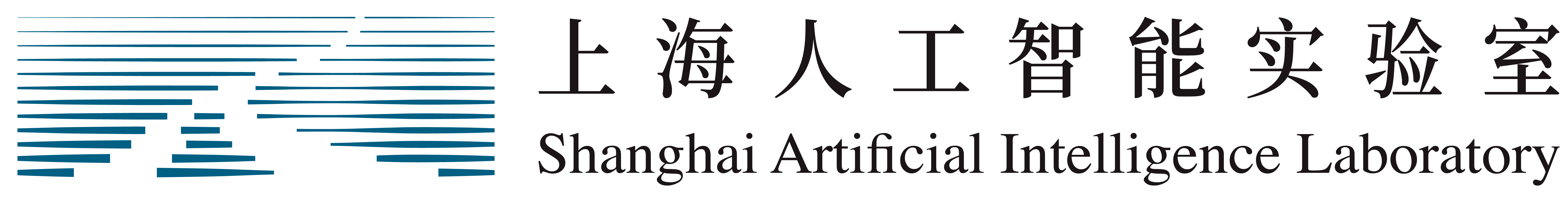}
            };
            
            \draw[line width=0.4pt] 
                (0.5cm,-\headerLinePosition) -- 
                (\paperwidth-0.5cm,-\headerLinePosition);
        \end{tikzpicture}%
    }%
}

\AddToShipoutPicture{
    \AtPageUpperLeft{%
        \begin{tikzpicture}[remember picture,overlay]
            \node[anchor=north west,inner sep=0pt] at (1cm,-\headerLogoTopMargin+5pt) {
                \includegraphics[width=120pt]{fig/logo-removebg-preview-Photoroom.png}
            };
            
            \node[anchor=north east,inner sep=0pt] at (\paperwidth-1.1cm,-\headerLogoTopMargin+2pt) {
                \includegraphics[width=180pt]{fig/logo_right.png}
            };
            
            \draw[line width=0.4pt] 
                (0.5cm,-\headerLinePosition) -- 
                (\paperwidth-0.5cm,-\headerLinePosition);
        \end{tikzpicture}%
    }%
}

\usepackage{etoolbox}
\makeatletter
\patchcmd{\section}{\normalfont}{\normalfont\bfseries}{}{}
\patchcmd{\subsection}{\normalfont}{\normalfont\bfseries}{}{}
\patchcmd{\subsubsection}{\normalfont}{\normalfont\bfseries}{}{}
\makeatother

\usepackage{fontawesome5}

\graphicspath{{fig/}}

\begin{document}




\title{A Survey of Scientific Large Language Models: From Data Foundations to Agent Frontiers}


\author{%
\begin{flushleft}
Ming Hu\textsuperscript{1,2}\thanks{$^\dagger$Corresponding Author, $^{\ddagger}$Project Leader, $^{\S}$Scientific Director}\quad
Chenglong Ma\textsuperscript{1,3}\quad
Wei Li\textsuperscript{1,4}\quad
Wanghan Xu\textsuperscript{1,4}\quad
Jiamin Wu\textsuperscript{1,5}\quad
Jucheng Hu\textsuperscript{1,6}\quad
Tianbin Li\textsuperscript{1}\quad
Guohang Zhuang\textsuperscript{1}\quad
Jiaqi Liu\textsuperscript{1,7}\quad
Yingzhou Lu\textsuperscript{8}\quad
% ...
\\[12pt]
\normalsize
$^1$Shanghai Artificial Intelligence Laboratory
$^2$Monash University
$^3$Fudan University\\
$^4$Shanghai Jiao Tong University
$^5$The Chinese University of Hong Kong \\
$^6$University College London
$^7$UNC-Chapel Hill
$^8$Stanford University
% ...


\medskip

\faGithub\ \textbf{Github Repository}:
\href{https://github.com/open-sciencelab/Awesome-Scientific-Datasets-and-LLMs}{https://github.com/open-sciencelab/Awesome-Scientific-Datasets-and-LLMs}

\end{flushleft}
  
}

\maketitle
\clearpage 

\onecolumn
\begin{abstract}
Scientific Large Language Models (Sci-LLMs) are transforming how knowledge is represented, integrated, and applied in scientific research, yet their progress is shaped by the complex nature of scientific data. This survey presents a comprehensive, data-centric synthesis that reframes the development of Sci-LLMs as a co-evolution between models and their underlying data substrate. We formulate a unified taxonomy of scientific data and a hierarchical model of scientific knowledge, emphasizing the multimodal, cross-scale, and domain-specific challenges that differentiate scientific corpora from general natural language processing datasets. We systematically review recent Sci-LLMs, from general-purpose foundations to specialized models across diverse scientific disciplines, alongside an extensive analysis of over 270 pre-/post-training datasets, showing why Sci-LLMs pose distinct demands—heterogeneous, multi-scale, uncertainty-laden corpora that require representations preserving domain invariance and enabling cross-modal reasoning. On evaluation, we examine over 190 benchmark datasets and trace a shift from static exams toward process- and discovery-oriented assessments with advanced evaluation protocols. These data-centric analyses highlight persistent issues in scientific data development and discuss emerging solutions involving semi-automated annotation pipelines and expert validation. Finally, we outline a paradigm shift toward closed-loop systems where autonomous agents based on Sci-LLMs actively experiment, validate, and contribute to a living, evolving knowledge base. Collectively, this work provides a roadmap for building trustworthy, continually evolving artificial intelligence (AI) systems that function as a true partner in accelerating scientific discovery. 
\end{abstract}



\begin{figure*}[th!]
    \centering
    \includegraphics[width=0.95\linewidth]{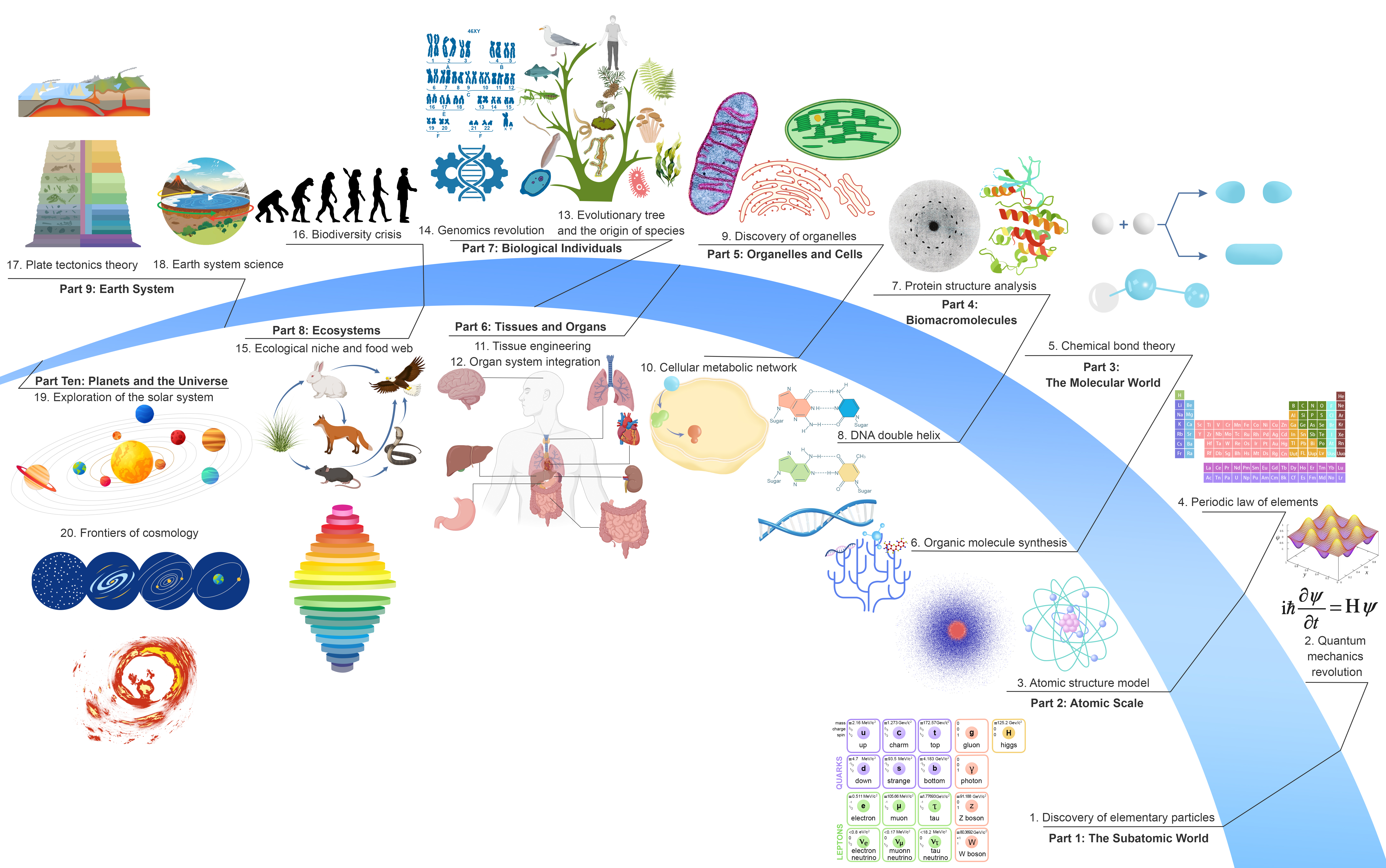}
    \caption{\textit{The song of humanity is a song of courage.} The diagram depicts the continuum of scientific inquiry spanning from subatomic particles through atomic and molecular structures, cellular and organismal biology, ecological systems, planetary sciences, to cosmological phenomena. Each tier represents distinct yet interconnected domains of investigation, illustrating the nested hierarchy of natural phenomena and the corresponding disciplinary frameworks employed in their study. This visualization encapsulates the expansion of scientific understanding from micro to macro dimensions, symbolizing humanity’s persistent pursuit of knowledge across all scales of nature.}
    \label{fig:4_period}
    \vspace{-5mm}
\end{figure*}

\twocolumn

\newpage
\onecolumn
\tableofcontents
\clearpage
\twocolumn
\newpage

\section{Introduction}
\label{sec:introduction}

\noindent\emph{``Science is built up with facts, as a house is with stones. But a collection of facts is no more a science than a heap of stones is a house.''}
\begin{flushright}
--- Henri Poincaré
\end{flushright}

\begin{figure*}[ht]
    \centering
    \includegraphics[width=0.48\linewidth]{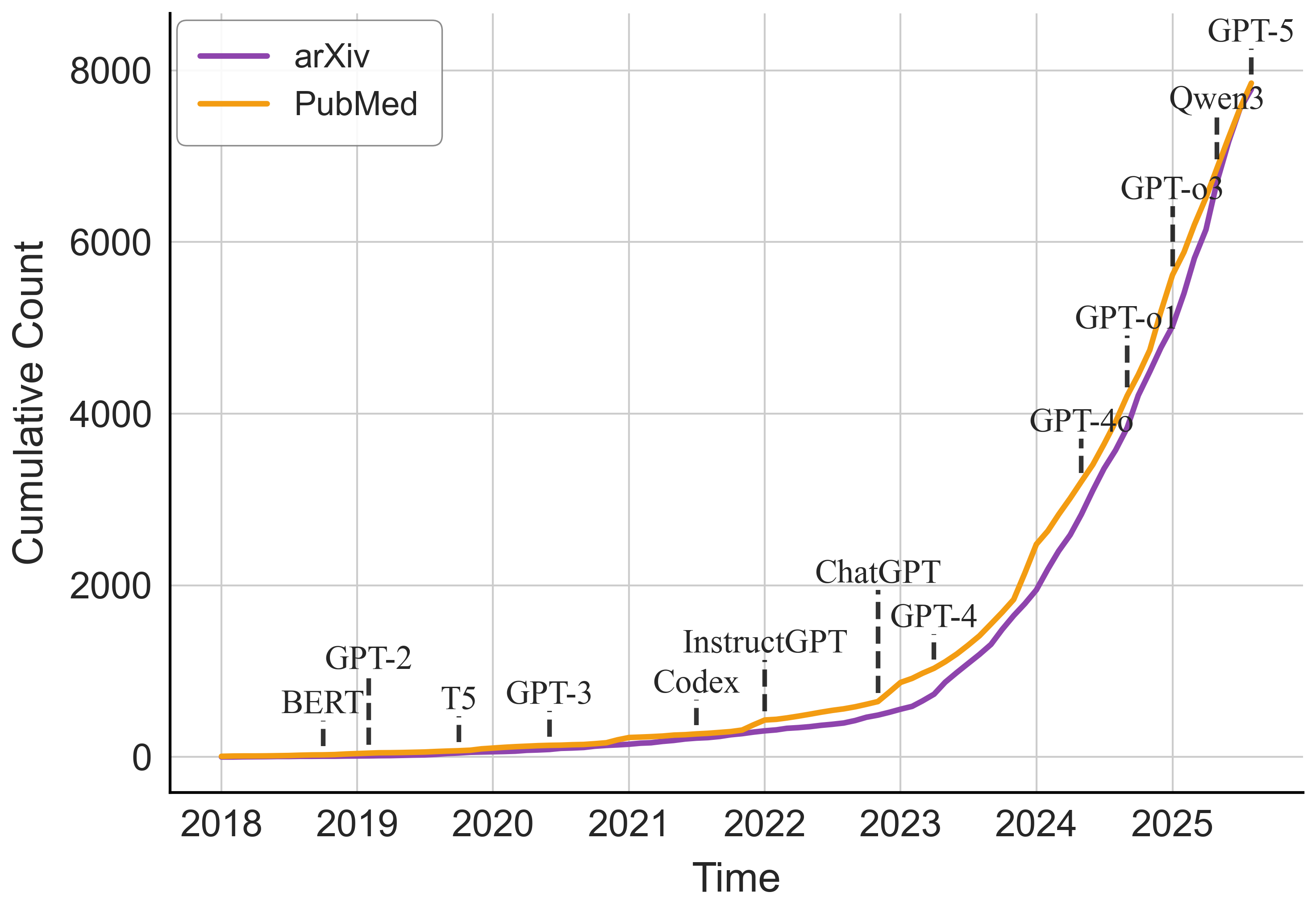}
    \hfill
    \includegraphics[width=0.48\linewidth]{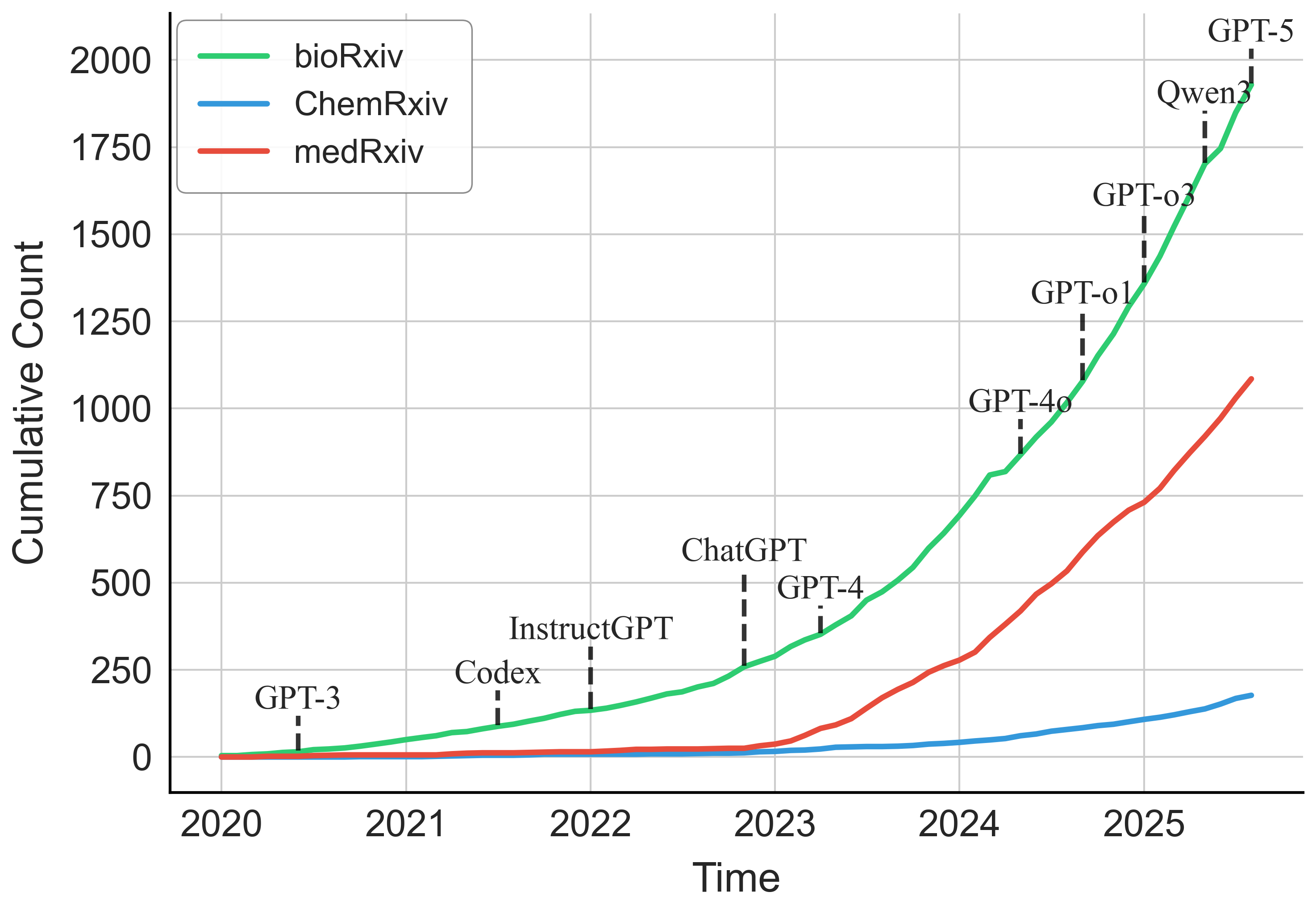}
    \caption{Cumulative trend of publications on major preprint platforms whose titles or abstracts mention the keyword ``language model'' or the combination ``language model + scientific domain'' (\eg, chemistry, physics, multi-omics, medicine, \etc).
    Left: Results from January 2018 to August 2025, from arXiv and PubMed. For arXiv, the matching includes ``language model'' in combination with additional science-related keywords; PubMed results are limited to occurrences in titles and abstracts. Both platforms show rapid growth.
    Right: Results from 2020 to August 2025, from bioRxiv, medRxiv, and ChemRxiv, all based on direct matches of ``language model'' in titles and abstracts. While the overall volumes are smaller than arXiv and PubMed, all three platforms, especially bioRxiv, show rapid acceleration, reflecting growing interdisciplinary interest in large language models across biomedical, chemical, and computational sciences.
}
    \label{fig:platform_comparison}
\end{figure*}

The rapid advancement of large language models (LLMs) has sparked a paradigm shift across numerous domains, demonstrating unprecedented transformative potential through task automation, productivity enhancement, and breakthrough innovations~\cite{wang2023scientific, jiang2024survey,zhang2025artificial,zhou2025survey,liu2024datasets} (Fig.~\ref{fig:platform_comparison}). These models have fundamentally transformed scientific research by introducing a unified approach that replaces traditional task-specific methods, extending beyond natural language processing to encompass diverse scientific data types, including molecules~\cite{liu2024drugllm}, proteins~\cite{xiao2025protein}, tables~\cite{fang2024large}, and complex metadata. LLMs have already revolutionized fields such as software engineering~\cite{ridnik2024code,jiang2024survey,hou2024large}, law~\cite{cui2023chatlaw,colombo2024saullm}, materials science~\cite{tom2024self,zimmermann202534}, healthcare~\cite{singhal2023large, GatorTronS, lu2024multimodal}, and biomedical research~\cite{huang2025biomni}, and have been applied across disciplines from mathematics~\cite{ahn2024large} and physics to chemistry~\cite{zhang2024chemllm}, biology~\cite{zhang2025scientific}, and geoscience~\cite{lin2023geogalactica}.

The evolution of scientific LLMs (Sci-LLMs) has undergone a paradigm shift through four distinct data-driven phases from 2018 to 2025 (Fig.~\ref{fig:4_period}). 
The initial \emph{transfer learning phase (2018--2020)} witnessed domain-specific adaptations of BERT~\cite{devlin2019bert} architecture, with models like SciBERT~\cite{beltagy2019scibert}, BioBERT~\cite{biobert}, and PubMedBERT~\cite{gu2021domain} trained on large-scale scientific corpora, showing that continued pre-training on domain literature yields sizable gains in downstream tasks that require scientific text understanding. These models provided reliable, static concept representations for specific downstream uses, but struggled to synthesize or generate novel scientific content at scale. 
The subsequent \emph{scaling phase (2020--2022)} embraced parameter and token-count expansion, marking a critical transition. Models like GPT-3~\cite{floridi2020gpt} with 175 billion parameters, along with later data/compute-optimal training rules~\cite{jordan2022chinchilla,kaplan2020scalinglaw} demonstrated that massive parameter scaling with diverse training data could achieve emergent knowledge integration capabilities, fundamentally altering the landscape of scientific AI. Galactica~\cite{taylor2022galactica} extended this lesson to science, with 120 billion parameters trained on more than 48 million scientific papers, textbooks, and encyclopedias, designing specialized tokenization schemes for mathematical formulas, chemical structures, and citations. MedPaLM-2~\cite{medpalm}, 
further instruction-tuned on multiple medical-domain datasets and achieved over 85\% accuracy on USMLE-style questions, becoming the first AI system to exhibit expert-level medical reasoning capabilities comparable to those of licensed physicians. However, scaling ran into a data wall for Sci-LLMs: unlike general-domain crawls with hundreds of billions to trillions of tokens, high-quality scientific text corpora were orders of magnitude smaller, with abundant scientific raw data underutilized in early large-scale attempts. 


The \emph{instruction-following phase (2022--2024)} shifted focus from capacity to alignment, introducing task adaptation via reinforcement learning from human feedback (RLHF). Examples include InstructGPT~\cite{ouyang2022training} and ChatGPT~\cite{openai2022chatgpt}, enabling more precise scientific task execution. Subsequently, foundational architectures represented by open-source LLMs (\eg, LLaMA~\cite{touvron2023llama}, Qwen~\cite{bai2023qwen}, ChatGLM~\cite{glm2024chatglm}, and Mistral~\cite{jiang2023mistral7b}) have enabled unprecedented diversity in scientific applications. Concurrently, the unprecedented expansion of instruction datasets has given rise to a series of milestone Sci-LLMs. Specifically, in the biomedical field, Meditron~\cite{chen2023meditron}, pre-trained on 48.1 billion tokens from medical literature, demonstrates the potential of open-source models in professional medical reasoning. ProteinChat~\cite{proteinchat}, trained on 1.5 million protein-prompt-answer triplets, facilitates protein research; LLaMA-Gene~\cite{llama-gene} integrates gigabytes of DNA, protein, and text data and 500 millions of instruction examples in DNA/protein tasks for training, achieving cross-modal biological sequence understanding. The multidisciplinary model SciGLM~\cite{zhang2024sciglm} leverages the efficient architecture of ChatGLM, fine-tuned on 254,000 carefully constructed instruction examples, achieving cross-disciplinary knowledge integration capabilities. Notably, several works demonstrate a strong correlation between data scale and model performance: HuatuoGPT-II~\cite{chen2023huatuogptii} utilizes an 11 TB medical corpus with million-scale documents for pre-training, 
while NatureLM~\cite{naturelm} is pre-trained on 143 billion tokens and fine-tuned using 45.1 million instruction-response pairs. This dual-drive paradigm of ``architectural diversity + data scaling'' has become the core framework for current scientific large language model development.

\begin{figure*}[h!]
    \centering
    \includegraphics[width=0.95\linewidth]{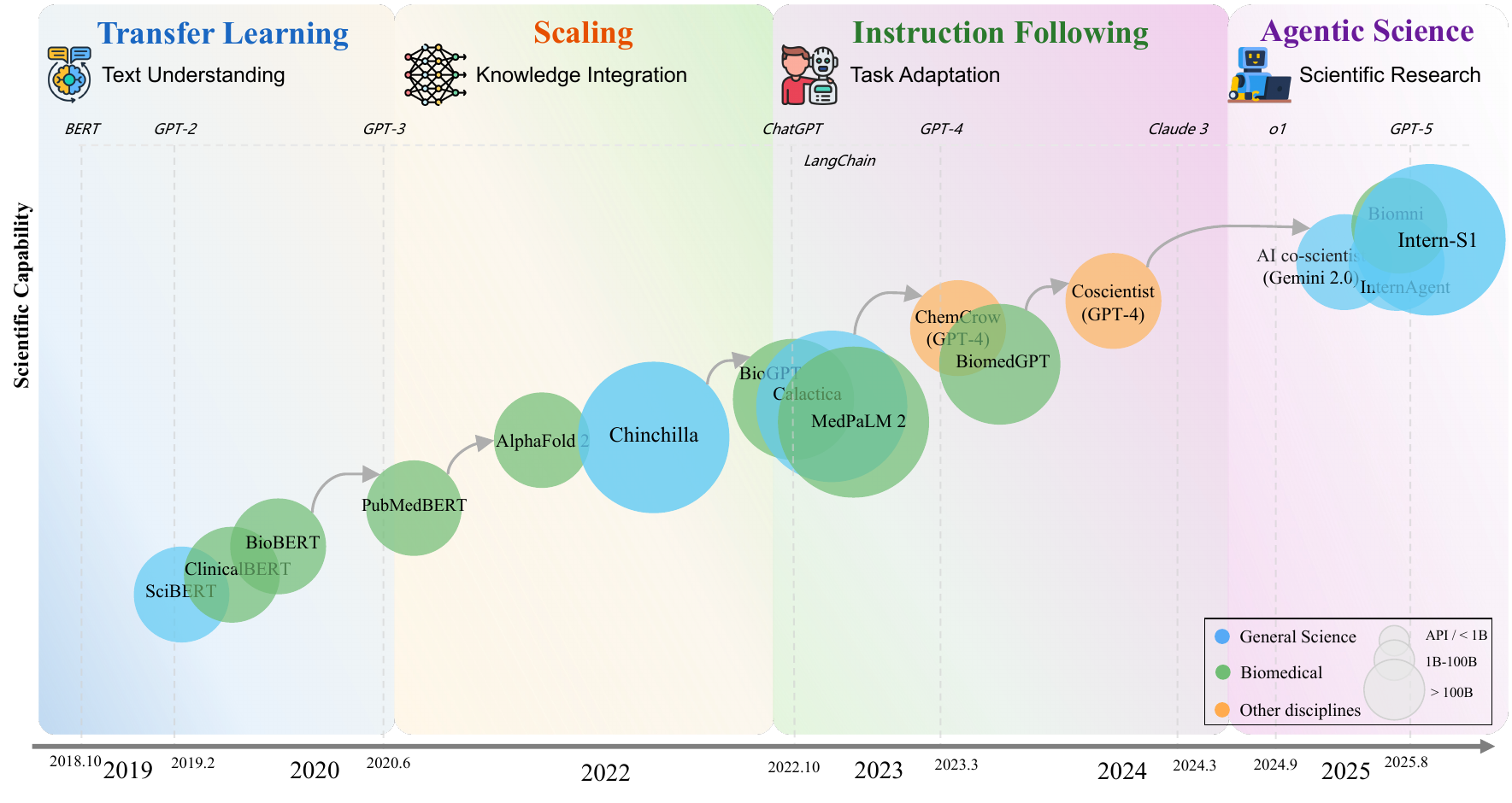}
    \caption{Evolution of Sci-LLMs reveals four paradigm shifts from 2018 to 2025, including (1) the progression from transfer learning approaches, (2) through the scaling era marked by knowledge integration in larger models, (3) instruction-following capabilities enabling flexible task adaptation, to (4) the latest paradigm introduces scientific agents—AI systems capable of autonomously conducting scientific research, from hypothesis generation and experimental design to data analysis and discovery. \textbf{Note:} Model positions reflect their release dates (x-axis) rather than strict paradigm classification. The four paradigms represent evolving trends in Sci-LLM development with overlaps and continuities, not mutually exclusive categories.}
    \label{fig:4_period}
    \vspace{-5mm}
\end{figure*}

Beyond excelling at analyzing existing scientific data, these models demonstrate remarkable potential in accelerating scientific discovery via hypothesis generation, theorem proving, experiment design, drug discovery, and weather forecasting, fundamentally reshaping how complex challenges are approached and solved in the era of AI-driven research~\cite{boiko2023autonomous,hong2023metagpt,gottweis2025towards}. As a prominent example of this trend, Intern-S1 \cite{bai2025interns1scientificmultimodalfoundation} is a scientific multimodal Mixture-of-Experts (MoE)~\cite{shazeer2017moe} foundation model with general understanding and reasoning capabilities alongside specialized expertise in scientific data analysis. Continually pre-trained on massive scientific data with 2.5 trillion tokens and enhanced with a Mixture-of-Rewards reinforcement learning, it surpasses existing closed-source state-of-the-art models in professional tasks such as molecular synthesis, reaction condition prediction, and crystalline thermodynamic stability prediction, while maintaining leading performance on general reasoning tasks. 

The latest paradigm of \emph{agentic science (2023--now)} is enabling AI systems with scientific agency, able to plan, act, and iterate across stages of discovery. Many works demonstrate end-to-end scientific workflows~\cite{boiko2023autonomous,yamada2025ai}, with increasing focus on multi-agent~\cite{ghafarollahi2025sciagents,ghareeb2025robin} and tool ecosystems~\cite{bran2023chemcrow,huang2025biomni}. Multi-agent designs emulate laboratory hierarchies from principal investigators to domain specialists, coordinating through formalized meeting protocols and critique–iteration loops~\cite{luo2025largeagent,swanson2025virtual}. Such systems generate scientific ideas with improved novelty and feasibility by explicitly modeling research teamwork~\cite{su2024many} and scientific law constraints~\cite{pu2025piflow}. At scale, cooperative frameworks manage entire research lifecycles (problem scoping, manuscript drafting, \etc), preserving persistent artifacts and audit trails~\cite{schmidgall2025agent}, while embodied variants integrate robotic execution with adaptive planning~\cite{song2025multiagent}. Parallel advances in tool integration center on knowledge-graph–driven orchestration~\cite{ding2025scitoolagentknowledgegraphdrivenscientific} and domain-scale agents interfacing with hundreds of software tools, databases, and instruments with provenance tracking~\cite{huang2025biomni}.

Despite these promising results, Sci-LLMs encounter fundamental challenges stemming from the \emph{unique characteristics of scientific data and knowledge representation}. Unlike the relatively homogeneous text corpora for general-purpose LLM development, scientific datasets exhibit extreme heterogeneity across modalities and formats. For instance, in chemistry alone, models must reconcile molecular strings, 3D molecular coordinates, spectroscopic data, and reaction mechanisms, each requiring distinct processing strategies~\cite{zhang2024comprehensive}. This heterogeneity extends beyond chemistry to encompass the full spectrum of scientific disciplines. In life sciences, models must simultaneously process genomic sequences, protein structures, multi-omics data, and clinical imaging~\cite{hasin2017multi,chen2021data,ruan2025large}, while astronomical applications demand integration of time-series photometry, spectroscopic observations, and multi-wavelength imaging across vastly different spatial and temporal scales~\cite{York2000_SDSS,GWTC3}. 

The challenge is further compounded by the hierarchical nature of scientific knowledge itself, which spans from raw observational data to abstract theoretical frameworks, each with its own representational requirements~\cite{rozzi2013linking,simon2012architecture}. Moreover, scientific data often embodies domain-specific semantics that resist straightforward tokenization or embedding. Mathematical equations carry precise symbolic relationships that must be preserved during processing~\cite{Dan2025_SymbolicNumerical,Wang2025_DrSR}, while crystallographic information files encode 3D structural constraints essential for materials science applications~\cite{jain2013commentary,dunn2020benchmarking}. Time-series data from instruments like Laser Interferometer Gravitational-Wave Observatory (LIGO) contain subtle signals buried in noise, requiring specialized preprocessing for physical interpretability~\cite{GWTC3,Evans2008_LHC}. These diverse data types cannot be adequately represented through conventional text-based approaches, necessitating novel architectures that preserve domain-specific invariance while enabling cross-modal reasoning~\cite{multimodal1,multimodal2,li2025mmscidatasetgraduatelevelmultidiscipline}. The integration of such heterogeneous data sources poses additional computational and methodological challenges. Cross-scale modeling, from quantum mechanical calculations to macroscopic phenomena, demands architectures capable of capturing multi-resolution dependencies~\cite{horstemeyer2009multiscale}. Furthermore, the uncertainty in experimental measurements require models to propagate error bounds and maintain scientific rigor throughout the reasoning process~\cite{Heisenberg1927,brennecke2013accounting,sivia2006data}. These constraints fundamentally distinguish scientific AI from general-purpose language modeling, requiring specialized solutions that respect the unique epistemological foundations of scientific inquiry.

\begin{figure}[t!]
    \centering
    \includegraphics[width=0.95\linewidth]{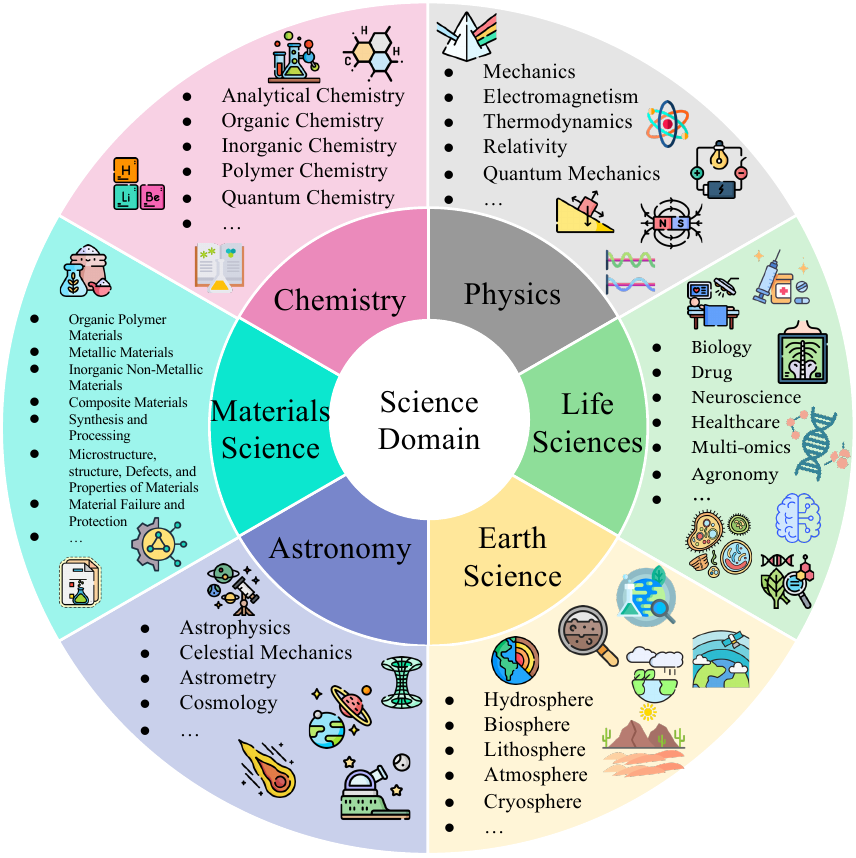}
    \caption{Six main scientific domains covered in this survey. The figure illustrates the primary disciplines investigated in our study on science-oriented large language models, encompassing Chemistry, Materials Science, Physics, Life Sciences, Astronomy, and Earth Science, along with representative subfields within each domain.}
    \label{fig:scientific_domains}
    \vspace{-5mm}
\end{figure}

The inherent complexity of scientific data and reasoning naturally extends to the evaluation of Sci-LLMs, where conventional natural language processing benchmarks prove insufficient for capturing domain-specific competencies. Recent efforts have produced comprehensive evaluation suites such as ScienceQA~\cite{lu2022learn}, which tests multimodal scientific understanding across elementary to graduate levels, and MMLU-Pro~\cite{wang2024mmlu}, which includes rigorous assessments in specialized fields like quantum physics and molecular biology. However, these benchmarks often fail to capture the nuanced requirements of scientific discovery, \eg, the ability to generate novel hypotheses, identify non-obvious connections between disparate findings, or design experiments that test theoretical predictions. To address this gap, Liu~\etal propose ResearchBench~\cite{liu2025researchbench}, a large-scale scientific discovery benchmark spanning 12 disciplines to systemically evaluate the hypothesis generation capabilities of LLMs.
Furthermore, researchers have also begun developing process-oriented evaluations that assess intermediate reasoning steps rather than just final answers, exemplified by frameworks like ScienceAgentBench~\cite{chen2024scienceagentbench} that evaluate models on complex scientific workflows, including literature review, experimental design, and result interpretation. Benchmarks such as MultiAgentBench~\cite{zhu2025multiagentbench} and WorkflowBench~\cite{fan2024workflowllm} now quantify collaboration, coordination, and workflow synthesis skills, marking a shift toward measurable, safety-aware, and reproducible science automation. The community has also recognized that scientific validity requires more than linguistic fluency; models must respect fundamental constraints such as physical laws, chemical valence rules, and biological feasibility~\cite{zhang2025scientific,liu2025chemauharnessreasoningllms,kesseli2025logicpybridginggapllms}. This has led to the integration of symbolic reasoning modules and constraint satisfaction systems that act as guardrails during generation, ensuring that model outputs remain within scientifically plausible bounds while still allowing for creative exploration at the frontiers of knowledge.


To address these gaps, several survey papers look into adjacent facets of the problem. A few works~\cite{he2025survey,mousa2025comparative} focused on models and tasks for biomedical data; Zhang~\etal~\cite{zhang2025scientific} examined Sci-LLMs under a broader perspective that involves both biological and chemical domains. Other works~\cite{zhang2024comprehensive} explored the application of Sci-LLMs in scientific discovery. Wei~\etal~\cite{wei2025from} and Wang~\etal~\cite{wang2025hitchhiker} reviewed scientific agent paradigms and system designs for autonomous research and scientific discovery. Ni~\etal~\cite{ni2025surveylargelanguagemodel} conducted a survey on existing benchmarks for LLMs involving several science fields. Chen~\etal~\cite{chen2025ai4research} provided a comprehensive survey on AI for autonomous scientific research, offering a systematic taxonomy and compiling resources across multiple disciplines. However, these reviews are theme-specific and limited to models with only a cursory touch on the underlying substrate---scientific datasets, throughout pre-training, post-training and evaluation. 
Complementing these perspectives, our survey contributes a unified, cross-disciplinary synthesis that explicitly \emph{links data foundations to agent frontiers}. 
We summarize the contributions as follows:
\begin{itemize}

    \item By introducing a unified taxonomy of scientific data and a hierarchical model of scientific knowledge, we provide a novel epistemological framework for analyzing the challenges in representing scientific information, from raw observational data and symbolic notations to abstract theoretical insights.
    
    \item We deliver a comprehensive and structured account of the rapidly evolving landscape of scientific large language models across six main scientific domains (\ie, physics, chemistry, life sciences, Earth Science, astronomy, and materials science; as in Fig.~\ref{fig:scientific_domains}). 
    
    \item By systematically analyzing over 270 pre- and post-training datasets, we provide a comprehensive panorama of current scientific datasets for Sci-LLM development, distilling the multimodal, cross-scale, and domain-specific challenges that distinguish Sci-LLMs from their general-purpose counterpart. 
    
    \item We conduct a comprehensive review of over 190 evaluation datasets for Sci-LLMs, discussing the shift of evaluation from static exams to research-level scientific discovery, the increasing employment and combination of domain-specific metrics, and the emergence of advanced evaluation methodologies. 


    \item We identify structural failures in scientific data curation and translate them into a forward-looking data development agenda that supports advanced scientific intelligence, advocating for a closed-loop feedback between autonomous scientific discovery and scientific data infrastructure. 
\end{itemize}
Collectively, these contributions establish a consolidated reference and a clear roadmap for building trustworthy, continually evolving Sci-LLMs capable of accelerating data-driven scientific discovery.

The paper is organized as follows: Sec.~\ref{sec:background} formulates a unified taxonomy of scientific data grounded in a hierarchical model of scientific knowledge. Sec.~\ref{sec:scillms} shows the landscape of Sci-LLMs across six main scientific domains. Secs~\ref{sec:pre-training_data}, \ref{sec:post-training_data}, and \ref{sec:evaluation_data} provide an extensive catalog and analysis of existing pre-training, post-training, and evaluation datasets for Sci-LLMs. Sec.~\ref{sec:analysis} analyzes how scientific data shapes LLM development and identify systemic issues that impede AI-readable corpora. Sec.~\ref{sec:future} outlines forward directions for scientific discovery empowered by advanced scientific agents and data ecosystems. Secs.~\ref{sec:outlook} and \ref{sec:conclusion} summarize challenges, outlook, and conclusion distilled from the paper. 

\section{Background}
\label{sec:background}

This section provides the foundations for understanding scientific AI systems. We first examine the diverse taxonomy of scientific data across disciplines (Sec.~\ref{sec:background_taxonomy}), followed by an analysis of the hierarchical structure of scientific knowledge (Sec.~\ref{sec:background_structure}), which reveals that scientific understanding forms a sophisticated multilevel system rather than a simple information repository. Then, we identify critical challenges unique to scientific AI (Sec.~\ref{sec:background_challenges}), including knowledge consistency, interpretability, and the integration of cross-scale multimodal data. We conclude by establishing frameworks for evaluating both data quality standards (Sec.~\ref{sec:background_quality}) and AI system capabilities specific to scientific domains (Sec.~\ref{sec:background_eval}). These elements collectively define the requirements for AI systems designed to support rigorous scientific discovery and reasoning.

\subsection{Taxonomy of Scientific Data}
\label{sec:background_taxonomy}

Scientific data manifests in striking diversity across disciplines, shaped by the fundamental questions and methodological paradigms unique to each field. In this subsection, we review and summarize the primary data types and modalities across scientific domains, examining how they appear and function within different scientific contexts, including: 
\textit{textual formats} (papers, experimental reports) in Sec.~\ref{sec:background_taxonomy_textual}, \textit{visual data} (medical scans, astronomical observations) in Sec.~\ref{sec:background_taxonomy_visual}, \textit{symbolic representations} (formulas, chemical structures) in Sec.~\ref{sec:background_taxonomy_symbolic}, \textit{structured data} (databases, knowledge graphs) in Sec.~\ref{sec:background_taxonomy_structured}, and \textit{time-series data} (neurophysiological recordings, astronomical light curves) in Sec.~\ref{sec:background_taxonomy_time}. In addition to these general types, we also discuss \textit{multi-omics integration} in Sec.~\ref{sec:background_taxonomy_multiomics} as a special case, as it represents an emerging paradigm that requires combining heterogeneous data across multiple biological layers (\eg, genomics, transcriptomics, proteomics). 
This taxonomy sets the stage for understanding how scientific data collectively support AI-driven scientific discovery across domains, and also establishes the foundation for developing multimodal large language models (MLLMs) which aim to process and integrate heterogeneous scientific data within a unified framework. 

\subsubsection{Textual Formats}
\label{sec:background_taxonomy_textual}

Scientific textual data forms the foundational substrate for knowledge representation across disciplines, encompassing a rich hierarchy from primary experimental documentation to synthesized knowledge repositories. At the most granular level, laboratory notebooks, experimental protocols, and field observations capture the raw process of scientific discovery, documenting not only successful experiments but also failed attempts and methodological refinements that prove invaluable for reproducibility and knowledge transfer~\cite{schnell2015ten}. This primary documentation feeds into specialized databases and repositories that have become central to modern scientific practice: genomic sequences in GenBank~\cite{benson2012genbank}, protein structures in RCSB~\cite{burley2021rcsb}, chemical compounds in PubChem~\cite{wang2017pubchem,kim2023pubchem}, and astronomical observations in NASA's Astrophysics Data System~(ADS)~\cite{kurtz2000nasa}, collectively housing petabytes of structured information linked to their textual descriptions and metadata.

The scholarly communication layer builds upon this foundation through peer-reviewed journals, comprehensive textbooks, and increasingly, preprint repositories that accelerate knowledge dissemination. Traditional venues like \textit{Physical Review Letters}, \textit{The Astrophysical Journal}, and \textit{Monthly Notices of the Royal Astronomical Society} maintain rigorous standards while platforms such as arXiv~\cite{arxiv} and ChemRxiv~\cite{kiessling2016chemrxiv} enable rapid sharing of emerging findings across physics, astronomy, chemistry, and interdisciplinary domains. This academic corpus is complemented by educational resources ranging from open-access textbooks like OpenStax series~\cite{OpenStax_CP2e,OpenStax_Physics} and The Feynman Lectures~\cite{FeynmanLectures} to specialized training materials including agricultural extension question-answering (QA) records~\cite{kpodo2024agxqa}, examination questions, and curated datasets for AI model evaluation such as ScholarChemQA~\cite{chen2025unveiling}, ScienceQA~\cite{saikh2022scienceqa}, and materials science benchmarks~\cite{auer2023sciqa,zaki2023mascqaquestionansweringdataset,cui2025curie}.

Beyond traditional academic outputs, scientific textual data increasingly encompasses regulatory documentation, real-time observational streams, and computational artifacts that reflect the evolving nature of modern research. Clinical trial registries~\cite{zarin2011clinicaltrials}, institutional review protocols~\cite{resnik2018ethics}, and biosafety guidelines~\cite{baltimore2015prudent} ensure responsible research conduct, while electronic health records~\cite{jensen2012mining,shickel2017deep}, citizen science annotations from projects like Galaxy Zoo~\cite{galaxy_zoo}, and real-time environmental monitoring data~\cite{kelling2009data} bridge laboratory findings with societal applications. The integration of computational approaches has spawned new textual categories, including bioinformatics pipelines~\cite{thiele2010protocol}, systems biology models~\cite{palsson2015systems}, synthesis planning frameworks~\cite{wu2025chematagentenhancingllmschemistry}, and code generation benchmarks~\cite{tang2024biocoder,tian2024scicode}, all requiring extensive documentation for reproducibility. This diverse textual ecosystem not only archives scientific progress but enables meta-analyses~\cite{gurevitch2018meta}, knowledge synthesis efforts, and increasingly sophisticated AI-driven discovery across the full spectrum of scientific inquiry.

\begin{figure*}[t!]
    \centering
    \includegraphics[width=0.95\linewidth]{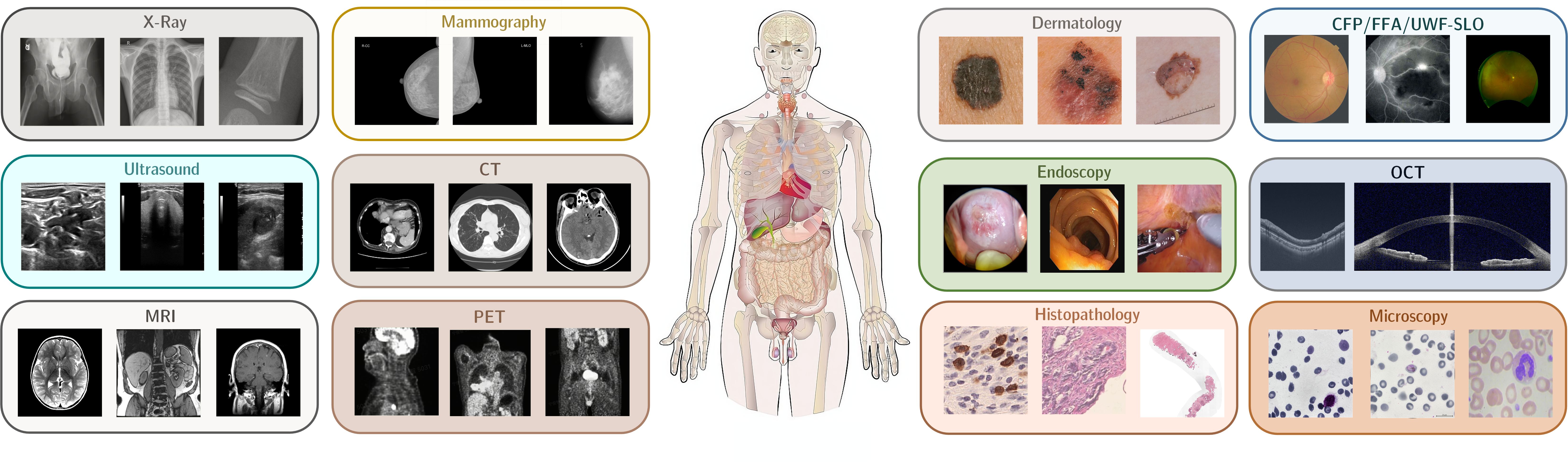}
    \caption{Examples of visual data across typical medical imaging modalities, involving radiology (PET, CT, mammography, X-ray, MRI, and ultrasound), dermatology, ophthalmology (CFP, FFA, UWF-SLO, and OCT), endoscopy, histopathology, and cellular microscopy. The figure is sourced from open-source medical datasets.}
    \label{fig:medical_modality}
    \vspace{-3mm}
\end{figure*}


\begin{figure}[t!]
    \centering
    \includegraphics[width=0.95\linewidth]{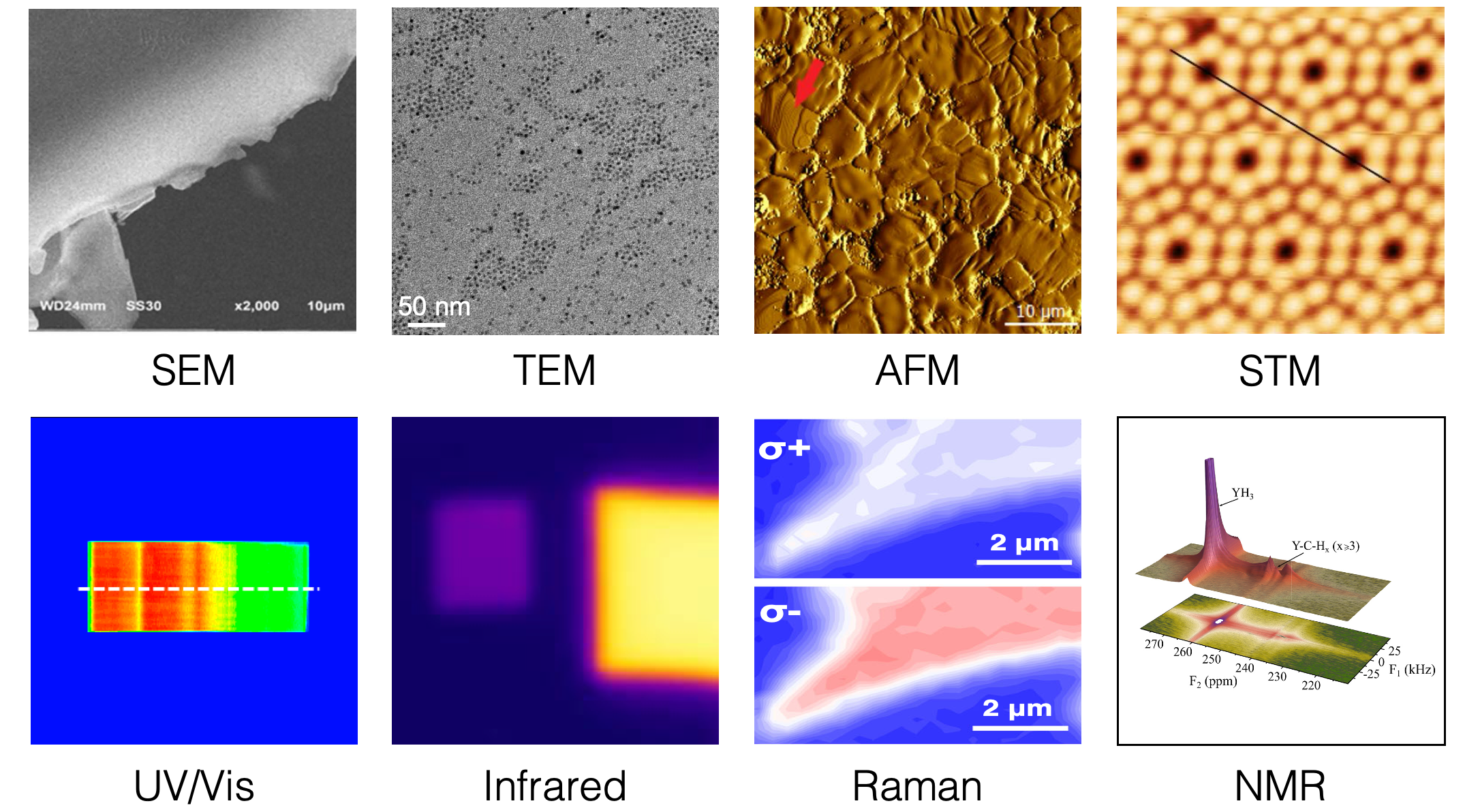}
    \caption{Examples of visual data in physics. SEM of epoxy with/without AlN~\cite{ali2023mineral}; TEM of W-doped Cu--Pt nanoalloys~\cite{liu2022light}; AFM topography of hyper-stoichiometric UO$_2$~\cite{tobi2025applicability}; STM of Si (111)-(7$\times$7) at multiple scan sizes~\cite{mansurov2022stm};
    UV/Vis contour map (500--680\,nm)~\cite{woo2023real}; Infrared thermographs of a directional emitter~\cite{fan2024directional}; Raman helicity-resolved maps of 1T-TaS$_2$~\cite{zhao2023spectroscopic}; NMR of yttrium hydrides~\cite{meier2021situ}. All panels are reused or adapted under the stated licenses (CC-BY-4.0 or CC-BY), with minor cropping only.}
    \label{fig:physics_visual_data}
    \vspace{-6mm}
\end{figure}

\subsubsection{Visual Data}
\label{sec:background_taxonomy_visual}

Visual data in scientific domains broadly fall into two categories: instrumental imaging that directly captures physical subjects through various sensing technologies, and diagrammatic representations that abstract and visualize concepts, relationships, and analytical results. These visual data span an extraordinary range of scales and modalities, from sub-atomic particle interactions to cosmic structures, providing essential foundations for multimodal AI systems to understand scientific phenomena. 


At the smallest scales, as shown in Fig.~\ref{fig:physics_visual_data}, advanced microscopy techniques, including scanning and transmission electron microscopy (SEM/TEM)~\cite{Goldstein2018_SEM,Williams2009_TEM}, atomic force microscopy (AFM)~\cite{Binnig1986_AFM}, and scanning tunneling microscopy (STM)~\cite{Binnig1982_STM}, reveal atomic structures and molecular arrangements critical for physics, materials science and chemistry. 
Visual spectrum data, including ultraviolet-visible spectrophotometry (UV/Vis)~\cite{Skoog2017_UVVis}, infrared~\cite{Stuart2004_IR}, Raman~\cite{Ferraro2003_Raman}, and nuclear magnetic resonance (NMR)~\cite{Claridge2016_NMR} spectroscopy, serve as molecular ``fingerprints'' across chemistry, materials science, and physics, with visual representations proven effective for spectrum learning~\cite{lu2024deep,liu2024deep}. 

In life sciences, light microscopy (brightfield, confocal) and fluorescence microscopy capture cellular structures and protein localizations, with datasets like the Human Protein Atlas~\cite{ponten2008hpa} and Broad Bioimage Benchmark Collection~\cite{ljosa2012bbbc} supporting cell segmentation and phenotype classification tasks. These microscopy images, typically stored in formats like TIFF~\cite{tool-tiff} or ND2~\cite{tool-nd2}, have been increasingly leveraged for training visual-language models~\cite{lozano2024microbench,burgess2025microvqa}. Moving up in scale, whole-slide digital pathology produces gigapixel images stored in SVS format, essential for cancer diagnosis, with large cohorts like TCGA~\cite{hutter2018tcga} and CPTAC~\cite{edwards2015cptac} providing thousands of images paired with diagnostic reports~\cite{he2020pathvqa,ikezogwo2023quilt,chen2025slidechat}.

\begin{figure*}[t!]
    \centering
    \includegraphics[width=0.95\linewidth]{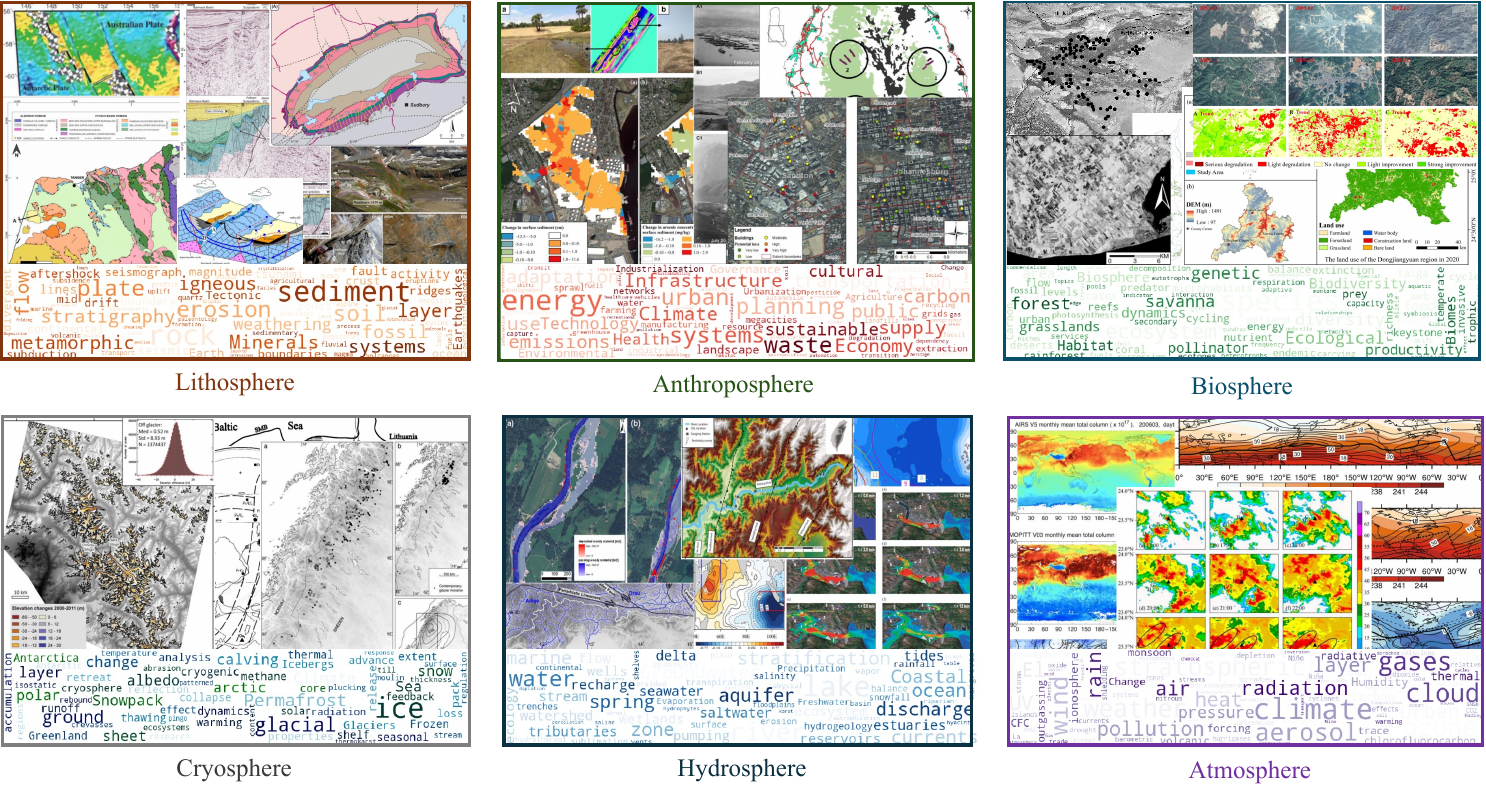}
    \caption{Data from Earth science's six major domains, including the lithosphere, anthroposphere, biosphere, cryosphere, hydrosphere, and atmosphere. Each panel consists of geospatial data, maps, satellite imagery, charts, \etc These data sources are highly diverse, encompassing a wide range of spatial and temporal resolutions, as detailed in Sec.~\ref{sec:background_structure_fact}. The figure is sourced from MSEarth~\cite{zhao2025msearth}, and authorization for its use has been obtained from the original author.}
    \label{fig:earth_vis_data}
    \vspace{-3mm}
\end{figure*}

\begin{figure}[t!]
    \centering
    \includegraphics[width=0.9\linewidth]{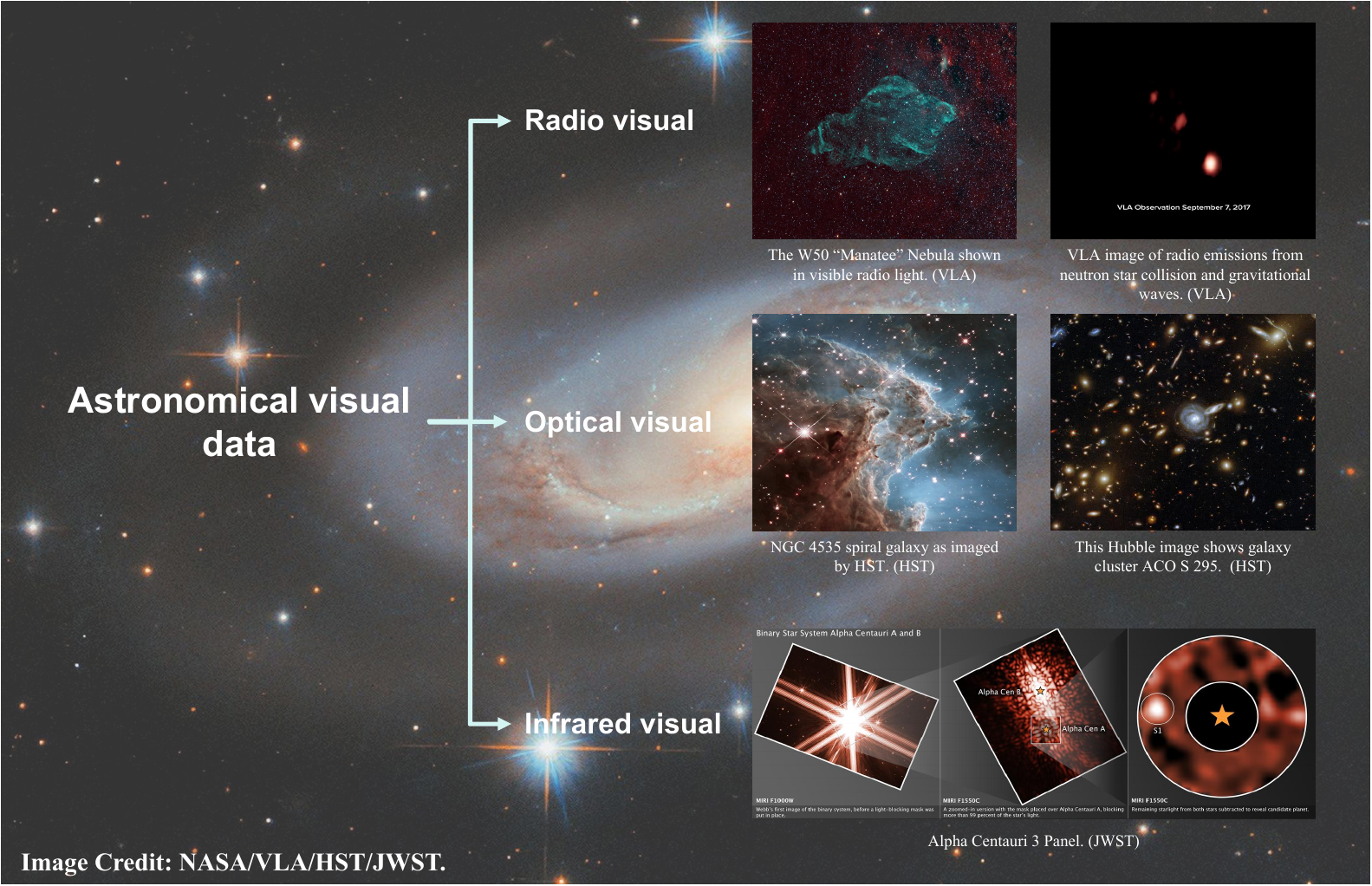}
    \caption{Examples of astronomical data, demonstrating the application of radio signals, optical signals, and infrared signals in imaging different astronomical objects. The image is sourced from \href{https://www.nasa.gov/}{NASA}.}
    \label{fig:astro_visual_data}
    \vspace{-8mm}
\end{figure}

At tissue and organ scales, radiological imaging encompasses multiple modalities including X-rays~\cite{johnson2019mimic,demner2015openi}, computed tomography (CT)~\cite{ji2022amos,hamamci2024developing,nlst2013tcia}, histopathology~\cite{kumar2017dataset}, magnetic resonance imaging (MRI)~\cite{wald2024openmind,d2024totalsegmentator}, ultrasound~\cite{aldhabyani2020busi,yang2025ahu}, positron emission tomography (PET)~\cite{gatidis2022autopet,gatidis2023autopet}, and mammography~\cite{suckling1994mammographic}, each revealing different aspects of internal anatomy and function. These images, commonly stored in DICOM~\cite{tool-dicom} or NIfTI~\cite{tool-nifti} formats with rich metadata, can be processed using specialized viewers like RadiAnt~\cite{tool-radiant} and MRIcroGL~\cite{tool-mricrogl} or programmatic libraries such as pydicom~\cite{tool-pydicom} and SimpleITK~\cite{tool-simpleitk}. Clinical imaging extends to specialized domains like ophthalmology with color fundus photography (CFP)~\cite{staal2004drive,hoover2000stare,de2023airogs}, fundus fluorescein angiography (FFA)~\cite{wu2024mm}, ophthalmology~\cite{ding2020primefp20} and optical coherence tomography (OCT)~\cite{liang2025novel,kermany2018identifying}, dermatology for skin lesion analysis~\cite{tschandl2018ham10000,codella2019skin} ophthalmic surgical microscopy for high-resolution intraoperative visualization in ophthalmic procedures~\cite{hu2024ophnet,hu2023nurvid,hu2025towards,li2025ophora}, and endoscopy for surgical guidance~\cite{jha2019kvasir,twinanda2016cholec,endovis2025}. These visual data, once paired with their descriptions and reports, hold great potential in developing healthcare MLLMs; visualization examples are shown in Fig.~\ref{fig:medical_modality}.

At macroscopic scales, natural photographs capture biodiversity through datasets like iNaturalist~\cite{van2018inaturalist}, while agricultural visual data span from micro-level plant imaging to macro-level UAV and satellite imagery for crop monitoring~\cite{arbab2024ageval,dongre2025mirage,chiu2020agriculture}. 
Earth science leverages satellite remote sensing~\cite{justice2002overview,moreira2013tutorial} and atmospheric datasets~\cite{hersbach2020era5,rasp2020weatherbench} for climate modeling and environmental monitoring. As shown in Fig.~\ref{fig:earth_vis_data}, due to the diversity of their collection sources, earth observation data exhibit significant variability. For instance, some data are obtained from ground-based observation stations, offering long-term and continuous records at specific locations. Other datasets are derived from multispectral remote sensing technologies, which provide comprehensive information on surface and atmospheric characteristics across larger spatial scales. Additionally, reanalysis data~\cite{hersbach2020era5} integrate observational records with numerical models, resulting in meteorological and environmental parameters with enhanced temporal and spatial consistency. These various types of data each possess unique features in terms of spatial coverage, temporal resolution, and observational content, offering a multi-dimensional information foundation for research in earth system science.
Beyond Earth, astronomical observations across the radio interferometry~\cite{Thompson2017_VLA} to optical~\cite{HST2025,York2000_SDSS} and infrared~\cite{JWST}, capture celestial phenomena, complemented by spectroscopic data from instruments like Large sky Area Multi-Object fiber Spectroscopic Telescope (LAMOST)~\cite{LAMOST} that reveal chemical compositions and stellar dynamics, as illustrated in Fig.~\ref{fig:astro_visual_data}.



Complementing direct imaging, diagrammatic figures and spectroscopic visualizations provide crucial abstractions of scientific knowledge that cannot be captured through photography alone. Molecular structure diagrams, increasingly recognized as natural interfaces for chemical AI systems~\cite{tan2025chemmllm}, have been curated into large-scale datasets for tasks ranging from image captioning to property prediction~\cite{edwards2022translation,wang2017pubchem,fu2021moler}. Schematic diagrams and conceptual illustrations from scientific literature~\cite{lozano2025biomedica,lin2023pmcoa,kembhavi2016aidc,jobin2019docfigure} distill complex processes and experimental setups into accessible forms, essential for both human understanding and AI interpretation. These diverse visual modalities from atomic-resolution microscopy to cosmic surveys, and from molecular diagrams to climate visualizations, collectively form a rich multimodal foundation for scientific AI systems. The integration of these varied visual elements into comprehensive datasets like MaCBench~\cite{alampara2024macbench} and MMSci~\cite{li2025mmscidatasetgraduatelevelmultidiscipline} enables models to synthesize knowledge across disciplines, though challenges remain in aligning dense visual information with semantic textual descriptions, particularly for complex phenomena in molecular biology, materials science, 
and mathematical physics that require advanced multimodal learning techniques.


\subsubsection{Symbolic Representations}
\label{sec:background_taxonomy_symbolic}

Symbolic representations constitute a fundamental data modality in scientific computing, providing abstract, non-numeric encodings of scientific entities, relationships, and laws that are both human-interpretable and machine-processable. These representations include molecular structures encoded as string notations, such as Simplified Molecular-Input Line-Entry System (SMILES) strings~\cite{weininger1988smiles}, International Chemical Identifier (InChI) codes~\cite{heller2009iupac},  Self-Referencing Embedded Strings (SELFIES)~\cite{selfies}), Crystallographic Information Files (CIF) for material structures, and parameterized equations for physics and Earth system modeling. The significance of symbolic data lies in its ability to encode complex scientific knowledge in compact, manipulable forms that preserve semantic meaning while enabling automated reasoning, transformation, and discovery operations critical for modern scientific computing.

\begin{figure}[t!]
 \centering
 \includegraphics[width=0.98\linewidth]{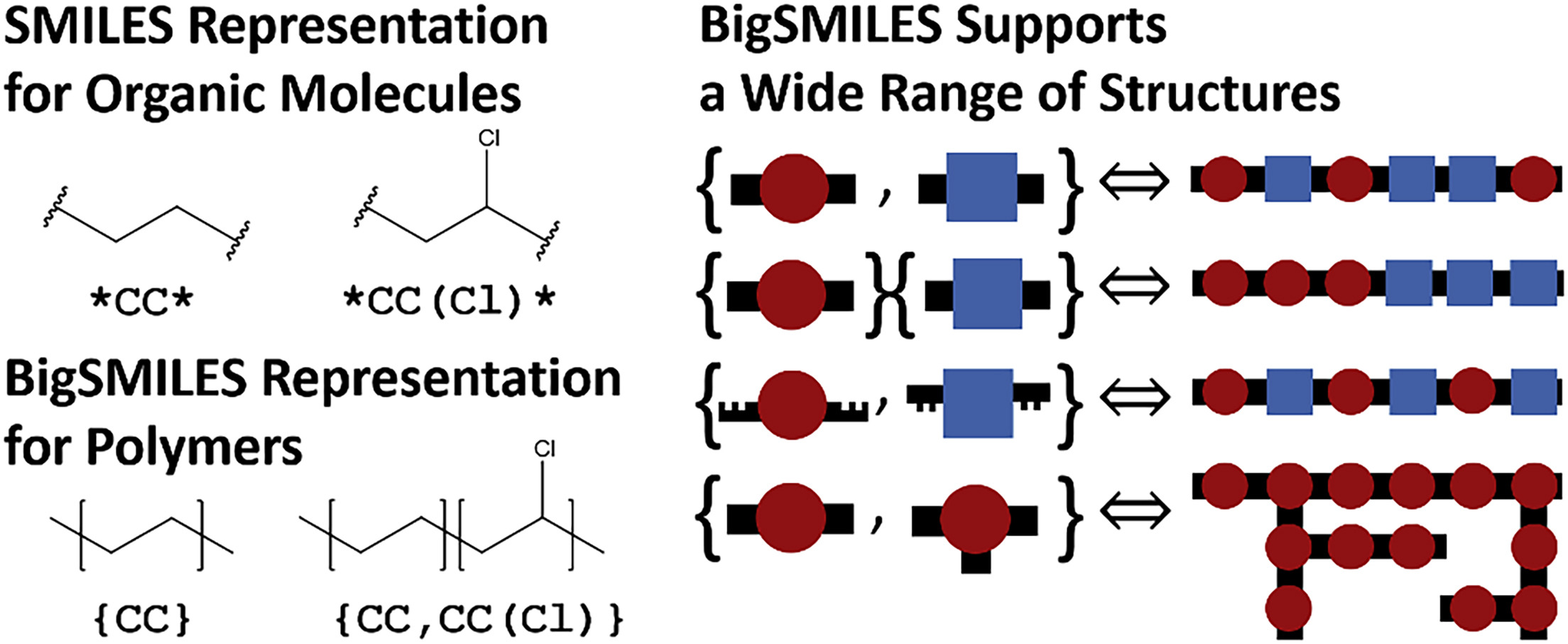}
 \caption{Schematic of BigSMILES representations from Lin \etal~\cite{lin2019bigsmiles}. Polymers are represented as monomers (repeating units) enclosed within curly brackets; the curly brackets indicate that the molecule is a stochastic object. The monomers are represented as SMILES strings, with additional information expressing the connectivity between monomeric units.}
 \label{fig:big_smiles}
\end{figure}

\begin{figure}[t!]
    \centering
    \includegraphics[width=\linewidth]{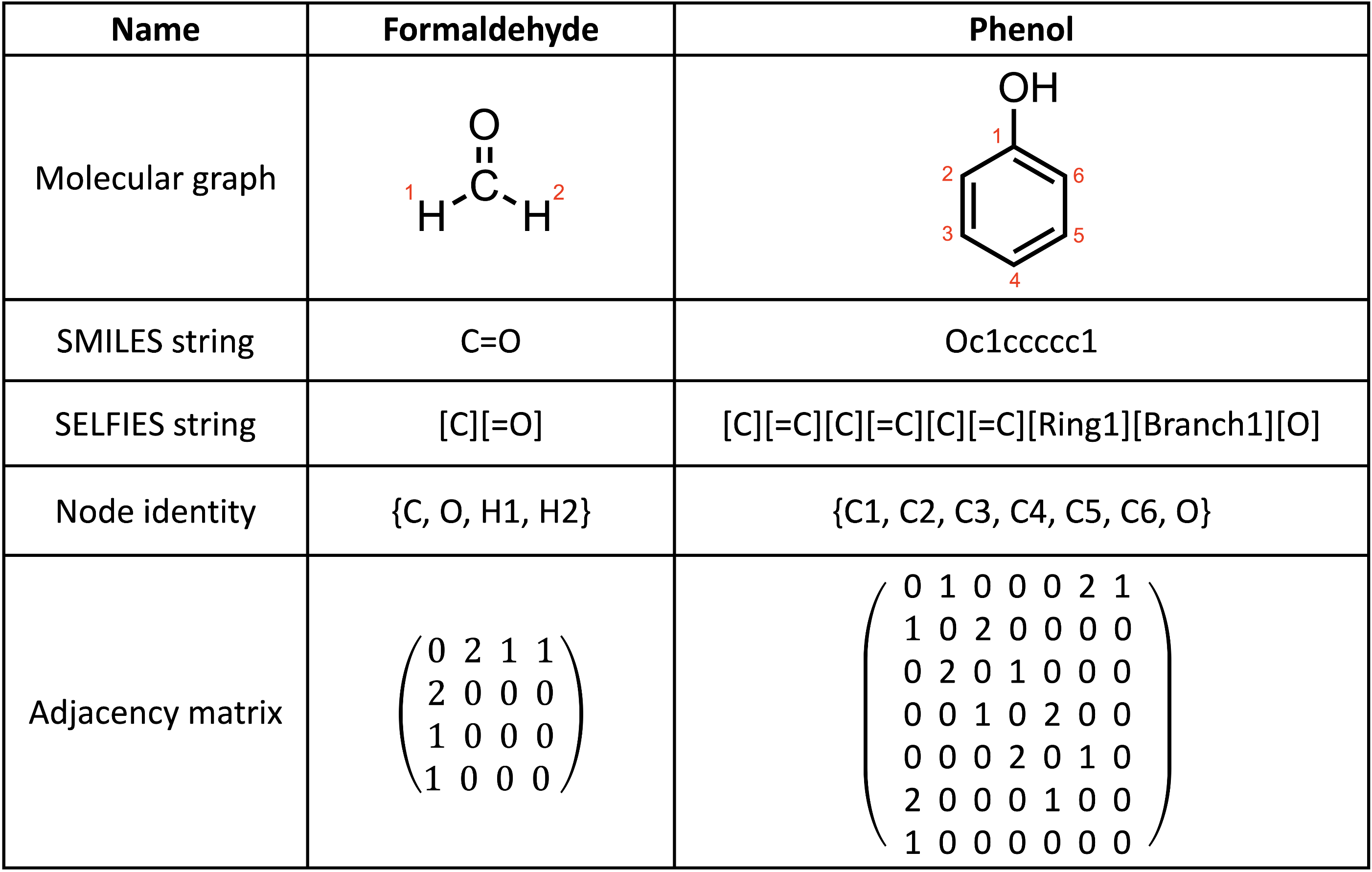}
    \caption{Exemplified symbolic representations (cheminformatics) of formaldehyde and phenol: molecular graph, SMILES and SELFIES string, node identity, and adjacency matrix. Hydrogens are typically omitted in SMILES and SELFIES strings. In the adjacency matrix, edge weights reflect bond types: 1 for single bonds, 2 for double bonds, and 3 for bonds in the aromatic ring.}
    \label{fig:molecular_graph}
\end{figure}

The most prevalent symbolic representations in chemistry and materials science are string-based molecular encodings, with SMILES~\cite{weininger1988smiles} being the de facto standard since the 1980s. SMILES is a specification in the form of a line notation for describing the structure of chemical species using short ASCII strings, encoding molecular structures using ASCII strings with specific rules: atoms are represented by their chemical element symbols (often with brackets omitted), bonds by symbols including ``-'' (single), ``='' (double), ``\#'' (triple), ``:'' (aromatic), rings by breaking cycles and adding matching numbers (\eg, ``O1CCOCC1'' for 1,4-Dioxane), aromatic rings using lowercase letters or alternating bonds (\eg, ``c1ccccc1'' for benzene), and branches using parentheses (\eg, ``CCC(=O)O'' for propionic acid). An extension of SMILES for polymers is BigSMILES~\cite{lin2019bigsmiles}, which represents polymers as stochastic objects with monomers enclosed in curly brackets, as illustrated in Fig.~\ref{fig:big_smiles}. However, SMILES suffers from syntactic fragility—small perturbations can render strings invalid. To address this, SELFIES (SELF-referencing Embedded Strings)~\cite{krenn2020self} was introduced in 2020, guaranteeing 100\% validity through formal grammar rules. SELFIES uses a vocabulary of tokens like ``[C]'', ``[=O]'', ``[Branch]'', ``[Ring]'' with localized markers for branches and rings, enabling robust left-to-right parsing that gracefully handles errors. Fig.~\ref{fig:molecular_graph} shows examples of Formaldehyde and Phenol's molecular graphs and corresponding SMILES and SELFIES strings. The difference between SMILES, BigSMILES, and SELFIES is demonstrated in Table~\ref{tab:string_representations}. Beyond strings, molecular graphs provide more intuitive representations where nodes correspond to atoms and edges to bonds, with adjacency matrices encoding connectivity and bond types~\cite{coley2017convolutional}. Recent benchmark~\cite{gao2022samples} reveals that SMILES remains most expressive for molecular optimization tasks, while SELFIES often underperforms due to redundancy.


For crystalline materials, the CIF format serves as the standard, encoding unit cell parameters (lattice constants $a,b,c$, angles $\alpha,\beta,\gamma$), atomic positions in fractional coordinates, space group symmetries, and experimental metadata in a structured key-value format readable by tools like pymatgen and VESTA. These representations underpin major databases including ZINC~\cite{irwin2012zinc}, ChEMBL~\cite{gaulton2012chembl}, USPTO~\cite{Lowe2017}, ICSD, and the Materials Project~\cite{jain2013commentary}, as well as benchmarks like MoleculeNet~\cite{wu2018moleculenetbenchmarkmolecularmachine} and MatBench~\cite{dunn2020benchmarking}.

In physics and astronomy, symbolic representations extend beyond structural encodings to encompass mathematical expressions, differential equations, and theoretical frameworks that enable automated scientific discovery. At the core are algebraic equations, differential/integral forms, and probability distributions, with recent work demonstrating that LLMs performing symbolic derivation, \ie, keeping variables symbolic before late-stage numerical substitution, tend to achieve higher accuracy on physics problem solving compared with numeric-first approaches~\cite{Dan2025_SymbolicNumerical}. Equation graphs represent variables and operators as nodes, enabling graph-based symbolic regression; for instance, graph networks trained on force-law data successfully recover Newton's law through message-passing outputs~\cite{Cranmer2019_SymbolicPhysics}. Building on this foundation, LLM-powered methods like Dual Reasoning Symbolic Regression integrate language model reasoning with reflective optimization for equation extraction~\cite{Wang2025_DrSR}. In astronomy, systems like PhyE2E~\cite{PhyE2E} demonstrate end-to-end neural symbolic regression, generating dimensionally consistent formulas from diverse sources including NASA's THEMIS mission data~\cite{THEMIS}, AI Feynman datasets~\cite{Feynman,Feynman2}, and solar observation data (SILSO)~\cite{SILSO}. Similarly, Earth science employ symbolic representations through mathematical formula fitting and regression for modeling complex phenomena governed by partially understood physics, such as the Navier-Stokes equations~\cite{constantin1988navier} in atmospheric motion, wave equations in seismology~\cite{bormann2012seismic}, and shallow-water equations in oceanography~\cite{le1998finite}. These models utilize parameterization schemes and regression analysis (least squares, Bayesian inference) to align theoretical predictions with observational data, demonstrating how symbolic representations serve as a bridge between empirical observations and theoretical understanding across scientific disciplines.

\begin{table*}[t!]
\centering
\caption{Comparison of SMILES, BigSMILES, and SELFIES representations.}
\label{tab:string_representations}
\resizebox{\textwidth}{!}{
\begin{tabular}{lccc}
\toprule
\textbf{Feature} & \textbf{SMILES}~\cite{weininger1988smiles} & \textbf{BigSMILES}~\cite{lin2019bigsmiles} & \textbf{SELFIES}~\cite{krenn2020self} \\
\midrule
\textbf{Primary domain} & Small molecules & Polymers and macromolecules & Small molecules \\
\textbf{Syntax basis} & ASCII strings with chemical rules & SMILES syntax + curly bracket extensions & Tokenized grammar rules \\
\textbf{Connectivity encoding} & Explicit bonds, rings, branches & Bonds, rings, branches + bonding descriptors (\texttt{[*]}) & Encoded via grammar tokens \\
\textbf{Stochastic representation} & Not supported & Supported via curly brackets & Not supported \\
\textbf{Polymer architecture} & Not supported & Supports block, random, graft, branched & Not supported \\
\textbf{Error tolerance} & Fragile—small changes can break validity & Same as SMILES for monomers & Guaranteed 100\% valid \\
\textbf{Typical example} & \texttt{CCO} (ethanol) & \texttt{\{[*]CC[*]\}} (polyethylene) & \texttt{[C][C][O]} (ethanol) \\
\textbf{Advantages} & Compact, widely supported & Encodes polymer connectivity & Robust to syntax errors \\
\textbf{Limitations} & Syntactic fragility & Still fragile at monomer level & Redundancy, longer strings \\
\bottomrule
\end{tabular}
}
\end{table*}

\subsubsection{Structured Data}
\label{sec:background_taxonomy_structured}

Structured data in scientific domains refers to information systematically organized through explicit, formal models that enable efficient querying, storage, and computational reasoning. Across disciplines, structured data follows a progression from simple tabular formats to complex knowledge representations. At the foundational level, data tables $T$ consisting of columns $\{c_i\}_{i=1}^C$ and rows $\{l_j\}_{j=1}^R$ serve as the basic organizational unit, with each cell $v_{ij}$ representing measurements or annotations. These tables, prevalent in resources like GEO~\cite{edgar2002gene}, dbSNP~\cite{sherry2001dbsnp}, and weather station datasets such as WEATHER-5K~\cite{han2024weather}, provide straightforward data organization but lack explicit semantics or inter-attribute relationships. Building upon this foundation, relational databases $D = \{T_1, T_2, \dots, T_N\}$ extend tables with schema-level constraints and referential integrity, where foreign key pairs $(c_i^{(k)}, c_j^{(h)})$ connect columns across tables, enabling complex queries over diverse entities as seen in Ensembl~\cite{cunningham2022ensembl} and UniProtKB~\cite{uniprot2023uniprot}.

The evolution toward more expressive representations includes ontologies and knowledge graphs that capture domain-specific semantics and relationships. Ontologies formally represent concepts and their relationships using languages like Web Ontology Language~\cite{mcguinness2004owl} or Open Biological and Biomedical Ontologies~\cite{smith2007obo}, defining classes, properties, and hierarchies for semantic interoperability and logical inference, exemplified by the Gene Ontology~\cite{ashburner2000gene} and Human Phenotype Ontology~\cite{kohler2021hpo}. A knowledge graph is a collection of relational facts \(G \subseteq \mathcal{E}\times \mathcal{R}\times \mathcal{E}\), where \(\mathcal{E}\) denotes the set of entities and \(\mathcal{R}\) the set of semantic relations. By integrating heterogeneous data into a unified semantic representation, knowledge graphs facilitate knowledge reasoning and discovery~\cite{lin2024improving, liang2024survey}, as exemplified by UMLS~\cite{bodenreider2004umls} and PrimeKG~\cite{chandak2023building}; similarly, CLLMate~\cite{li2024cllmate} aligns meteorological records with climate events. Taken together, these developments form a structured data ecosystem supported by standardized exchange formats—including CSV, XML, JSON, YAML, HDF5, ROOT, FITS, and NetCDF—that ensure traceability and interoperability across disciplines. Large-scale repositories have emerged as critical infrastructure, from molecular libraries like ZINC~\cite{irwin2012zinc} and ChEMBL~\cite{gaulton2012chembl} storing compounds in SMILES format~\cite{anderson1987smiles,huang2021therapeutics}, to physics archives like CODATA~\cite{CODATA2022} and particle physics databases~\cite{PDG2024}, astronomical catalogs including SIMBAD~\cite{SIMBAD} and VizieR~\cite{VizieR}, materials databases such as the Materials Project~\cite{jain2013commentary} and MatBench~\cite{dunn2020benchmarking}.

The sophistication of structured data extends to specialized property datasets that enable targeted scientific investigations. In chemistry, ADMET (Absorption, Distribution, Metabolism, Excretion, and Toxicity) databases~\cite{wang2016admet,huang2021therapeutics} provide comprehensive pharmacokinetic properties including absorption (Bioavailability~\cite{bioavail}, HIA~\cite{hou2007adme}), distribution (BBB~\cite{martins2012bayesian}, FreeSolv~\cite{mobley2014freesolv}), metabolism (Clearance-AstraZeneca~\cite{gaulton2017chembl}), excretion (VDss~\cite{lombardo2016silico,chen2021data}), and toxicity (ClinTox~\cite{wu2018moleculenetbenchmarkmolecularmachine}, ToxCast~\cite{richard2016toxcast}, Tox21~\cite{ncats2014tox21}) measurements crucial for drug discovery. Similarly, gravitational-wave catalogs like GWTC~\cite{GWTC3} document events with detailed source parameters in machine-readable formats, while materials databases provide multi-property coverage including electronic, thermodynamic, and mechanical behaviors computed under standardized protocols. These structured resources leverage persistent identifiers and metadata standards, facilitating rich scholarly analyses through bibliographic knowledge graphs like INSPIRE-HEP~\cite{INSPIREAPI} and NASA ADS~\cite{kurtz2000nasa}, ultimately enabling robust predictive modeling and efficient exploration of vast scientific spaces across all disciplines.

\subsubsection{Time-Series Data}
\label{sec:background_taxonomy_time}

\begin{figure}
 \centering
  \includegraphics[width=0.9\linewidth]{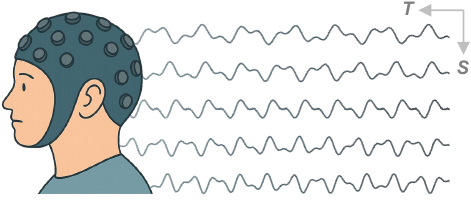}
  \captionsetup{type=figure}
    \caption{Five-channel EEG recording setup and corresponding time series data. Horizontal axis: time (T); Vertical axis: individual EEG channels showing brain electrical activity patterns recorded from scalp electrodes. Figure is adapted from CSBrain \cite{zhou2025csbrain}.}
  \label{fig:EEG_signal}
  \vspace{-5mm}
\end{figure}

Time series data, characterized by sequences of temporal data points collected at certain intervals~\cite{esling2012time,box2015time,lim2021time}, constitutes a fundamental data modality across scientific disciplines, capturing dynamic phenomena from nanoseconds to decades. These data enable the analysis of temporal patterns, periodicity, and system evolution across vastly different scales—from molecular dynamics tracking atomic positions $\{\mathbf{X}^{(t)} \in \mathbb{R}^{N \times 3}\}_{t = 0}^T$, velocities $\{\mathbf{V}^{(t)} \in \mathbb{R}^{N \times 3}\}_{t = 0}^T$, and forces $\{\mathbf{F}^{(t)} \in \mathbb{R}^{N \times 3}\}_{t = 0}^T$ in datasets like MD17~\cite{chmiela2017machine} and ISO17~\cite{schutt2017schnet,schutt2018schnet}, to astronomical observations monitoring stellar brightness variations for exoplanet detection in missions like Kepler~\cite{kepler} and Five-hundred-meter Aperture Spherical Telescope (FAST)~\cite{FAST}. The temporal resolution spans milliseconds in neurophysiological recordings such as electroencephalogram (EEG)~\cite{binnie1994electroencephalography} capturing brain oscillations~\cite{yeung2004detection} and event-related potentials~\cite{light2010electroencephalography} (Fig.~\ref{fig:EEG_signal}), to hourly meteorological variables in the ERA5 dataset~\cite{hersbach2020era5} with 0.25-degree spatial resolution, and continuous seismic waveforms from Incorporate Research Institutions for Seismology~\cite{ahern2015iris} and United States Geological Survey networks~\cite{USGS2025_EarthquakeAPI} for earthquake monitoring.



The diversity of time-series modalities reflects the multiscale nature of scientific phenomena. In biological systems, time-series data capture dynamics from molecular-level gene expression patterns revealing temporal responses~\cite{bar2012studying,shalek2014single,deng2014single} to clinical monitoring through electrocardiogram (ECG)~\cite{willems1991diagnostic} for cardiac rhythm analysis~\cite{agrafioti2009ecg}, electromyogram (EMG)~\cite{kugelberg1947electromyograms} for muscle activity~\cite{merlo2003fast}, and continuous glucose monitoring~\cite{rodbard2016continuous,klonoff2017continuous}. Neuroimaging modalities provide complementary temporal and spatial resolutions: functional magnetic resonance imaging (fMRI) detects blood-oxygen-level-dependent (BOLD) signals~\cite{kim2012biophysical} for mapping brain networks~\cite{van2010exploring}, while magnetoencephalography (MEG) measures magnetic fields from neuronal activity~\cite{fred2022brief,vrba2001signal}. In chemistry, molecular spectrum data mainly include Raman, infrared (IR), ultraviolet (UV), $^1$H nuclear magnetic resonance (NMR), and $^{13}$C NMR spectroscopy~\cite{telle2007laser}, revealing structural and compositional information enabling AI-driven representation learning~\cite{lu2024deep}. Physics leverages high-frequency strain data from LIGO/Virgo at 16,384 Hz for gravitational wave detection~\cite{GWTC3}, while SDO~\cite{Pesnell2012_SDO} provides Atmospheric Imaging Assembly Extreme Ultraviolet images every 12 seconds and Helioseismic and Magnetic Imager (HMI) vector-magnetogram-derived Space-weather HMI Active Region Patches features at 12-minute cadence to forecast space weather~\cite{Bobra2015_SHARP}.

These temporal datasets serve critical roles in understanding system dynamics, enabling predictive modeling, and monitoring critical events. Longitudinal clinical studies utilize serial MRI, CT, and clinical report data~\cite{menachemi2011benefits} to model disease trajectories~\cite{vos2013preclinical,niu2024enhancing}, while synoptic astronomical surveys like The Zwicky Transient Facility~\cite{ZTF} and Legacy Survey of Space and Time~\cite{LSST} generate calibrated image sequences for transient detection. Earth science integrates atmospheric data from WeatherBench~\cite{rasp2020weatherbench} and WEATHER-5K~\cite{han2024weather}, oceanic measurements from the Hybrid Coordinate Ocean Model (HYCOM)~\cite{chassignet2007hycom} and NOAA Tides~\cite{guest2006solutions}, and geophysical recordings for comprehensive Earth system monitoring. The standardization of these diverse time-series formats facilitates cross-disciplinary AI applications~\cite{IRIS2025_SeismicData,morid2023time,di2023explainable}, establishing time-series analysis as a cornerstone methodology for extracting insights from dynamic scientific phenomena across all scales.

\subsubsection{Multi-omics Integration}
\label{sec:background_taxonomy_multiomics}


\begin{figure}[t!]
    \centering
    \includegraphics[width=0.5\linewidth, angle=-90]{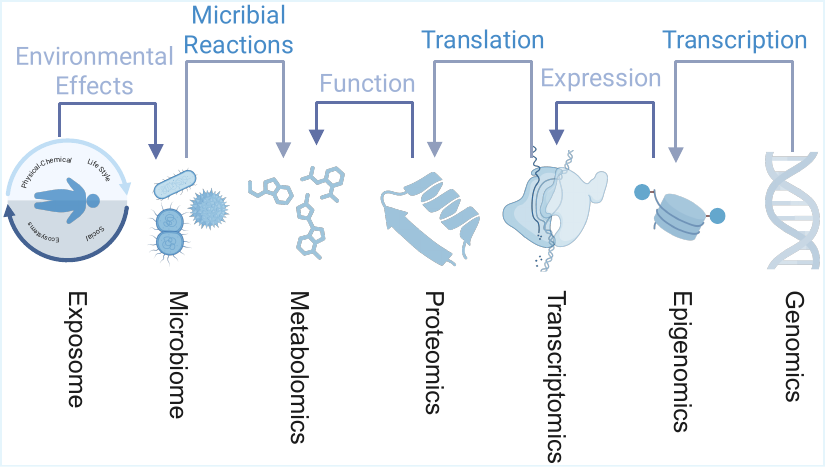}
    \caption{Multi-omics data landscape.}
    \label{fig:multiomics}
\end{figure}

\begin{figure}[t!]
    \centering
    \includegraphics[width=\linewidth]{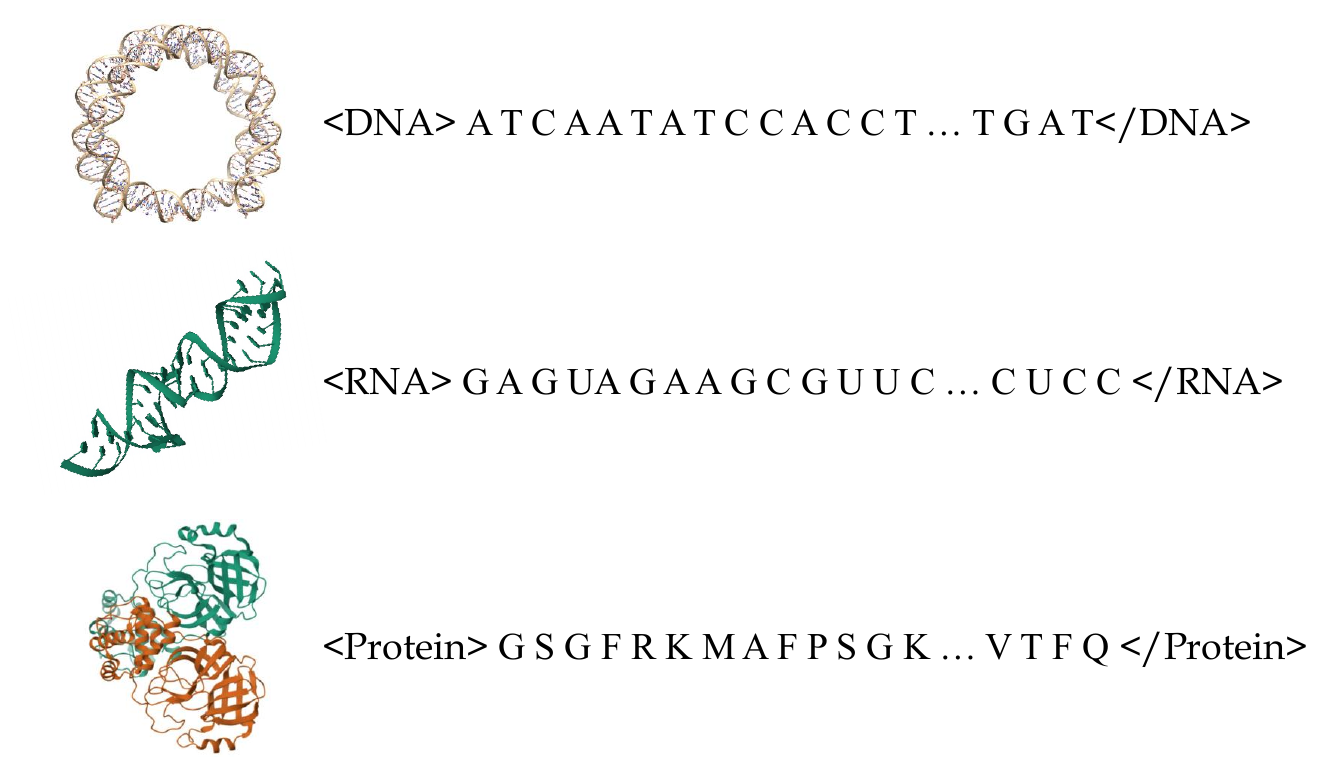}
\caption{Symbolic representations and 3D structure visualizations across different scientific domains: DNA, RNA and Protein. The DNA structure is split into chain I and chain J from PDB 1KX5 \cite{Davey2002im} and visualized by UCSF Chimera \cite{chimera2004software}.
The RNA structure is from the RNAsolo with ID 7ELQ~\cite{adamczyk2022rnasolo,bernard2024state}.
The protein snapshot is from the PDB bank with ID 7CAM \cite{Wang2020-vw}. The DNA and protein are adapted from NatureLM \cite{naturelm}.
}
\label{fig:example_symbolic_vis}
\vspace{-5mm}
\end{figure}

Driven by rapid advances in high-throughput technologies, multi-omics has emerged as a powerful approach for capturing the complexity of living systems through the integrated analysis of multiple layers of biological data~\cite{hasin2017multi}. As illustrated in Fig.~\ref{fig:multiomics}, the multi-omics landscape encompasses seven major data modalities: genomics (capturing genetic sequences and variations), epigenomics (mapping regulatory modifications), transcriptomics (profiling gene expression), proteomics (analyzing protein abundance and function), metabolomics (measuring small molecule metabolites), microbiome (characterizing microbial communities and their functions/interactions), and exposome (tracking environmental effects). These omics layers are interconnected through biological processes, from transcription and translation at the molecular level to environmental interactions at the systems level, offering complementary insights that together enable a more comprehensive understanding of biological processes than any single layer alone~\cite{subramanian2020multi,xie2024tracing}. At the molecular core of this framework, biological information flows from DNA to RNA to proteins, with each biomolecule existing in both symbolic sequence representations and three-dimensional structural forms (Fig.~\ref{fig:example_symbolic_vis}).



Multi-omics technologies have continued to advance, offering improved resolution, accuracy, and scalability, along with enhanced methods for integrating data across different biological domains~\cite{rappoport2018multi,chen2021data,lin2024st, ren2024comet}. As a result, multi-omics has emerged as a cornerstone of modern scientific research, providing deeper insights into the molecular mechanisms underlying health and disease, unraveling complex regulatory networks, and driving data-informed discoveries across diverse biological domains~\cite{karczewski2018integrative}.

Genomics encompasses a vast and evolving ecosystem of structured, symbolic and sequence-based representations. \emph{(i)} Reference genomes, such as those hosted by Ensembl~\cite{hubbard2002ensembl} and UCSC Genome Browser~\cite{kent2002human}, provide curated nucleotide sequences and annotated genomic elements across thousands of species. \emph{(ii)} Genetic variation, arising from differences in DNA sequences across individuals or populations, is a central focus of genomics. Population-scale resources such as GWAS Catalog~\cite{gwas_cata}, dbSNP~\cite{sherry2001dbsnp} and gnomAD~\cite{karczewski2020mutational} catalog common and rare variants, providing estimates of allele frequencies across diverse cohorts, while ClinVar connects specific variants to clinical phenotypes and pathogenicity interpretations~\cite{landrum2016clinvar}. \emph{(iii)} Functional genomics maps, such as those from ENCODE and Roadmap Epigenomics~\cite{kundaje2015integrative}, layer chromatin accessibility, histone marks, DNA methylation, and transcription factor binding profiles onto the genome to reveal regulatory landscapes.  \emph{(iv)} Spatial genome resources~\cite{lieberman2009comprehensive,rao20143d}, including Hi-C datasets and 3D genome browsers, reconstruct chromatin topology to explore long-range regulatory interactions. Genomic data are inherently symbolic and sequential, with rich metadata and controlled vocabularies~\cite{kohler2021hpo}—features that make them well-suited for conversion into prompt-based representations for language models~\cite{zhou2023dnabert,dalla2025nucleotide}. Emerging methods already leverage large-scale variant catalogs~\cite{karczewski2020mutational} and knowledge graphs~\cite{chandak2023building} to train foundation models for genotype-phenotype reasoning, while multi-resolution integration with imaging or epigenetics supports causal inference at cellular and organismal scales. 

Transcriptomics captures the dynamic and context-specific landscape of gene expression, linking genome to phenotype in time and space. Its data ecosystem spans multiple layers that together provide a comprehensive view of transcriptional activity. \emph{(i)} Transcript annotations from sources like GENCODE~\cite{frankish2019gencode} and RefSeq~\cite{o2016reference} define exon–intron structures, splice variants, and isoform-level expression. \emph{(ii)} At the foundational level, bulk RNA-seq and single-cell RNA-seq repositories such as GEO~\cite{edgar2002gene}, and ArrayExpress~\cite{kolesnikov2015arrayexpress} house millions of transcriptomic profiles across tissues, conditions, and perturbations.  \emph{(iii)} Expression atlases, such as the Human Cell Atlas or GTEx~\cite{gtex2020gtex}, enable comparative and tissue-specific analyses of transcriptional activity.  \emph{(iv)} Spatial transcriptomics platforms, including 10x Genomics Visium~\cite{tenxgenomics2025visium}, Slide-seq~\cite{rodriques2019slideseq}, and Stereo-seq~\cite{chen2022stereoseq}, link gene expression profiles to precise tissue coordinates, enabling spatially resolved analyses of cell-cell interactions, microenvironmental heterogeneity, and histopathological context. Public repositories like SpatialDB~\cite{fan2020spatialdb} aggregate thousands of such datasets across diverse species and conditions, facilitating cross-study comparisons and integration with histology images. 
\emph{(v)} Gene co-expression networks, such as STRING~\cite{szklarczyk2019string} co-expression edges, provide functional grouping of genes based on correlated activity. 
These transcriptomic resources form a rich, structured, and temporally resolved representation of cellular states, readily convertible into graph-, token-, or prompt-based formats for integration with other omics layers in large-scale modeling.

Proteomics is often described as \emph{multimodal}, but, strictly speaking, the field rarely couples images with free text in the way vision-language benchmarks do.  Instead, it juggles \emph{molecular representations} drawn from distinct information channels: \emph{(i)} structured knowledge bases such as UniProtKB deliver expertly curated sequences, domains and post-translational modifications for more than 250 million proteins~\cite{uniprot2023uniprot}, among them, the reviewed subset UniProtKB/Swiss-Prot (0.57 million entries, as of August 2025) is the most widely used; while the Protein Data Bank (PDB) stores atomic coordinates for experimentally determined folds~\cite{berman2000protein} 
; \emph{(ii)} interaction networks fuse biochemical and genetic evidence—STRING merges literature, co-expression and synteny to build genome-wide association graphs\cite{STRING2023}, whereas BioGRID~\cite{BioGRID2019} and IntAct~\cite{IntAct2004} record bench-validated contacts; \emph{(iii)} symbolic ontologies provide a shared semantic layer, with the Gene Ontology defining controlled terms for function, process and localization~\cite{GO2021}; \emph{(iv)} image resources such as the Human Protein Atlas place thousands of proteins into tissue and cellular context by immunohistochemistry and fluorescence microscopy~\cite{HPA2015}; \emph{(v)} computational structure repositories, notably the AlphaFold Protein Structure Database, extend empirical coverage with high-confidence models for millions of previously unsolved proteins~\cite{AlphaFoldDB2024}; and \emph{(vi)} time-resolved quantitative datasets from mass-spectrometry pipelines are shared through the ProteomeXchange consortium~\cite{ProteomeXchange2020}, with PRIDE as its flagship archive~\cite{PRIDE2023}. Seamlessly combining these heterogeneous modalities yields synergistic insight, \eg, PDB experimental structures and AlphaFold DB predicted models (surfaced via PDBe-KB) jointly constrain interaction graphs from STRING, BioGRID, and IntAct; ontology-aware statistics translate large-scale microscopy screens into testable biological hypotheses; and longitudinal mass spectrometry experiments connect dynamic post-translational regulation to spatial relocalization inferred from imaging. Although corpora already formatted as dialogue for LLM training remain scarce, the underlying repositories constitute machine-readable graphs, tables and sequences that can be converted into textual prompts or retrieval-augmented contexts with minimal templating. Emerging pipelines therefore marry graph databases with transformer representation learning, reconcile identifiers across formats, and propagate uncertainty, all under FAIR standards~\cite{wilkinson2016fair} (Findable, Accessible, Interoperable, Reusable) such as MIAPE~\cite{taylor2007minimum} and ProteomeXchange-XML~\cite{ProteomeXchange2020}. As these resources expand and model architectures mature, a genuinely integrative, causally grounded ``digital proteome'' becomes feasible, where each protein is simultaneously encoded as sequence, structure, dynamic profile, network node and spatial image, ready for LLM-driven reasoning across the molecular landscape.

Beyond the molecular central dogma, additional omics layers provide complementary biochemical and environmental perspectives. Metabolomics profiles small-molecule metabolites to capture biochemical activity and phenotypic state, with repositories such as the Human Metabolome Database~\cite{wishart2018hmdb} and MetaboLights~\cite{haug2013metabolights} supporting pathway-level integration with other omics. Microbiome studies characterize the composition and functional potential of microbial communities through metagenomic and metatranscriptomic sequencing, with resources like the Human Microbiome Project~\cite{human2012structure} and MGnify~\cite{mitchell2020mgnify} enabling host–microbe interaction analyses. Exposome research examines the totality of environmental exposures, including diet, pollutants, and lifestyle factors, using chemical assays, wearable sensors, and curated biomarker databases such as Exposome‑Explorer~\cite{neveu2016exposome}. These layers extend multi-omics frameworks by linking molecular phenotypes to ecological and environmental contexts.
 
From precision medicine and cancer research to environmental science and agriculture, multi-omics data now empower researchers to tackle complex, interdisciplinary problems and generate holistic models of biological and ecological systems~\cite{misra2019integrated,xie2024pathmethy, he2025generalized}.


\subsection{Hierarchical Structure of Scientific Knowledge}
\label{sec:background_structure}

Scientific knowledge fundamentally differs from a flat collection of information. Instead, it manifests as a sophisticated hierarchical system that mirrors the progressive nature of human cognition and the evolutionary path of scientific discovery from phenomena to essence, from the concrete to the abstract. This inherent stratification resonates with established knowledge hierarchy models, most notably the DIKW (Data-Information-Knowledge-Wisdom) pyramid articulated by Ackoff~\cite{ackoff1989data} and systematically analyzed by Rowley~\cite{rowley2007wisdom}, which posits that knowledge emerges through qualitative transformations rather than mere accumulation. However, as Zeleny~\cite{zeleny1987management} observed in mapping knowledge forms from ``know-nothing'' through ``know-what'' and ``know-how'' to ``know-why,'' scientific inquiry demands a more nuanced taxonomy that captures both procedural and explanatory dimensions. Building upon these theoretical foundations while addressing the unique epistemological requirements of scientific practice, we propose a five-tiered framework encompassing factual, theoretical, methodological-technological, modeling-simulation, and insight levels. This stratification reflects what Baskarada and Koronios~\cite{baskarada2013data} characterize as the need to contextualize knowledge hierarchies within specific domains, incorporating the computational and instrumental dimensions essential to contemporary science. Each level represents not merely a repository of information but a distinct mode of understanding, exhibiting emergent properties that reflect the transformative nature of scientific knowledge construction. The following sections will systematically examine each stratum, revealing how this hierarchical architecture facilitates both the organization of existing knowledge and the generation of novel scientific insights.



\begin{figure}[tp!]
    \centering
    \includegraphics[width=0.95\linewidth]{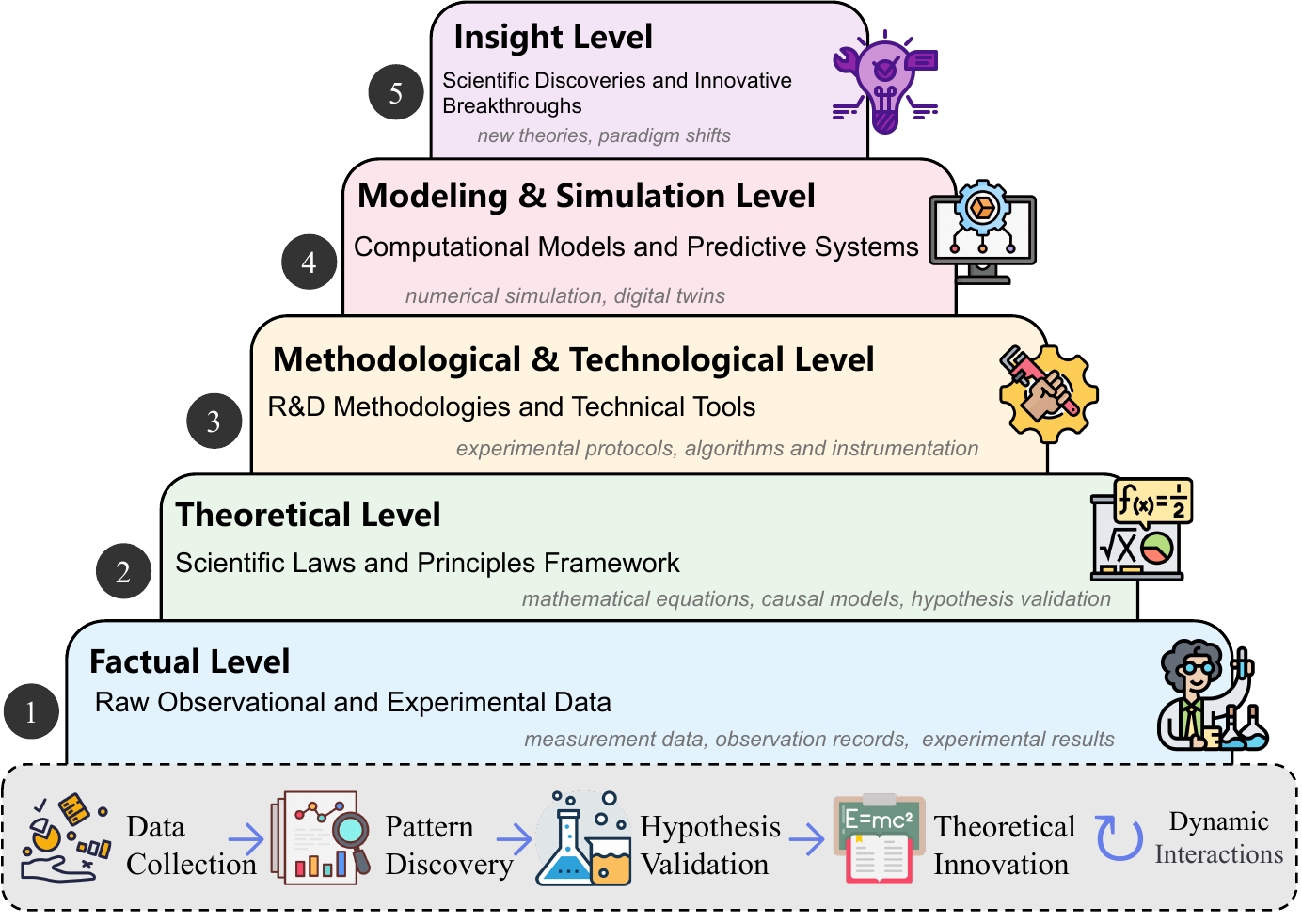}
    \caption{Hierarchical structure of scientific knowledge. The framework comprises five levels: factual (raw data), theoretical (laws and principles), methodological/technological (methods and tools), modeling/simulation (computational models), and insight (discoveries). The bottom panel illustrates the iterative cycle linking these levels through data collection, pattern recognition, hypothesis testing, and theory development.}
    \label{fig:sci_structure}
    \vspace{-2mm}
\end{figure}

To this end, we organize this subsection into five interconnected components, each representing a distinct level of scientific knowledge, as shown in Fig.~\ref{fig:sci_structure}. These levels include: 
the \textit{Factual Level} (Sec.~\ref{sec:background_structure_fact}), the \textit{Theoretical Level} (Sec.~\ref{sec:background_structure_theo}),  the \textit{Methodological and Technological Level} (Sec.~\ref{sec:background_structure_method}), the \textit{Modeling and Simulation Level} (Sec.~\ref{sec:background_structure_modeling}), and the \textit{Insight Level} (Sec.~\ref{sec:background_structure_insight}). In addition, we discuss \textit{Dynamic Interactions and Evolution} (Sec.~\ref{sec:background_structure_dynamic}) which highlights the iterative feedback loops across levels that collectively drive scientific progress. Finally, we conclude this subsection with the implication of such hierarchy (Sec.~\ref{sec:background_structure_implication}), which not only underscores the progressive deepening from data to discovery but also provides a structured foundation for developing Sci-LLMs that can effectively capture and utilize the multifaceted nature of scientific data.

\subsubsection{Factual Level}
\label{sec:background_structure_fact}

At the foundation of scientific knowledge lies the factual level—direct observational data, experimental measurements, and empirical evidence that constitute our primary interface with the physical world. This raw, unprocessed information serves as the bedrock for all subsequent scientific understanding.

Factual data is characterized by its objectivity and minimal human intervention. When astronomers collect astronomical imaging data, such as multi-band images~\cite{HST}, and additional light curves and spectra from distant galaxies, particle physicists capture collision events at the Large Hadron Collider~\cite{Evans2008_LHC}, gravitational-wave detectors measure strain signals~\cite{GWTC}, or biologists sequence genetic material~\cite{nurk2022complete}, they obtain direct representations of nature's state. Despite instrumental limitations, these data fundamentally reflect objective reality independent of theoretical frameworks.

Modern experiments generate data of unprecedented dimensionality and structural complexity. High-energy physics experiments like A Toroidal LHC Apparatus (ATLAS) and Compact Muon Solenoid (CMS) produce order-of-tens of terabytes of collision data per second~\cite{Evans2008_LHC}, while LIGO and Virgo release strain data sampled at 16,384 Hz~\cite{GWTC3}. This heterogeneity spans all domains: multi-channel neural recordings capture brain dynamics at millisecond resolution~\cite{buzsaki2012brain}, single-cell RNA sequencing reveals cellular heterogeneity with millions of transcripts~\cite{regev2017human}, multi-omics platforms integrate genomic, proteomic, and metabolomic data~\cite{hasin2017multi}, agricultural sensors monitor crop phenotypes across spatial and temporal scales~\cite{yang2017unmanned}, and Earth observation satellites generate multi-spectral imagery for climate monitoring~\cite{savtchenko2004terra}.

Critical to scientific data is its spatiotemporal context. Astronomical observations acquire meaning only when anchored by precise coordinates and timestamps, enabling cross-instrument calibration and transient detection. Self-supervised models that jointly encode images, spectra, and light curves demonstrate that meaningful representations emerge through multimodal fusion~\cite{rizhko2024self}. Similarly, seismic wave arrivals at distributed stations enable earthquake triangulation and Earth structure probing~\cite{fountoulakis20252024,evangelidis2021seismic}, while drug discovery relies on temporal pharmacokinetic profiles~\cite{reichel2015pharmacokinetics} and agricultural yield predictions depend on phenological timing~\cite{lobell2020eyes}. In the field of Earth science, the spatiotemporal characteristics of data are particularly prominent. This is primarily reflected in the fact that spatial scales of Earth science data often need to be mapped to specific geographic resolutions. For example, in~\cite{xu2024generalizing}, global meteorological variables are represented using a $128\times256$ tensor, providing a spatial discretization suitable for modeling over the entire globe. Regarding temporal resolution, different tasks require data at distinct time intervals. For some short-term nowcasting tasks~\cite{gong2024cascast, gong2024postcast}, data are typically recorded at 10-minute intervals, enabling the capture of rapidly evolving atmospheric phenomena. In contrast, for medium-range forecasting tasks~\cite{chen2023fengwu, xu2024extremecast}, data are usually sampled every 6 hours to balance data volume with the relevant timescales for prediction.

Inherent uncertainties and noise are integral to factual data. Quantum experiments face fundamental measurement limits~\cite{Heisenberg1927}, biological studies contend with individual variation and technical noise~\cite{brennecke2013accounting}, astronomical observations are severely degraded by atmospheric turbulence~\cite{hardy1998adaptive}, and clinical trials must account for patient heterogeneity~\cite{FDA2018HTE}. These uncertainties inform confidence bounds and guide robust analytical methods across all scientific disciplines.

\subsubsection{Theoretical Level}
\label{sec:background_structure_theo}

The theoretical level transcends empirical observations through diverse forms of abstraction and formalization. Beyond mathematical equations such as Newton's mechanics~\cite{Newton1687-hn}, Maxwell's electromagnetism~\cite{Maxwell1865-lu}, Schrödinger's quantum mechanics~\cite{PhysRev.28.1049}, and Hodgkin-Huxley neural dynamics~\cite{hodgkin1952quantitative}, scientific theories employ multiple representational frameworks.

Conceptual models capture fundamental principles: the central dogma in molecular biology~\cite{crick1970central}, plate tectonics in geoscience~\cite{vine1963magnetic}, and the Standard Model in particle physics~\cite{weinberg1967model}. Classification systems organize knowledge hierarchically: Linnaean taxonomy~\cite{linnaeus1789systema}, the periodic table~\cite{mendeleev1869relationship}, Gene Ontology~\cite{ashburner2000gene}, and astronomical object catalogs~\cite{VizieR}. Network representations reveal systemic relationships: protein interaction networks~\cite{barabasi2004network}, metabolic pathways~\cite{jeong2000large}, ecological food webs~\cite{dunne2002network}, and brain connectomes~\cite{sporns2005human}.
Computational models bridge theory and prediction: climate circulation models~\cite{washington2005introduction}, molecular dynamics simulations~\cite{karplus2002molecular}, population genetics algorithms~\cite{excoffier2005arlequin}, and pharmacokinetic compartmental models~\cite{rowland2011clinical}. Statistical frameworks quantify uncertainty: Bayesian inference in phylogenetics~\cite{huelsenbeck2001bayesian}, machine learning in multi-omics integration~\cite{hasin2017multi}, and cosmological parameter estimation~\cite{aghanim2020planck}.

These diverse theoretical representations exhibit hierarchical organization and domain-specific validity. Mathematical formalisms enable precise predictions; conceptual models provide intuitive understanding; classification systems facilitate knowledge organization; network models reveal emergent properties; computational approaches handle complexity. Together, they transform raw data into actionable scientific knowledge, creating a multi-layered theoretical infrastructure that supports discovery, prediction, and technological innovation across disciplines~\cite{simon2012architecture}.

\subsubsection{Methodological and Technological Level}
\label{sec:background_structure_method}

Between raw facts and abstract theories lies a crucial intermediate layer of methods and tools that transform theoretical predictions into testable hypotheses and raw data into theoretical insights.

Scientific methodology has evolved from simple comparative studies to sophisticated experimental designs across disciplines. Revolutionary techniques open new frontiers: CRISPR-Cas9 enables precise genomic editing~\cite{doudna2014new}, ultracold atom Bose-Einstein condensation paved the way for quantum simulation~\cite{AndersonScience.269.5221.198}, and high-throughput sequencing enables multi-omics profiling~\cite{hasin2017multi}.

Computational methods bridge theory and experiment. Monte Carlo algorithms~\cite{metropolis1953equation} underpin simulations from protein folding to climate modeling. Machine learning extracts patterns from massive datasets, \eg, AlphaFold~\cite{jumper2021highly} predicts protein structures, while algorithms identify astronomical objects and reconstruct neural circuits~\cite{helmstaedter2013connectomic}. Statistical frameworks ensure rigorous inference: particle physics commonly adopts a five-sigma threshold for discovery~\cite{aad2012observation}, while Bayesian approaches provide principled uncertainty quantification across fields~\cite{sivia2006data}.

Instrumental technologies extend observation into new realms. From Ruska's electron microscope~\cite{ruska1987development} to modern cryo-electron microscopes (cryo-EM), from LIGO's detection of $10^{-21}$-level spacetime strains~\cite{GWTC3} to single-cell sequencing~\cite{stuart2019integrative}, these tools fundamentally alter what questions we can ask. This creates feedback loops where better instruments enable deeper theories, which guide development of more sophisticated technologies.

\subsubsection{Modeling and Simulation Level}
\label{sec:background_structure_modeling}

This level involves utilizing numerical simulations to replicate complex systems. Virtual experiments enable researchers to test hypotheses and predict phenomena otherwise difficult or costly to study.

Contemporary modeling emphasizes multi-scale integration. Materials science connects quantum calculations at atomic scales to macro-level material behaviors~\cite{horstemeyer2009multiscale}. Climate modeling integrates short-term atmospheric processes with long-term ocean dynamics, bridging local weather and global climate change~\cite{taylor2012overview}. Astronomy links transient events like supernovae to long-term galaxy evolution spanning billions of years~\cite{vogelsberger2020cosmological}. Physics-informed neural networks merge physical laws and data-driven approaches, enabling effective data-physics fusion for fluid dynamics simulations with notable demonstrations from aerospace to biomedical applications~\cite{raissi2019physics, cai2021PINN}. Life sciences employ multi-scale models to explore molecular interactions and biological systems~\cite{kitano2002systems}. Computational simulations accelerate drug discovery by predicting molecular interactions~\cite{sliwoski2014computational}. Multi-omics approaches integrate genomic, proteomic, and metabolomic data to decipher disease mechanisms and guide personalized treatment~\cite{hasin2017multi}. Neuroscience simulations range from synaptic processes to brain-wide activity~\cite{markram2015reconstruction}, while agronomic models forecast crop performance under varying environmental conditions~\cite{asseng2013uncertainty}. Rigorous verification and validation processes ensure model reliability, confirming computational accuracy and predictive validity against experimental data, which is critical in nuclear engineering, aerospace, and medical certifications~\cite{oberkampf2010verification}.

Thus, the modeling and simulation level serves as a foundational tool, supporting modern scientific exploration and informed decision-making.
\subsubsection{Insight Level}
\label{sec:background_structure_insight}

At the apex of the scientific hierarchy, the insight level represents transformative moments when disparate knowledge coalesces into revolutionary understanding. Cross-disciplinary fusion has repeatedly catalyzed such breakthroughs: Shannon's information theory meeting molecular biology birthed bioinformatics, revealing life as an information processing system~\cite{shannon1948mathematical, watson1953molecular}; neuroscience converging with physics produced brain imaging technologies that decode neural activity patterns~\cite{ogawa1990brain}; astronomical spectroscopy combined with quantum mechanics unveiled stellar nucleosynthesis, explaining element formation across the cosmos~\cite{burbidge1957synthesis}. These interdisciplinary insights demand intellectual flexibility to recognize patterns across traditional boundaries, from protein folding dynamics mirroring energy landscape theory in physics~\cite{dill1997levinthal}, to agricultural genomics borrowing population genetics models to enhance crop resilience~\cite{varshney2012draft}.

Scientific revolutions often emerge from careful attention to anomalies that challenge existing frameworks. Classical physics predicted unbounded ultraviolet radiance at short wavelengths under the Rayleigh-Jeans law; Planck's quantization of energy in 1900 resolved this ``ultraviolet catastrophe'' and birthed quantum theory~\cite{Planck1901-vl}. Similarly, the discovery of reverse transcriptase shattered the central dogma of molecular biology~\cite{baltimore1970viral}, while anomalous galactic rotation curves revealed dark matter's existence~\cite{rubin1980rotational}. In pharmacology, unexpected drug side effects have led to therapeutic breakthroughs: sildenafil's transition from angina treatment to erectile dysfunction exemplifies serendipitous discovery through anomaly recognition~\cite{boolell1996sildenafil}. True conceptual innovation transcends problem-solving to introduce novel frameworks: Darwin's natural selection fundamentally altered our view of life's relationship to time~\cite{darwin2009origin}; plate tectonics unified previously disparate geological phenomena~\cite{wegener1980entstehung}; systems biology's emergence revealed that biological function arises from network interactions rather than isolated components~\cite{kitano2002systems}.

In the era of multi-omics and big data, extracting genuine insight requires navigating information overload through human-AI collaboration. Machine learning excels at pattern recognition across genomic, proteomic, and metabolomic datasets, uncovering disease signatures invisible to traditional analysis~\cite{hasin2017multi}. Yet human judgment remains essential for distinguishing correlation from causation, contextualizing discoveries within theoretical frameworks, and recognizing which patterns reflect fundamental principles. The future of scientific insight lies in this synergy, where computational power amplifies human creativity to reveal nature's hidden connections across scales from quantum to cosmic, from molecular to ecological.

\subsubsection{Dynamic Interactions and Evolution}
\label{sec:background_structure_dynamic}

Scientific progress emerges from dynamic interactions between hierarchical levels of knowledge, creating intricate feedback loops that drive discovery forward. This process manifests through three primary mechanisms: bottom-up induction, top-down deduction, and horizontal method transfer.

Inductive processes transform observations into theoretical understanding across disciplines. In astronomy, Kepler's analysis of Brahe's observations yielded planetary motion laws, later unified by Newton's gravitational theory. Modern life sciences follow similar trajectories: genomic sequencing reveals patterns explained through molecular and evolutionary models; neuroimaging data drives theories of brain function; agricultural field trials inform crop optimization strategies; and multi-omics integration uncovers systems-level biological principles. In physics, deduction channels theoretical insights into experimental design. Einstein's 1916 prediction of gravitational waves guided decades of detector development, culminating in LIGO and Virgo's detection of spacetime strains ($\sim10^{-21}$ m) from binary black-hole mergers in 2015, confirming century-old predictions and inaugurating gravitational-wave astronomy~\cite{abbott2016observation}.

Horizontal method transfer catalyzes unexpected advances. X-ray crystallography transitioned from mineralogy to revealing biomolecular structures; machine learning algorithms developed for image recognition now predict protein folding and drug-target interactions; network analysis from sociology illuminates ecological interactions and neural connectivity; spectroscopic techniques from physics enable remote sensing in Earth science and metabolomics profiling. This evolution follows a spiral pattern where theories transcend and include predecessors, \ie, classical mechanics subsumed within relativity and quantum mechanics, Mendelian genetics integrated with molecular biology, revealing why earlier frameworks succeeded within their domains while pointing toward a more comprehensive understanding. Such dynamic interactions are essential for developing AI systems that capture science's creative essence beyond pattern matching.


\subsubsection{Implications for Sci-LLMs}
\label{sec:background_structure_implication}
This hierarchical framework carries profound implications for the development and deployment of Sci-LLMs. Each level offers distinct computational challenges and opportunities for language model integration. At the \emph{factual level}, LLMs must learn to parse heterogeneous data formats, extract patterns from high-dimensional observations, and maintain spatiotemporal context, which is essential for tasks like automated literature mining and experimental data interpretation. The \emph{theoretical level} demands that models internalize mathematical formalisms, causal relationships, and domain-specific ontologies, enabling them to reason about scientific laws and generate testable hypotheses. The \emph{methodological level} requires LLMs to understand experimental protocols, computational workflows, and instrumental constraints, facilitating automated experiment design and method recommendation. At the \emph{modeling and simulation level}, language models can serve as interfaces between natural language queries and complex computational engines, translating scientific questions into simulation parameters and interpreting results. Finally, the \emph{insight level} challenges LLMs to perform cross-domain synthesis and creative hypothesis generation, capabilities that emerge from training on the full spectrum of scientific knowledge rather than isolated datasets. By incorporating data from all five levels, Sci-LLMs can transcend simple information retrieval to become active participants in the scientific discovery process, bridging human intuition with computational power.





\subsection{Key Challenges in Scientific AI}
\label{sec:background_challenges}

In the field of scientific AI, especially within LLMs and MLLMs, several key challenges must be addressed to enable meaningful scientific understanding and reasoning. These challenges include interpretability (Sec.~\ref{sec:background_challenge_interpretability}), cross-scale and multimodal integration (Sec.~\ref{sec:background_challenges_cross}), as well as dynamic knowledge evolvement (Sec.~\ref{sec:background_challenges_knowledge}), all of which are essential for enhancing the effectiveness of these models in scientific applications.

\subsubsection{Interpretability in Scientific AI}
\label{sec:background_challenge_interpretability}

Interpretability remains a major bottleneck. Scientific reasoning is inherently logical, based on clear explanations and justifications. However, LLMs and MLLMs are typically perceived as ``black-box'' models, making it difficult to understand the rationale behind a model's reasoning or output. This challenge is particularly acute in scientific domains, where understanding the ``why'' and ``how'' behind an answer is just as important as the answer itself. Interpretability is crucial for building trust in Sci-LLMs, especially in high-stakes fields such as drug discovery and climate modeling. 
In LLM/MLLM area, prompting or training the model with chain-of-thought (CoT)~\cite{wei2022chain,cot2} emerges as an effective technique to elicit explicit, natural-language reasoning capability of LLMs. CoT enables the model to write a step-by-step reasoning trace, breaking down complex tasks before giving the final answer. This makes the reasoning path more transparent and provides clearer insights into its decision-making. 
The recent work, BioReason~\cite{fallahpour2025bioreason}, introduces this multi-step reasoning strategy into DNA foundation models, enabling deep, interpretable biological reasoning from complex genomic data. By integrating a DNA foundation model with an LLM and constructing a biological CoT, BioReason empowers the LLM to directly process and reason with genomic information, fostering multimodal biological understanding. Through reinforcement learning, the model refines its multi-step reasoning capabilities, leading to biologically coherent deductions and outperforming traditional single-modality models on biological reasoning benchmarks.
Overall, conducting CoT reasoning in scientific AI models is particularly challenging due to the complexity and domain-specific nature of scientific knowledge. Unlike generalist models, scientific reasoning involves hypothesis-driven logic grounded in empirical evidence, requiring a precise understanding across disciplines such as biology, chemistry, and physics.
Therefore, more work is needed to develop transparent models that can offer both scientific accuracy and explainable reasoning.

\subsubsection{Cross‑scale and Multimodal Integration} 
\label{sec:background_challenges_cross}

Another major hurdle in the application of LLMs and MLLMs to scientific reasoning is their ability to handle cross‑scale and multimodal integration. Scientific data is often characterized by hierarchical structures that span multiple scales, from microscopic phenomena (\eg, molecular dynamics in chemistry) to macroscopic phenomena (\eg, weather patterns or ecosystem behavior). For example, in computational biology, understanding the behavior of a cell involves integrating data from individual molecules to entire tissues, which can require models to simultaneously process both fine-grained details and large-scale systems. Traditional LLMs excel at processing textual data but struggle to model spatiotemporal dependencies across scales. 
Moreover, scientific reasoning frequently involves multimodal data, typically combining text, images, numerical data, and experimental results. This requires models to seamlessly integrate heterogeneous data sources~\cite{multimodal1, multimodal2}. 
The challenge is further exacerbated when the information comes from different experimental setups or different measurement modalities, each requiring tailored processing pipelines that preserve important domain-specific features. For instance, bioinformatics deals with an extensive variety of data, including DNA, RNA, protein sequences, and drug molecules~\cite{ruan2025large}. MLLMs have the potential to address this complexity by integrating text, images, audio, and other modalities. They offer promising opportunities to enhance scientific understanding by connecting disparate data points and inferring relationships across these varied modalities. Initiatives such as the National Institutes of Health's ``Advancing Health Research through Multimodal AI''~\cite{NIH2025} exemplify this trend, aiming to develop data-driven multimodal AI approaches to model, interpret, and predict complex biological, behavioral, and health systems. However, significant challenges persist in achieving seamless multimodal integration. MLLMs frequently struggle with complex multimodal and multi-step reasoning tasks, often relying on shallow multimodal cues or defaulting to text-dominant reasoning rather than truly integrated understanding. A major bottleneck in their development is the scarcity of appropriate, high-quality multimodal scientific datasets.

To address these challenges, models need to move beyond isolated data streams and embrace a holistic integration of cross-scale and multimodal information to create truly unified frameworks that can seamlessly integrate complex scientific data and perform rigor scientific reasoning.

\subsubsection{Dynamic Knowledge Evolvement}
\label{sec:background_challenges_knowledge}

One of the most prominent challenges in applying LLMs and MLLMs to scientific domains is ensuring knowledge update and evolvement. In scientific research, knowledge evolves dynamically, with new discoveries constantly challenging existing theories. This makes it difficult for models trained on static datasets to maintain consistency with the most current body of scientific knowledge. Models that fail to continuously update their knowledge bases risk generating outdated or conflicting information, which can undermine their utility in domains like medical research, physics, or environmental science. 
To fix this, we need to explore new methods like automated knowledge injection and model adaptation. These approaches would allow models to continuously integrate new research findings, ensuring they remain coherent and aligned with the rapidly changing world of scientific discovery.
%

\subsection{Quality Standards for Scientific Datasets}
\label{sec:background_quality}

Assessing the quality of scientific data is essential for developing robust scientific AI models. In this subsection, we outline four complementary dimensions that together characterize data quality in scientific contexts. First, \textit{accuracy} (Sec.~\ref{sec:background_quality_acc}) assesses how faithfully data represent the underlying phenomena. Second, \textit{completeness} (Sec.~\ref{sec:background_quality_complete}) concerns the extent to which datasets capture all relevant elements across content, structure, and temporal coverage. Third, \textit{timeliness} (Sec.~\ref{sec:background_quality_timeliness}) measures the update frequency and responsiveness of datasets to real-world changes. Finally, \textit{traceability} (Sec.~\ref{sec:background_quality_trace}) ensures transparency and reproducibility by documenting provenance, metadata, and version histories. Together, these aspects provide a systematic framework for evaluating the reliability, usability, and long-term value of scientific datasets, standardizing data management practices and guiding optimal AI deployment.


\subsubsection{Accuracy} 
\label{sec:background_quality_acc}

Accuracy is one of the fundamental dimensions of scientific data quality, reflecting how closely data represent the real world in terms of spatial positioning, temporal annotation, and signal fidelity. High-accuracy data not only enhances the training efficiency and inference precision of AI models, but also directly impact the credibility of scientific conclusions. For example, in geospatial datasets, Landsat 8 satellite imagery, after ground control point correction, achieves a geolocation error of 15 to 30 meters, indicating high spatial precision~\cite{storey2014landsat}. In contrast, location information from some social media platforms is often only annotated at the city level, offering coarse granularity that hinders fine-grained modeling~\cite{cheng2010you}. In the physical sciences, the Materials Project provides data generated via first-principles calculations, controlling model errors, and ensuring reliable accuracy in band structure and lattice constants~\cite{jain2013commentary}. Common methods for assessing accuracy include mean squared error (MSE), root mean square error (RMSE), temporal alignment deviation, and signal-to-noise ratio (SNR), typically quantified by comparing with ground truth or high-quality benchmark datasets~\cite{batini2006data,pipino2002data}.

\subsubsection{Completeness} 
\label{sec:background_quality_complete}

Completeness refers to the extent to which a scientific data set adequately covers content, structural fields, and temporal span, whether it contains all the data elements that should have been collected. It serves as a foundation for systematic and logical data analysis. 
In genomics, completeness is often evaluated by sequencing depth; coverage below 10$\times$ is generally insufficient to accurately detect mutations, and modern whole genome sequencing standards typically require an average coverage of 30$\times$ or more~\cite{li2009sequence,sims2014sequencing}. 
In the field of materials science, data integrity directly determines the success or failure of data-driven discovery of new materials~\cite{ren2018accelerated}. 
Methods for assessing completeness include missing value statistics, field coverage analysis, breakpoint detection, and time series gap identification. For example, in Earth science, SCDNA~\cite{Tang2020_SCDNA} filled in missing data for precipitation, minimum temperature, and maximum temperature to ensure the data integrity across all weather stations, which improved the accuracy of spatial interpolation. Tools such as OpenRefine~\cite{tool-openrefine} and DataCleaner~\cite{tool-datacleaner} can automatically detect missing entries, structural anomalies, and null fields, thus improving the overall quality of datasets.

\subsubsection{Timeliness} 
\label{sec:background_quality_timeliness}

Timeliness measures data update frequency, the latency between data collection and release, and the speed at which data respond to real-world changes. This is crucial for applications like emergency response, trend forecasting, and dynamic modeling. For instance, during the COVID-19 pandemic, the Johns Hopkins University dataset was released at daily intervals, enabling rapid epidemic modeling and policy decision-making on a global scale~\cite{dong2020interactive}. In remote sensing, NASA’s MODIS satellite products are updated daily, supporting timely environmental monitoring and disaster assessment~\cite{lewis2009remote}. In contrast, traditional datasets like ImageNet~\cite{deng2009imagenet} and MNIST have not been updated for years, making them suitable for algorithm benchmarking but less relevant for contemporary applications. Meanwhile, open knowledge bases like Wikidata allow real-time user editing and provide API-based updates, representing a higher level of ``interactive timeliness''~\cite{vrandevcic2014wikidata}. Timeliness can be systematically quantified using indicators such as collection-to-release time lag, average update interval, event response delay, and timestamp consistency~\cite{redman1998impact,batini2006data}.

\subsubsection{Traceability} 
\label{sec:background_quality_trace}

Data traceability refers to the ability to track the complete journey of data from its origin and transformations to its final use. 
Traceability has increasingly become a critical supplementary metric for evaluating scientific data security and trustworthiness, especially in the context of open science and data reuse. Highly traceable data should include complete metadata, change logs, version control records, and accountability information, meeting the ``Findability'' and ``Reusability'' criteria of the FAIR principles~\cite{wilkinson2016fair}. For example, each record on the OpenAIRE platform~\cite{tool-openaire} includes a unique DOI, data acquisition description, and license details, significantly enhancing verifiability and reuse credibility. 
Moderately traceable data may provide basic metadata but often lack processing chains, revision histories, or algorithmic documentation, limiting users' ability to assess reliability. Low-traceability data typically lack source documentation and coherent annotation, rendering them difficult to verify. For instance, web-scraped research images or code snippets without provenance or revision records pose considerable risks in academic usage~\cite{krotov2020tutorial}. Recently, technologies such as blockchain and cryptographic hash signatures are being explored to build traceability chains and verifiable records for scientific data~\cite{liang2017provchain}.

\subsection{Dimensions for Evaluating Scientific AI}
\label{sec:background_eval}

General-purpose LLM benchmarks primarily assess core natural language processing and general reasoning abilities. Key evaluation dimensions typically include
language understanding, fluency, factual knowledge recall,
reasoning and problem-solving.
These benchmarks are designed to evaluate broad linguistic competence and general cognitive skills across everyday or non-specialized domains. Even when covering technical subjects (\eg, STEM topics in MMLU), they often assume only basic computational skills and high school–level science knowledge. Evaluations of factuality and alignment are typically grounded in general content.
In contrast, science-focused LLM benchmarks require mastery of the depth, precision, and rigor characteristic of academic research. Beyond the general dimensions listed above, scientific LLMs must be evaluated on their ability to engage with domain-specific scientific knowledge, reason with formal systems (\eg, equations, symbolic logic), retrieve and synthesize scholarly information, and support hypothesis generation or experimental design.

\subsubsection{Expert-Level Scientific Knowledge Comprehension and Retrieval} 
Unlike general-purpose language models, scientific AI models must retrieve, comprehend, and apply cutting-edge research knowledge across diverse scientific disciplines with domain-level expertise. This knowledge extends beyond general encyclopedic facts to include domain-specific equations, physical constants, technical terminology, and theoretical constructs. A model’s ability to access, interpret, and reason over external academic knowledge is a critical dimension of evaluation, serving as a cornerstone for enabling automated scientific discovery.
Key evaluation aspects include information retrieval, literature-based fact verification, and the integration of heterogeneous scientific knowledge. For example, SciBench~\cite{scibench} introduces benchmark tasks requiring the retrieval of mathematical equations, chemical laws, and physical theorems; SciKnowEval~\cite{sciknoweval} spans domains from biology to materials science, assessing tasks such as molecule identification and reaction prediction; and SciQA~\cite{auer2023sciqa} leverages the Open Research Knowledge Graph to support complex cross-domain scientific questions. This dimension challenges models on both the breadth and depth of scientific understanding, emphasizing accuracy, completeness, and the ability to engage with knowledge beyond surface-level facts.

\subsubsection{Scientific Reasoning and Problem Solving}
Scientific problems often require multi-step reasoning rooted in the principles of the scientific method. Effective models must be capable of formulating and decomposing complex problems, applying relevant scientific laws and theories, and performing precise numerical computations. 
SFE~\cite{zhou2025scientists}, for example, emphasizes advanced reasoning skills of Sci-LLMs, including the evaluation of scientific attribute understanding and comparative analysis.
Error analyses of science-focused benchmarks reveal that key reasoning capabilities include logical decomposition, causal inference, deductive problem solving, and abstract reasoning. These tasks extend beyond the scope of general mathematical puzzles found in standard LLM benchmarks, demanding the ability to reason about experimental procedures, derive theoretical formulas, and interpret results within a scientific framework.

\subsubsection{Multimodal Scientific Data}
Science AI models should incorporate various modalities other than language. The ability to understand data diagrams, including figures and tables, and to conduct quantitative and statistical analysis to identify scientific trends, is crucial. Furthermore, expert AI models need to comprehend specialized scientific data that requires domain-specific knowledge, such as chemical structures and laboratory images, for high-level reasoning. SciBench~\cite{scibench} notably includes a multimodal subset with figures and graphs, highlighting that assessing the ability to interpret visual scientific information is a dimension beyond typical LLMs and even MLLMs. On the other hand, it remains to be seen whether current science AI models can fully incorporate and leverage all these diverse data types effectively for truly advanced scientific discovery.

\begin{figure*}[t!]
    \centering
    \includegraphics[width=0.85\linewidth]{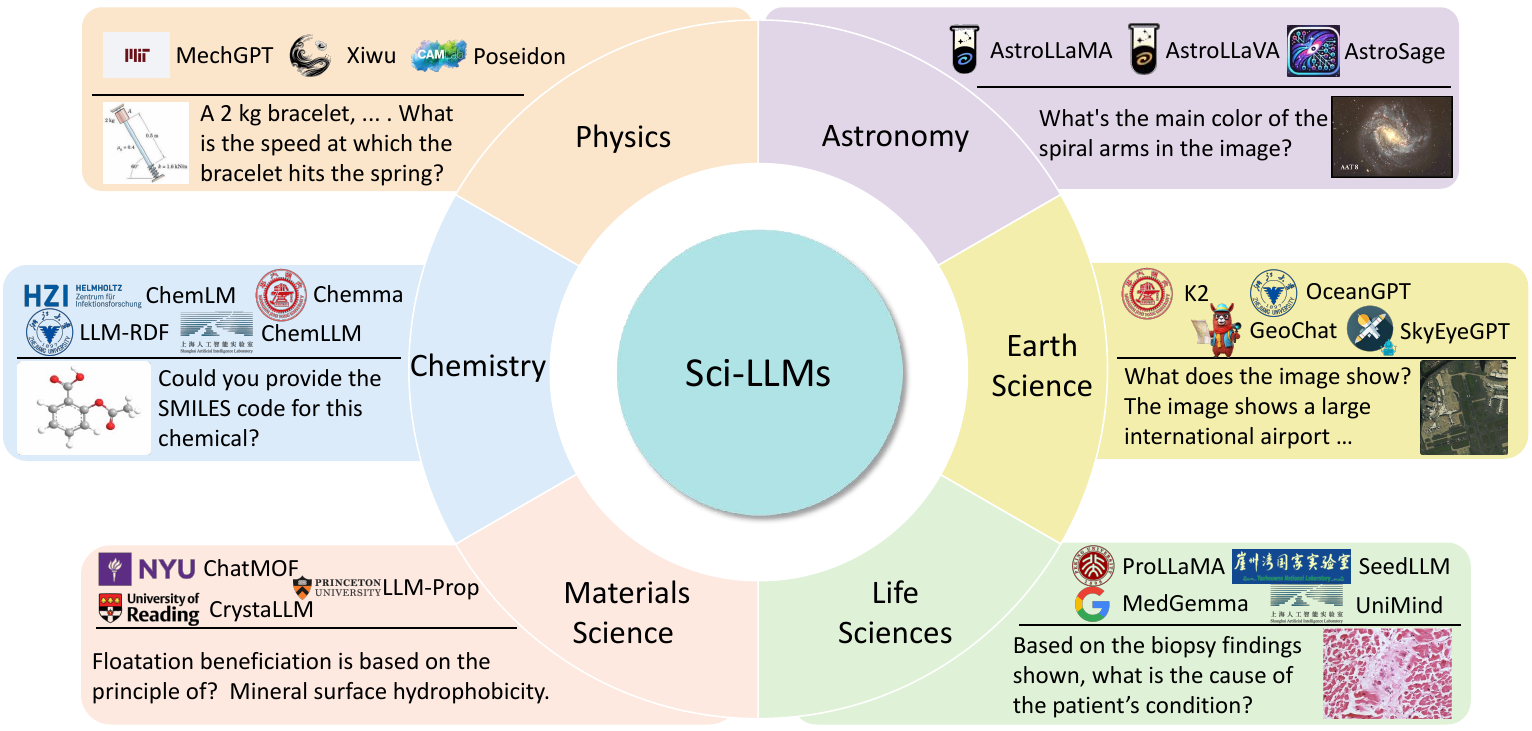}
    \caption{Research scopes of Sci-LLMs across six scientific subjects: physics, chemistry, materials science, life sciences, Earth science, and astronomy. For each subject, we present representative domain-specific Sci-LLMs and example questions that the Sci-LLMs are able to solve.
    }
    \label{fig:llm_scopes}
\end{figure*}
\section{Scientific Large Language Models}
\label{sec:scillms}

Sci-LLMs are emerging as powerful tools for modeling, understanding, and reasoning across diverse scientific domains. This section begins with a brief touch on the architecture and training of general LLMs, establishing the groundwork for their scientific extensions (Sec.~\ref{sec:scillms_intro}), followed by a survey of general-purpose Sci-LLMs (Sec.~\ref{sec:scillms_general}). We then introduce major scientific LLMs across six natural science domains (Sec.~\ref{sec:scillms_specific}), including physics (Sec.~\ref{sec:scillms_specific_physics}), chemistry (Sec.~\ref{sec:scillms_specific_chemistry}), materials science (Sec.~\ref{sec:scillms_specific_materials}), life sciences (Sec.~\ref{sec:scillms_specific_life}), astronomy (Sec.~\ref{sec:scillms_specific_astronomy}), and Earth science (Sec.~\ref{sec:scillms_specific_earth}), each with unique data modalities, modeling challenges, and scientific applications. 
Fig.~\ref{fig:llm_scopes} illustrates the research scope of Sci-LLMs covered in this survey. 

\subsection{Introduction of Large Language Models}
\label{sec:scillms_intro}

LLMs~\cite{brown2020language,bai2023qwen,grattafiori2024llama} exhibit strong capabilities in understanding, generating, and interacting with human language.
LLMs can comprehend the intricate relationships among massive amounts of text, sequential, and visual data in queries, and generate corresponding answers following user instructions.
Existing LLMs are mainly based on a decoder-only transformer architecture~\cite{vaswani2017attention}, which converts human natural language into a sequence of textual tokens. 
When equipped with specific modality encoders, data from other modalities, such as images or videos, can also be converted into tokens and processed by LLMs. 
Then, LLMs generate or expand information when given an input or condition by extracting relationships between tokens. 
The generated tokens are then decoded into text or other modalities that humans can understand.
To enable such capability, LLMs are usually pre-trained on vast and diverse data using the next-token prediction objective~\cite{brown2020language}.
This process encodes world knowledge into the LLMs and serves as the foundation for their capabilities.
The post-training process is crucial for activating and enhancing the task-specific knowledge that LLMs acquire from the large-scale data, allowing them to understand user instructions and solve complex tasks in practical applications.

\subsection{General-purpose Sci-LLMs}
\label{sec:scillms_general}

\begin{figure*}[t!]
  \centering
    \includegraphics[width=0.95\linewidth]{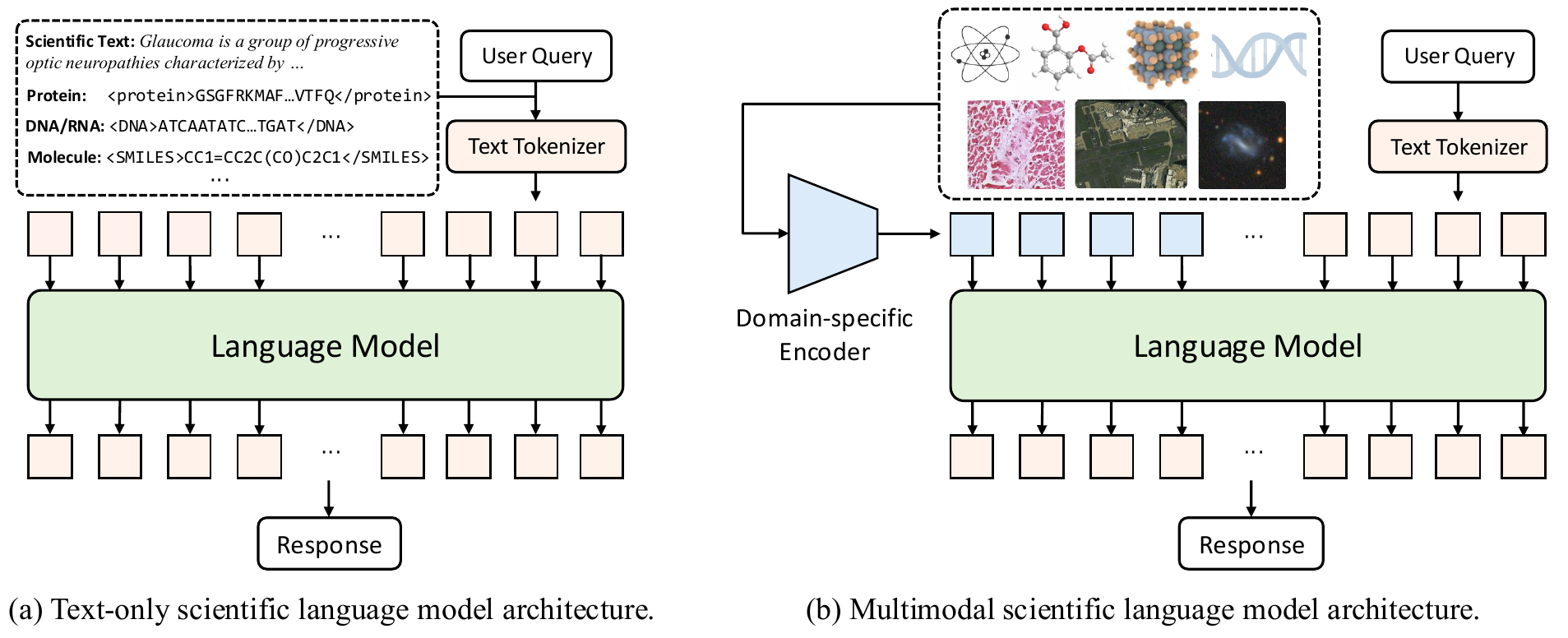}
  \caption{
  Illustration of common model architectures for existing scientific large language models.
  \textbf{(a) Left:} Text-only language model architecture showing the processing pipeline where user queries are processed through a text tokenizer, with scientific text inputs (including disease descriptions, DNA/RNA sequences, protein sequences, and SMILES molecular representations) as part of the query, to generate responses.
  \textbf{(b) Right:} Multimodal model architecture featuring a domain-specific encoder that processes diverse scientific data types (molecular structures, DNA structures, microscopic images, \etc) alongside text inputs, enabling comprehensive scientific question-answering capabilities through the integration of textual and non-textual scientific information.
    }
  \label{fig:llm_model_arch}
\end{figure*}

Current scientific LLMs are mainly developed from existing general-purpose LLMs through post pre-training or fine-tuning on data from specific scientific tasks~\cite{eger2025transforming}.
They do not alter the model architecture of existing LLMs. Instead, domain-specific encoders are used to convert scientific data, such as medical images and protein sequences, into tokens compatible with the LLM backbones.
Fig.~\ref{fig:llm_model_arch} demonstrates the architecture of LLMs in the scientific domain.
They can achieve significant performance improvement on certain scientific tasks, but they cannot push the capability boundary of existing LLMs due to the limited data scale and task diversity.
For example, \emph{DARWIN} models~\cite{xie2023darwin} are fine-tuned on the open-source LLaMA-7B~\cite{touvron2023llama} using about 60 K science-focused instruction examples covering physics, chemistry, and materials science. These instructions are carefully collected from science exams and scholarly papers.
\emph{SciGLM}~\cite{zhang2024sciglm} is further fine-tuned from a general-purpose LLM with the proposed SciInstruct dataset, which enhances the model’s ability to understand intricate scientific concepts, derive symbolic equations, and solve numerical problems. SciInstruct is built through self-reflective annotation to alleviate data scarcity in science domains such as Physics and Chemistry.

However, directly fine-tuning open-source LLMs does not significantly enhance their scientific capabilities because of a limited training corpus. Therefore, to improve performance on scientific tasks, a model should be pre-trained on large-scale scientific data, thereby strengthening its capabilities for scientific tasks.
For example, \emph{Galactica}~\cite{taylor2022galactica} is a 120 B parameter decoder-only model trained on 106 B tokens drawn from papers, reference materials, encyclopedias, and other scientific sources. Its corpus mixes text with scientific sequence representations such as protein sequences and chemical formulae, as well as LaTeX and code. At release, Galactica reported state-of-the-art results on PubMedQA\cite{jin2019pubmedqa} and MedMCQA-dev\cite{pal2022medmcqa} and strong performance on mathematical reasoning and technical knowledge probes, as well as chemistry, biology and physics capabilities

\emph{SciDFM}~\cite{sun2024scidfm} adopts a MoE~\cite{shazeer2017moe} architecture with 5.6 B active parameters routed across eight experts. It is pre-trained from scratch on 300 B science-domain tokens covering mathematics, chemistry, biology, geography, and general science, together with 270 B general-domain tokens. This broad training corpus strengthens the model’s scientific capabilities. SciDFM is then fine-tuned on customized instruction-tuning data derived from open-source datasets to improve its performance on downstream scientific benchmarks.
\emph{OmniScience}~\cite{prabhakar2025omniscience} is built on the LLaMA-3.1-70B model and undergoes domain-adaptive pre-training on a carefully curated corpus of papers, journals, and textbooks that span general science and electrochemistry. The model is then instruction-tuned to improve its understanding of science-specific task prompts. Finally, OmniScience distills knowledge from the advanced reasoning model DeepSeek-R1 by fine-tuning on the s1K-1.1 dataset~\cite{muennighoff2025s1}, thereby gaining multi-step reasoning capability for complex scientific problems. Intern-S1~\cite{bai2025interns1scientificmultimodalfoundation} is a recently released open-source scientific multimodal foundation model developed by Shanghai AI Laboratory. It adopts a MoE architecture with 241B total parameters and 28B activated per inference step. The language backbone is based on Qwen3-235B MoE and is extended with specialized encoders, including InternViT-6B for vision and a time-series signal encoder, together with a dynamic tokenizer designed for scientific formats such as SMILES and FASTA. Trained on over 5 trillion tokens, including 2.5T from scientific domains, Intern-S1 delivers competitive performance on general reasoning tasks while surpassing both open- and closed-source systems across multiple scientific benchmarks, such as molecular synthesis planning, materials property prediction, and crystal stability.

Recently, beyond large-scale pre-training, existing general-purpose LLMs~\cite{o1} propose test-time scaling by introducing Chain-of-Thought reasoning process~\cite{wei2022chain}.
Such a paradigm demonstrates potential in solving complex scientific tasks, which require reasoning from multiple perspectives and drawing accurate and interpretable conclusions~\cite{platt1964strong}.
To achieve this, recent work tries different approaches.
For example, \emph{DeepSeek-R1}~\cite{guo2025deepseek}, built upon DeepSeek-V3-Base through cold-start training and reasoning-oriented Reinforcement Learning (RL) processes, achieves results comparable to OpenAI-o1 model~\cite{o1} on scientific benchmarks such as MMLU-Pro~\cite{wang2024mmlu}, and GPQA-Diamond~\cite{Rein2023GPQAa}.
\emph{Qwen3}~\cite{yang2025qwen3} draws inspiration from DeepSeek-R1 and is designed with a MoE architecture. It integrates both thinking and non-thinking modes into a unified framework, allowing the model to respond adaptively based on task difficulty to avoid unnecessary computational overhead compared to DeepSeek-R1. 
\emph{Kimi K2}~\cite{team2025kimi} scales the MoE architecture to 1 trillion parameters with 32 billion activated parameters.
K2 undergoes a multi-stage post-training process which is powered by the proposed large-scale agentic data. 
Therefore, K2 can interact with real and synthetic environments using diverse tools, demonstrating its potential for addressing complex scientific tasks.
\emph{Gemini 2.5 Pro}~\cite{comanici2025gemini} is a reasoning model that can process multimodal and long-contextual data. 
It can handle text, image, video, and audio inputs with a total context length of 1 million tokens. 
This capability makes it well-suited for complex scientific tasks that involve processing sensor or sequence data from different devices or databases.
\emph{Grok 4}~\cite{grok4} is trained with large-scale RL at pre-training scale, achieving 50.7\% accuracy on the Humanity’s Last Exam (HLE)~\cite{phan2025humanity} and demonstrating strong scientific reasoning capabilities.

The test-time scaling strategy demonstrates strong potential for enhancing generalization in scientific tasks by understanding multimodal scientific data and reasoning with scientific tools.
In the future, further scaling up the RL process could lead to frontier intelligence surpassing human capabilities, enabling novel scientific discoveries and allowing models to design and conduct experiments using real-world tools in support of the proposed hypotheses.
Moreover, developing a virtual laboratory environment~\cite{swanson2025virtual} where LLMs can conduct experiments and collect experimental feedback would accelerate the training process toward more powerful general-purpose scientific intelligence.


\begin{figure*}[t!]
    \centering
    \includegraphics[width=\linewidth]{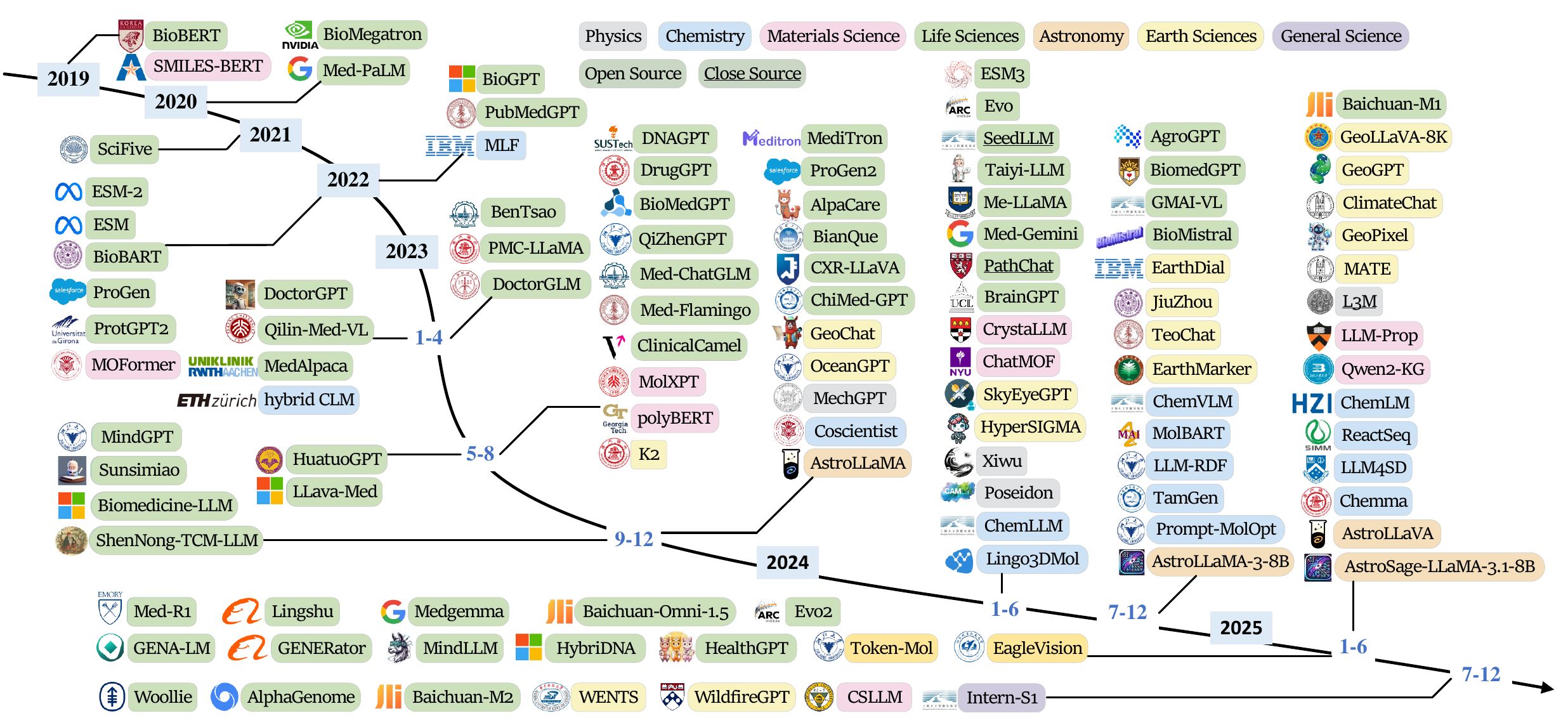}
    \caption{Chronological overview of notable Sci-LLMs categorized by six scientific domains, spanning from 2019 through early 2025. Due to the rapid expansion of the field, this figure presents a selective overview. For detailed information, please refer to Tab.~\ref{tab:sci_llms}.}
    \label{fig:llm_development} 
\end{figure*}

\subsection{Domain-specific Sci-LLMs}
\label{sec:scillms_specific}

In some cases, domain-specific scientific LLMs can be more helpful for particular scientific tasks. Such models can be constructed with well-curated, domain-specific datasets and training schemes tailored to the target subject. Below, we introduce recent domain-specific scientific LLMs that cover eight subjects. Fig.~\ref{fig:llm_development} demonstrates the development of scientific LLMs across six subjects.

\subsubsection{Physics}
\label{sec:scillms_specific_physics}

In the field of physics, scientific LLMs have begun to take a significantly different path compared to traditional symbolic modeling and numerical simulation. By integrating LLMs with physics engines and visual modules, these models are not only capable of processing natural language descriptions of physical systems but also able to explicitly estimate physical parameters, simulate dynamic evolution, and represent physical laws symbolically. They are evolving from language understanding systems into intelligent tools that interact directly with scientific workflows.

\emph{LLMPhy}~\cite{cherian2024llmphy} is a representative model that combines program synthesis with physics simulation feedback. Its core idea is to use an LLM to generate symbolic code that can be executed by a non-differentiable physics engine, enabling iterative refinement of physical parameters like friction, stiffness, damping, and rotational inertia. In Phase 1, the system uses trajectories extracted from auxiliary videos, while in Phase 2, it processes multi-view images with a vision-language model to reconstruct the scene layout. The TraySim dataset, which provides paired multi-view images and video trajectories, supports a closed-loop “analysis-by-synthesis” framework that allows the model to align simulation results with physical reasoning.
\emph{POSEIDON}~\cite{herde2024poseidon} is a model designed for learning solution operators of partial differential equations (PDEs). While it does not use natural language as input, its core is a scalable Operator Transformer with strong temporal modeling capabilities enabled by time-conditioned layer normalization. The architecture uses a multi-scale Vision Transformer~\cite{liu2022swinv2} to encode spatial fields, and a U-Net-style module to encode input into latent space. The transformer backbone, along with time-conditioned layer normalization, enables continuous-in-time evaluations, with optional autoregressive rollouts during inference. It is trained in two stages: large-scale pretraining using an ``all-to-all'' strategy on PDE trajectories  (\eg, Euler, Navier-Stokes), and small-sample fine-tuning on specific PDE tasks. POSEIDON achieves high sample efficiency, requiring only 20 samples to match the performance of the widely-used FNO that needs 1024 samples.
\emph{Xiwu}~\cite{zhang2024xiwu} is a domain-specific model for high energy physics (HEP), built on Vicuna-13B-v1.5. Its system includes a data engine, LLM core, vector memory, and user-facing interfaces. It is trained in three stages: continued pretraining on 750M HEP-specific textual tokens, supervised fine-tuning on 26k human-verified QA pairs, and real-time learning via a just-in-time system where expert users can inject, correct, or update knowledge in a vector storage for retrieval. Xiwu outperforms Vicuna-13B in 95\% of win-or-draw rate and surpasses GPT-4 in most tasks involving HEP software code generation for BESIII Offline Software System(BOSS) ~\cite{zou2024offline}, demonstrating its domain specialization.

In summary, these models reflect the diverse design and training strategies adapted for physics tasks. They range from symbol-to-simulation systems and PDE operator learners, to physics QA transformers and high-energy physics retrievers. As they evolve, further integration of multimodal capabilities, improved spatiotemporal reasoning, and unified knowledge representation frameworks will be essential for expanding their scientific utility.

\subsubsection{Chemistry}
\label{sec:scillms_specific_chemistry}

In this section, we review the latest advances in LLMs in the field of chemistry. We begin by examining their model architectures, training strategies, and the core data modalities employed, such as molecular structures, reaction data, spectroscopic information, and scientific literature. Next, we provide a comprehensive, domain-specific overview of their applications across key areas in chemistry, including molecular design, reaction prediction, retrosynthetic analysis, catalyst discovery, quantum chemistry, and materials science. Finally, we critically discuss the major challenges and ethical considerations in the field, and offer a perspective on future research directions and opportunities for AI-driven chemical innovation. 

\emph{ChemLLM}~\cite{zhang2024chemllm} is one of the earliest LLMs that is specifically designed for chemistry. It also curates \textit{ChemData}, a specialized instruction-tuning dataset, and \textit{ChemBench}, a comprehensive benchmark covering nine core chemistry tasks. 
\emph{InstructMol}~\cite{cao2023instructmol} aligned molecular structure and text via using a light-weighted projector, following LLaVA's alignment strategy. It leverages a two-stage training scheme which starts with the multimodal alignment object followed task-specific instruction tuning. InstructMol supports several tasks, including molecule property prediction, molecule description generation, retrosynthesis prediction, \etc 
\emph{ChemDFM}~\cite{zhao2024chemdfm} is pre-trained on 34 billion tokens from chemical literature and textbooks, and fine-tuned with 2.7 million instruction pairs. As a result, ChemDFM is capable of understanding and reasoning over chemical knowledge through natural, free-form dialogue. 
\emph{ChemMLLM}~\cite{tan2025chemmllm} is proposed to mitigate the gap in generating molecular images, establishing a unified MLLM for chemical understanding and generation across text, molecular SMILES string, and molecular images. 
\emph{Chem3DLLM}~\cite{jiang2025chem3dllm} addresses the inability of traditional large language models to generate accurate 3D molecular conformations due to incompatible formats, lack of multimodal alignment, and absence of chemical priors. It introduces a reversible text encoding of 3D structures, enabling lossless compression and integration within a language‑model token space. A protein‑embedding projector aligns protein pocket representations, while reinforcement learning with chemical validity rewards enforces physical plausibility, yielding state‑of‑the‑art results in structure‑based drug design.


\subsubsection{Materials Science}
\label{sec:scillms_specific_materials}

LLMs have also been widely explored in diverse tasks in materials science. 
Recent studies have applied a transformer-based encoder to learn material representations.
For example, \emph{SMILES-BERT}~\cite{smiles-bert} is pre-trained on a vast collection of SMILES corpora to learn molecular representations for property prediction.  
Similarly, \emph{polyBERT}~\cite{polybert} employs a DeBERTa-based encoder~\cite{he2020deberta} trained on about 100 million hypothetical polymer SMILES, enabling end-to-end polymer fingerprinting.  
\emph{MatBERT-bandgap}~\cite{matbert} was pre-trained on about 2 million materials science abstracts, learning latent compositional features.  
\emph{Regression Transformer}~\cite{regtrans} adopts a novel multi-task scheme, translating property regression into sequence outputs by tokenizing continuous values and alternating masked-language and regression training phases.
Regression Transformer can simultaneously predict numeric properties and generate molecular strings, effectively merging regression and generation tasks. 

Some work applies general-purpose LLMs by utilizing Retrieval-Augmented Generation (RAG) equipped with professional databases to solve related tasks.
Knowledge integration is another strength. 
\emph{Qwen2-KG}~\cite{qwen2kg} uses Qwen2-72B together with a retrieved materials knowledge graph to answer questions about framework materials.  By combining chain-of-thought retrieval with graph facts, it achieves about 91.7\% accuracy on a held-out QA benchmark, outperforming LLM-only baselines and providing cited sources. 

Recent work has investigated fine-tuning a ChatGPT-style model (\ie, decoder-only transformer) to materials science.
For example, \emph{MolXPT}~\cite{molxpt} is built on GPT-2~\cite{radford2019language} and learns from combined PubMed abstracts and molecular SMILES data, while \emph{GPT-MolBERTa}~\cite{gptmolberta} is fine-tuned from a BERT-like encoder on about 326K molecular descriptions synthesized by ChatGPT. 
In the molecular domain, \emph{MolGPT}~\cite{Bagal2022} uses a GPT-style causal LM objective: it is pre-trained on millions of SMILES strings and then fine-tuned for conditional generation (scaffold or property guidance).
\emph{XYZTransformer}~\cite{xyztrans} is designed to process molecular structural data and is a decoder-only transformer trained directly on the raw 3D coordinates of molecules. 
\emph{CrystaLLM}~\cite{crystalm} is fine-tuned from the LLaMA-2 model using text-formatted crystal structures, harnessing billions of parameters to capture atomistic symmetries. 
CrystaLLM can generate metastable materials with a frequency of about 49\% when given desired features, and significantly outperforms diffusion-based models.
In synthesis planning, Okabe \etal develop three LLMs (\emph{LHS2RHS}, \emph{RHS2LHS}, \emph{TGT2CEQ})~\cite{lhs2rhs} to predict chemical equations: given reactants they predict products (LHS→RHS) or vice versa, and they can generate balanced chemical equations for target compounds. Fine-tuning on text-mined inorganic syntheses raised reaction-prediction accuracy to about 90\%, enabling rapid synthesis route inference.
\emph{CSLLM}~\cite{csllm} is fine-tuned from LLaMA-3-8B to predict the synthesizability and precursors of crystal structures.  
CSLLM reaches approximately 98.6\% accuracy on the synthesizability prediction task, which is vastly higher than that of Density Functional Theory (DFT)-based filters, for identifying experimentally realized crystals.
CSLLM can also predict the likely precursors and synthesis methods (solid vs solution), illustrating how LLMs can capture complex experimental domain knowledge. 
\emph{MechGPT}~\cite{buehler2024mechgpt} is tailored for material mechanics and multiscale modeling. It is built on the LLaMA2 model and fine-tuned using LoRA techniques with domain-specific question-answering (QA) data. Although its current inputs are text-based, the model is designed to eventually incorporate image and structural modalities. Its demonstrated capabilities include knowledge retrieval, hypothesis generation, and the construction of interpretable ontological knowledge graphs for structural insight. While MechGPT’s input is currently text-only, its architecture is designed to accommodate future multimodal extensions.

\subsubsection{Life Sciences} 
\label{sec:scillms_specific_life}

LLMs, pre-trained on large-scale scientific data in the field of life sciences, can adapt to a wide spectrum of downstream tasks, ranging from generating accurate diagnostic reports to designing previously unknown protein structures or novel drugs~\cite{zhang2025scientific}.
These tasks are closely related to human health.
In this part, we review the development of LLMs in the field of life sciences, including model architectures, training schemes, and applications. 



\textbf{Multi-Omics.}
DNA, RNA, and protein sequences have been seen as the ``language of life'' in computational biology in recent years~\cite{tan2009dna}.Recent advances in multi-omics research have developed two complementary families of domain-specific language models: \textit{(i)} encoder-centric genomics/protein language models (GLMs/PLMs) that are trained from scratch on biological sequences to learn molecular representations and biological constraints like the EVO series~\cite{nguyen2024sequence, brixi2025genome} and ESM series~\cite{rives2021biological,esm2, hayes2025esm3}; and \textit{(ii)} LLM-augmented systems that integrate omics data into instruction-following text LLMs to generate natural-language outputs, typically leveraging models from category (i) as omics encoders.

For category (i), \emph{EVO}~\cite{nguyen2024sequence} represents a groundbreaking advancement in genomic foundation modeling, being trained on an extensive dataset comprising over 80,000 bacterial and archaeal genomes, as well as millions of predicted phage and plasmid sequences, totaling 300 billion nucleotide tokens. This model establishes scaling laws for DNA that complement those discovered in language and vision domains. Additionally, EVO seamlessly integrates across DNA, RNA, and proteins, achieving zero-shot function prediction that rivals specialized language models. \emph{EVO2}~\cite{brixi2025genome} scales training to 9.3 trillion DNA bases spanning all domains of life and extends context to genome scale (up to 1M tokens), accurately predicting functional impacts of genetic variation and supporting genome-level design. Focusing on proteomics foundation models, the methodological landscape is evolving from unconditional sequence modeling toward a closed loop of controllable generation—cross-modal semantic alignment—interactive reasoning: at the outset, 
\emph{ESM-1b}~\cite{rives2021biological} demonstrates that large-scale unsupervised modeling of 250M protein sequences learns representations with emergent structure/function information enabling accurate long-range contact prediction and remote-homology organization. \emph{MSA Transformer}~\cite{rao2021msa} applies axial attention directly to multiple-sequence alignments to capture coevolutionary dependencies, yielding unsupervised structural features and strong contact-prediction signals. \emph{ESM-1v}~\cite{meier2021zeroshot} introduces a protein LM whose zero-shot likelihoods match state-of-the-art supervised predictors on deep mutational scanning benchmarks for functional effect prediction. \emph{ESM-IF1}~\cite{hsu2022if1} performs inverse folding by training on millions of predicted backbones, achieving 51\% native sequence recovery (72\% for buried residues) on held-out structures and generalizing to complex design settings. \emph{ESM-2/ESMFold}~\cite{lin2023esmfold} enables direct single-sequence, atomic-level 3D structure prediction without MSAs, delivering strong accuracy at substantially higher speed than traditional MSA-based pipelines. \emph{ESM3}~\cite{hayes2025esm3} unifies sequence, structure, and function in a single multimodal generative model that reasons across modalities and designs novel, functional proteins far from known families. \emph{ProtGPT2}~\cite{ferruz2022protgpt2} learns the “grammar” of proteins via large-scale unsupervised training, enabling \emph{de novo} generation close to natural sequence statistics; on controllability and scale, \emph{ProGen}~\cite{madani2020progen} injects functional and localization conditions into autoregressive modeling, and \emph{ProGen2}~\cite{progen2} further expands parameters and corpora to improve generalization and fitness prediction; along the path from general LLMs to domain adaptation. \emph{Ankh}~\cite{ankh} attains strong baselines under lower compute through protein-oriented training and architectural optimizations; centering on text-driven design,

As instances of (ii), \emph{LLaMA-Gene}~\cite{llama-gene} adapts LLaMA2-7B via domain-adaptive continued pretraining on a 39.5B-token DNA corpus from reference genomes, followed by instruction tuning with 800K synthetic multi-omics QA pairs (variant interpretation, promoter prediction, transcript identification). \emph{ProLLaMA}~\cite{prollama} migrates general models to protein multi-tasking via vocabulary pruning and instruction alignment. \emph{ProteinDT}~\cite{proteindt}, \emph{PAAG}~\cite{paag}, and \emph{Pinal}~\cite{pinal} map natural-language intent to controllable sequences—respectively via text-protein alignment, annotation-sequence multi-level domain alignment, and a two-stage pipeline—and support sequence editing; for \emph{interactive analysis}, \emph{ProteinGPT}~\cite{proteingpt} and \emph{ProteinChat}~\cite{proteinchat}/\emph{ProtChatGPT}~\cite{protchatgpt} align sequence/structure representations with LLMs to enable multi-turn QA around function and structure; in \emph{cross-modal translation}, \emph{ProTranslator}~\cite{protranslator} and \emph{BioTranslator}~\cite{biotranslator} achieve zero-shot transfer between text and protein/biological data (text $\leftrightarrow$ protein/data); to enhance \emph{interpretability}, \emph{Prot2Text}~\cite{prot2text} directly generates free-text functional descriptions from sequences. \emph{Evolla}~\cite{evolla} is a protein-language generative model with 80 billion parameters, designed to decode the molecular language of proteins by integrating sequence and structural information on proteins, together with user-query information. This capability is enabled by the proposed 546 million protein-centric question-answer pairs. Taken together, this line of work progresses from “learning the protein grammar” to “conditional controllable generation” and onward to “cross-modal alignment and dialogue-centric agents,” converging toward a design-analysis-feedback loop for proteomics foundation models.

Recent work proposes to understand DNA, RNA, and protein sequences simultaneously.
\emph{NatureLM}~\cite{naturelm} presents a general Sci-LLM instruction-tuned across genomics, proteomics, biochemistry, and materials science. Its post-training data comprises over 1.1M instruction pairs generated from curated databases such as UniProt, Ensembl, and GenBank, spanning protein functions, gene regulatory elements, and variant effects, formatted in English QA and reasoning chains. 
\emph{ChatNT}~\cite{chatnt} further establishes a multi-task conversational agent trained on curated instruction datasets across DNA, RNA, and protein domains. It integrates 361M English and DNA tokens from 18 task categories (\eg, methylation, splicing, polyadenylation, protein melting), and uses a unified text-to-text objective with an English-aware Perceiver projection to align genomic sequences with natural language prompts. Collectively, these models highlight the shift toward cross-omics instruction tuning that enables unified biological reasoning across diverse molecular inputs.

\textbf{Molecular and Cellular Biology. } Some studies propose applying LLMs in the field of molecular and cellular biology. These LLMs focus on understanding the morphology and function of cells in living organisms.

For example, \emph{MolecularGPT}~\cite{liu2024moleculargpt} is an instruction-tuned large language model (LLaMA-2-7B–based) for molecular property prediction that operates on SMILES strings, enabling zero or few-shot inference across diverse biological molecules. It is obtained by QLoRA fine-tuning on a hybrid instruction set spanning over 1,000 property tasks compiled from sources such as ChEMBL bioassays, CHEMBL physico-chemical properties, and QM9 (HOMO/LUMO), with about 3.5 GB of training tokens.
\emph{scGPT}~\cite{cui2024scgpt} is transformer-based single-cell foundation model with a specialized masked-attention scheme that jointly learns gene and cell embeddings to support cell-type annotation, batch correction, perturbation-response prediction, and gene network inference. It is pretrained in a self-supervised manner on over 33 million normal human cells from the CELLxGENE atlas and then adapted via task-specific fine-tuning pipelines for diverse downstream single-cell applications.

LLMs have also emerged as powerful tools for de novo molecular design. By treating chemical structures as ``languages'' (\eg, using SMILES notation), models like ChemBERTa\cite{chithrananda2020chemberta} and MolBERT\cite{li2021mol} generate novel molecules with desired properties. For instance, Edwards \etal~\cite{edwards2022translation} fine-tuned GPT-3 on chemical datasets to produce drug-like molecules, achieving hit rates comparable to high-throughput screening. In drug discovery, LLMs accelerate lead optimization. They predict bioactivity by analyzing sequence data from proteins and ligands. A notable example is the integration of LLMs with reinforcement learning in models like Chemformer, which designs molecules for specific targets, such as COVID-19 inhibitors~\cite{hatakeyama2023prompt}. These approaches reduce synthesis trials by 50-70\%, as validated in virtual screening benchmarks.
\textbf{Healthcare and Medical Science. } Recent LLMs in the field of healthcare and medical science are primarily adapted from existing general-purpose LLMs~\cite{jiang2025omniv}. 
These models are typically further pre-trained on domain-specific corpora, such as clinical reports, medical literature, and imaging data. They are then fine-tuned with medical instruction-response pairs to serve diverse user groups, including doctors, students, and patients.

Due to computational costs, recent medical LLMs only perform supervised fine-tuning (SFT) on general LLMs using medical-related instruction data. This process introduces the capability to solve medical tasks to general LLMs.
For example, \emph{BioMistral}~\cite{biomistral}, \emph{BioMedLM}~\cite{biomedlm}, \emph{ClinicalCamel}~\cite{toma2023clinical}, and \emph{MedAlpaca}~\cite{han2023medalpaca} collect medical question-answering pairs and doctor-patient dialogue data, and perform SFT on open-source LLMs such as LLaMA, achieving performance improvements on several medical benchmarks, such as MedMCQA~\cite{pal2022medmcqa}, PubMedQA~\cite{jin2019pubmedqa} and MedQA~\cite{jin2020disease}.
\emph{Med-PaLM} series~\cite{medpalm} are developed from a 540B parameter LLM, PaLM, are directly instruction-tuned on PaLM, and using a combination of prompt engineering technologies~\cite{10.1145/3560815} to adpat to medical quesiton-answering tasks. 
\emph{Apollo}~\cite{apollo} is a lightweight multilingual medical LLM, which collects medical data covering the six most widely spoken languages. Such a lightweight model can be deployed in hospitals to help protect the privacy of medical data.
\emph{HuatuoGPT}~\cite{huatuogpt} is fine-tuned from general LLMs using medical instruction and conversation data from both real-world sources and ChatGPT, in order to introduce medical-specific skills and to distill capabilities from powerful general LLMs.

Only performing SFT on existing general LLMs cannot further improve model performance in the healthcare field. Further scaling up the pre-training scale is beneficial to model performance.
For example, \emph{PMC-LLaMA}~\cite{pmcLLaMA} is based on LLaMA and was pre-trained on data containing 4.8 million biomedical academic papers and 30,000 medical textbooks. \emph{HuatuoGPT-II}~\cite{chen2023huatuogptii} combines pre-training and instruction tuning, using over 5.2 million medical documents from encyclopedias, books, and web corpora, as well as 142,000 medical instructions. It is based on the Baichuan2-Base models.
\emph{CHIMED-GPT}~\cite{tian2023chimed} collects over 214 million multilingual tokens in Chinese and English from medical textbooks and encyclopedia data, and is pre-trained on Ziya-13-v2~\cite{fengshenbang}. This work also conducts RLHF~\cite{bai2022training} to further enhance the safety of the model’s responses.
\emph{Zhongjing}~\cite{yang2024zhongjing} also conducts complete training process including pre-training, instruction-tuning and RLHF. Besides, Zhongjing supports multi-turn dialogues to meet real-world diagnosis requirements.
\emph{Me-LLaMA}~\cite{meLLaMA} is continually pre-trained on LLaMA2 with 129 billion tokens from biomedical datasets, research papers, and clinical notes, and is then fine-tuned on 214,000 instruction tuning samples from clinical domains. \emph{Baichuan-M1}~\cite{wang2025baichuan} is trained from scratch and further scales up the pre-training process, using over 20 trillion tokens, which include both general data and medical-related data such as clinical information and patient records. Baichuan-M1 achieves significant performance across more than 17 medical-related benchmarks.

Clinical practice is inherently multimodal. The diagnostic process requires physicians to synthesize information from diverse sources, including the patient's verbal descriptions (text/audio), physical signs (visual), medical imaging (visual), and laboratory findings (structured data). Accordingly, MLLMs capable of processing multiple data modalities are considered a critical path forward in the evolution of medical AI~\cite{liu2024medcot}. 
Recent work investigates the use of MLLMs in the medical field, mainly focusing on two primary tasks: medical reports generation~\cite{reale2024vision} and medical Visual Question Answering (VQA)~\cite{lin2023medical, wang2025v2t}. \emph{LLaVA-Med}~\cite{llavamed}, as a pioneering work in this domain, successfully transferred the capabilities of a general-purpose MLLM, \ie, LLaVA, to the biomedical field. It is fine-tuned on the visual instruction-tuning data from PubMed papers and can understand medical images.
\emph{CXR-LLaVA}~\cite{cxrllava} and \emph{Radiology-LLaMA2}~\cite{liu2023radiologyllama2bestinclasslargelanguage} are specifically developed for chest X-ray (CXR) imaging. They utilize GPT-4 to extract impressions and findings from radiology reports in order to enhance their ability to interpret X-ray images, and they can generate reports in a clinical style.
\emph{Med-Flamingo}~\cite{medflamingo} is continually pre-trained on paired and interleaved medical image-text data from publications and textbooks, and can solve medical VQA tasks through few-shot learning without further fine-tuning on the VQA datasets.

Moreover, several works aim to extend the medical MLLMs capability to diverse medical tasks requiring more modality information and reasoning capabilities.
\emph{HuatuoGPT-Vision}~\cite{huatuogptvision} and \emph{GMAI-VL}~\cite{li2025gmaivlgmaivl55mlarge} collect large-scale medical multimodal data from PubMed papers and open-source medical image datasets. They are pre-trained on extensive medical image-caption pairs and further fine-tuned on data containing diverse instructions in the medical field. Therefore, they can solve a wide range of tasks from different departments.
\emph{MedGemma}~\cite{sellergren2025medgemma} further extends the in-context length of MLLMs and can process long-context data such as medical videos or patient electronic health records.
\emph{HuatuoGPT-o1}~\cite{chen2024huatuogpt} aims to introduce complex medical reasoning capability by fine-tuning the model on question-answer pairs with complex reasoning trajectories and conducting RL with verifier-based rewards to enhance complex reasoning. \emph{Medground-r1}~\cite{xu2025medground} leverages GRPO with spatial-semantic rewards to enhance medical image grounding without CoT annotations.
\emph{GMAI-VL-R1}~\cite{su2025gmai} introduces multimodal medical reasoning capability by directly applying RL to verifiable multiple-choice VQA data, thereby enhancing performance on medical image diagnosis and VQA tasks without collecting complex reasoning data.


\textbf{Agriculture.} In this section, we examine the emerging family of agricultural LLMs, covering their architectural choices, training strategies, and domain-specific capabilities.
\emph{SeedLLM}~\cite{yang2025seedllm} is a domain-specific large language model for seed science, built on Qwen2.5-7B. It is pre-trained on RiceCorpus (a bilingual corpus of 1.38 million agronomy papers) and GeneralCorpus~\cite{penedo2024fineweb,huang2025opencoder}, targeting terminology and knowledge from modern breeding research. The fine-tuning stage uses QAs from both general and agricultural domains, synthesized using GraphGen~\cite{chen2025graphgen}, a knowledge-graph-based generation framework. SeedLLM is evaluated on SeedBench, a multi-task benchmark co-designed with domain experts for seed breeding applications. The model remains closed-source.
\emph{PLLaMA}~\cite{yang2024pllama} is an open-source language model tailored for plant science. It extends LLaMA-2 with 7B and 13B parameter variants, and is continuously pre-trained on 1.5 million plant-related scholarly articles curated from the S2ORC corpus. Fine-tuning employs 1,030 instruction samples adapted from LIMA. The model is evaluated on a held-out plant science quiz set, showing strong comprehension of plant genetics, physiology, and breeding concepts.
\emph{AgroGPT}~\cite{awais2025agrogpt} is an open-source multimodal assistant for agronomic consultation, with 3B and 7B vision-language variants based on LLaVA. While no raw pretraining corpus is used, AgroGPT is fine-tuned on AgroInstruct—a dataset of 70k synthetic QA pairs created from agricultural images using LLM-generated captions and instructions. It is evaluated on AgroEvals, a domain-specific benchmark for fine-grained crop disease and pest identification. AgroGPT demonstrates superior performance over generalist models and human baselines in image-based agronomic reasoning.

\textbf{Neuroscience.} Recent advances in LLMs for neuroscience have integrated both neuroscience literature and neural data from multiple modalities such as EEG and fMRI to improve interpretability and performance on brain related tasks.
\emph{BrainGPT}~\cite{luo2025large} is a domain specialized language model for neuroscience, fine tuned from Mistral 7B using low rank adaptation on 1.3 billion tokens from neuroscience literature. Evaluated on BrainBench, a benchmark for neuroscience-related question answering, BrainGPT outperformed both general models and human experts.
\emph{EEG-GPT}~\cite{kim2024eeg} is a domain-specific LLM based on OpenAI’s GPT-3 (da Vinci), designed for EEG classification and interpretation. It achieves strong few-shot performance using only 2\% of training data and employs tree-of-thought reasoning with specialist EEG tools for interpretable, step-wise decision-making.
\emph{NeuroLM}~\cite{jiang2024neurolm} is a multi-task foundation model that integrates EEG signals into a language modeling framework. It trains a vector-quantized tokenizer to convert EEG data into discrete neural tokens, and fine-tunes a GPT-2 ~\cite{radford2019language} language model with multi-channel autoregression and instruction tuning. The model is evaluated on neural decoding tasks including sleep stage classification, epilepsy detection, motor imagery decoding, and emotion recognition, demonstrating that incorporating neural representations significantly enhances brain signal analysis.
\emph{UMBRAE}~\cite{xia2024umbrae} unifies multimodal brain decoding by aligning fMRI signals with pretrained CLIP ~\cite{radford2021learning} visual features via a universal brain encoder. Cross-subject training promotes subject-agnostic representations, which are connected through adapter modules to a Vicuna-7B/13B-based multimodal language model for semantic captioning and spatial grounding.
\emph{MindGPT}~\cite{chen2025mindgpt} is a GPT‑2–based model that decodes visual stimuli from non-invasive brain recordings into natural language. It integrates a CLIP-guided encoder with cross-attention mechanisms to align brain, visual, and linguistic representations, enabling accurate semantic interpretation of visual experiences.
\emph{MindLLM}~\cite{qiu2025mindllm} is a subject-agnostic model for fMRI-to-text decoding that combines a neuroscience-informed encoder with Vicuna-7B. Trained via brain instruction tuning, it supports a wide range of tasks—including perception, memory retrieval, symbolic language processing, and reasoning—achieving flexible and accurate semantic interpretation of brain activity.
\emph{UniMind}~\cite{lu2025unimind} is a general-purpose EEG foundation model that leverages InternLM2.5 to unify multi-task brain decoding by bridging the modality gap between neural signals and language representations. It introduces a Neuro-Language Connector to distill spatiotemporal EEG patterns into LLM-interpretable embeddings and employs a Task-aware Query Selection mechanism for adaptive task-specific decoding, achieving robust performance across diverse EEG tasks without task-specific fine-tuning.
\emph{Neuro-GPT}~\cite{cui2024neuro} is built on the open-source GPT-2 model, combined with a convolutional-transformer EEG encoder trained using self-supervised learning. It reconstructs masked EEG segments from large-scale clinical data and demonstrates strong generalizability in downstream motor imagery classification tasks.


\subsubsection{Astronomy}
\label{sec:scillms_specific_astronomy}

In this section, we review recent advances in astronomy-specific LLMs, highlighting representative models such as AstroLLaMA~\cite{astrollama}, AstroLLaVA~\cite{astrollava}, and AstroSage~\cite{MCQ3}. These models are generally built upon LLaMA-2 or LLaMA-3 architectures, with LLMs focusing on text understanding and generation, and MLLMs incorporating visual encoders (\eg, CLIP ViT-L/14) and projection layers to integrate astronomical images with text. Most models follow a two-stage training strategy: continual pre-training (CPT) using large-scale astronomy literature (\eg, arXiv abstracts, Wikipedia, textbooks) to enhance general domain understanding, and SFT using domain-specific tasks, such as question answering, multiple-choice reasoning, and synthetic dialogue generation. Low-Rank Adaptation (LoRA)~\cite{hu2022lora} and other parameter-efficient tuning methods are commonly used for resource-effective adaptation.  These developments lay the foundation for a new generation of astronomy-focused models, which we detail below in terms of their architectures, training pipelines, and domain-specific capabilities.

\emph{AstroLLaMA}~\cite{astrollama} is an astronomy-specific language model fine-tuned from LLaMA-2. It focuses on traditional language modeling tasks, with text as the modality. The model was fine-tuned using over 300,000 astronomy abstracts (approximately 95 million tokens) from the arXiv database and employs LoRA to improve resource efficiency. In a text generation task, the model was tested by having it produce astronomy-related abstracts. The results showed that AstroLLaMA achieved a 32.5\% reduction in perplexity compared to LLaMA-2, generating text that was more specific to the astronomy field and possessed deeper insights. Furthermore, AstroLLaMA's embedding space better reflects the semantic differences within astronomical text. Despite issues such as knowledge gaps and the generation of fictitious data, AstroLLaMA outperformed general-purpose models overall.
\emph{AstroLLaVa}~\cite{astrollava} is a multimodal visual-language model for astronomy that combines images and text. Built on the LLaVA 1.5 architecture, its visual encoder uses the CLIP ViT-L/14 model, and its language model is based on LLaMA 7B. Fine-tuning data is sourced from publicly available images and captions from NASA's ``Astronomy Picture of the Day'' (approximately 9,962 image-text pairs), the European Southern Observatory (approximately 14,617 image-text pairs), and the NASA/ESA Hubble Space Telescope (approximately 5,204 image-text pairs). GPT-4 is used to generate a synthetic dialogue dataset from the image captions. Training utilizes a two-stage fine-tuning strategy: in the first stage, only the visual-language projection layer is trained using astronomical image-text pairs, with the pre-trained visual encoder and language model fixed. In the second stage, synthetic astronomy question-answer pairs are used for instruction tuning, resulting in end-to-end fine-tuning of the entire model. The evaluation used the Galaxy 10 DECaLS dataset~\cite{G10}. The model was tasked with describing galaxy images from the G10 test set. The results show that AstroLLaVA performs slightly better than the LLaVA 1.5 model in the task of describing galaxy images.
\emph{AstroSage-LLaMA-3.1}~\cite{MCQ4} is based on Meta's LLaMA-3.1 model. Like AstroSage-LLaMA-3.1-8B, this model is trained in two main phases: CPT and SFT. It also employs a model merging strategy to combine the strengths of multiple models. However, during the SFT phase, it is fine-tuned using a diverse dataset, including the LLaMA-Nemotron-Post-Training Dataset, the OpenHermes 2.5 dataset, and domain-specific QA datasets. Evaluation was performed using the AstroMLab-1 benchmark, which consists of 4,425 high-quality, human-verified multiple-choice questions from the Annual Review of Astronomy and Astrophysics paper, which were not included in the training set. The results show that AstroSage-LLaMA-3.1 achieved an accuracy of 86.2\% without enabling inference mode, surpassing all other open weights and proprietary models tested, proving that domain specialization can significantly improve the performance of the model in a specific domain.

These astronomy-specific models reflect the increasing maturity and specialization of LLMs and MLLMs in science. With better perplexity, semantic understanding, and strong performance on domain benchmarks, they show the value of targeted pretraining and fine-tuning.

\subsubsection{Earth Science}
\label{sec:scillms_specific_earth}

The application of LLMs in Earth science is undergoing a significant transformation, moving from general-purpose models to highly specialized, domain-adapted solutions. This shift is driven by the necessity to handle unique data characteristics, such as immense volume, high granularity, and diverse modalities. The advancements discussed here are rooted in the development of sophisticated, domain-specific datasets and innovative architectural designs tailored for scientific inquiry.

A foundational challenge in adapting LLMs for scientific domains is the scarcity of high-quality, expert-level instruction data. To bridge this gap, several works have focused on creating specialized text-based datasets.
In the field of geoscience, \emph{K2}~\cite{deng2024k2} was trained on GeoSignal, the first supervised instruction dataset enabling models to understand and respond to complex queries from geoscientists.
Similarly, \emph{ClimateChat}~\cite{chen2025climatechat} was built upon the ClimateChat-Corpus, a large-scale, high-precision dataset constructed through a semi-automated pipeline combining self-QA, web scraping, and self-instruct methods to enhance expertise in climate change topics. 
For ocean science, \emph{OceanGPT}~\cite{bi2023oceangpt} leveraged the DoInstruct Framework, which uses a multi-agent approach to automatically generate expert-level instructions, overcoming the prohibitive cost of manual annotation.

In the multimodal domain, the unique characteristics of remote sensing (RS) imagery have necessitated the creation of equally specialized datasets. 
\emph{EagleVision}~\cite{jiang2025eaglevision} was trained on the proposed EVAttrs-95K, the first large-scale dataset designed for fine-grained object-level understanding, enabling comprehension and description of intricate object attributes in RS imagery beyond simple classification.
\emph{EarthMarker}~\cite{zhang2024earthmarker} was supported by the RSVP dataset, containing approximately 3.65 million multimodal pairs of image-point-text and image-region-text, enabling nuanced interpretations guided by visual prompts.
For pixel-level grounding, \emph{GeoPixel}~\cite{shabbir2025geopixel} was trained on GeoPixelD, which provides over 50,000 grounded phrases and 600,000 object masks, achieving end-to-end segmentation in high-resolution images.
To address ultra-high-resolution imagery, \emph{GeoLLaVA-8K}~\cite{wang2025geollava} utilized the Background Token Pruning and Anchored Token Selection methods, enabling complex dialogue and reasoning on images up to 8K resolution.

\begin{figure*}[t!]
  \centering
  \begin{subfigure}[t]{0.32\textwidth}
    \centering
    \includegraphics[width=\linewidth]{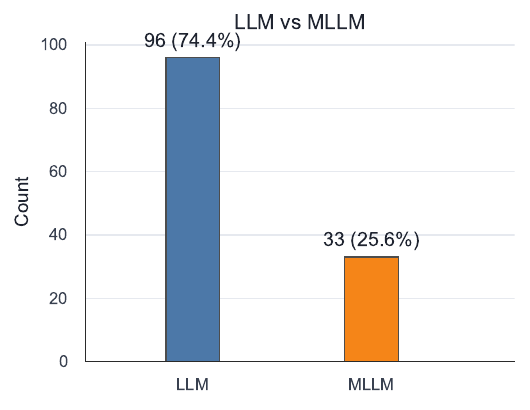}
    \caption{LLM vs MLLM ratio.}
    \label{fig:llm_mllm_ratio}
  \end{subfigure}
  \hfill
  \begin{subfigure}[t]{0.32\textwidth}
    \centering
    \includegraphics[width=\linewidth]{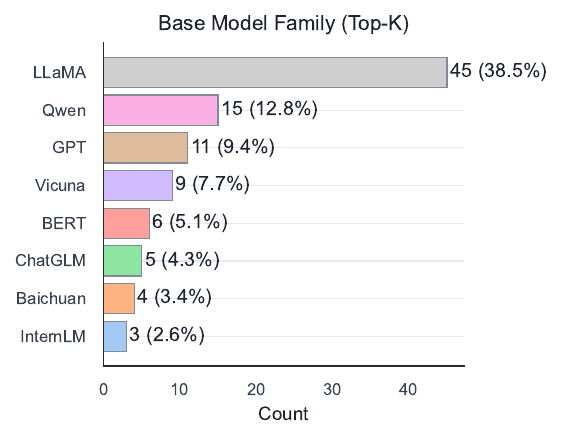}
    \caption{Base model family distribution (Top-K).}
    \label{fig:family_dist}
  \end{subfigure}
  \hfill
  \begin{subfigure}[t]{0.32\textwidth}
    \centering
    \includegraphics[width=\linewidth]{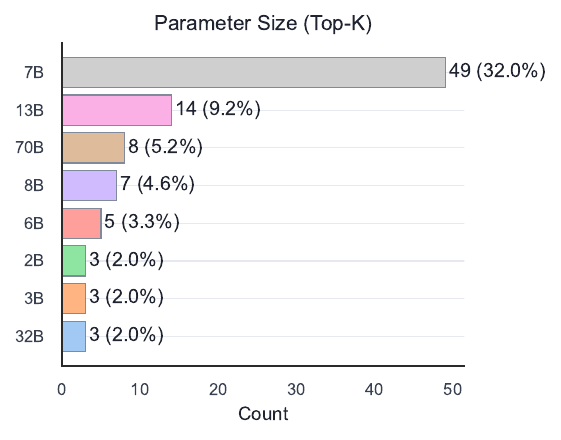}
    \caption{Parameter size distribution (Top-K).}
    \label{fig:param_size_dist}
  \end{subfigure}

  \caption{
    Statistical overview derived from Table~\ref{tab:sci_llms}.
    (a) Sci-LLM vs Sci-MLLM counts.
    (b) Base model family distribution; only top-$K$ are shown.
    (c) Parameter size distribution (all variants of multi-scale models are counted individually); only top-$K$ are shown.
  }
  \label{fig:sci_model_overview}
\end{figure*}

The scope of these models extends beyond static image analysis to encompass dynamic and multi-source data. \emph{EarthDial}~\cite{soni2025earthdial} was trained on EarthDial-Instruct, the largest remote sensing instruction-tuning dataset, comprising over 11 million instruction pairs across modalities like RGB, Synthetic Aperture Radar, and multispectral data, enabling reasoning over diverse Earth observation data.
HyperSIGMA~\cite{wang2025hypersigma} unifies HSI interpretation across tasks and scenes, scalable to over one billion parameters. 
SelectiveMAE~\cite{wang2024harnessing} dynamically encodes and reconstructs semantically rich patch tokens, thereby reducing the inefficiencies of traditional MIM models caused by redundant background pixels in RS images. 
RoMA~\cite{wang2025roma} enhances scalability for high-resolution images through a tailored auto-regressive learning strategy.
Furthermore, \emph{TEOChat}~\cite{irvin2024teochat} was powered by the proposed TEOChatlas, the first instruction-following dataset for temporal Earth observation data, making it the first vision-language assistant capable of engaging in dialogues about change detection and time-series analysis.
These innovative models, and the specialized datasets that train them, represent a significant step toward enabling more dynamic and comprehensive analysis for applications like environmental monitoring and disaster response.

\subsection{Sci-LLMs Analysis}
\label{sec:sci-llms_analysis}

Our survey highlights key trends in the development of Sci-LLMs. Roughly three quarters of current models are text-only LLMs, while MLLMs comprise only about one quarter (Fig.~\ref{fig:llm_mllm_ratio}). This imbalance reflects the dominance of text-based scientific sources (\eg, papers, patents, manuals) and the scarcity and cost of fine-grained multimodal supervision. Where MLLMs emerge—such as in medical imaging, life sciences, or remote sensing—they typically rely on smaller but higher-quality paired datasets that enable stronger cross-modal alignment.
Looking forward, as scientific discovery increasingly depends on integrating heterogeneous signals (\eg, astronomy that requires optical, radio, and X-ray observations to confirm cosmic events~\cite{angeloudi2024multimodal}, or climate science that unites satellite images, numerical models, and field reports~\cite{de2023machine}), the demand for Sci-MLLMs capable of synthesizing diverse modalities will grow. Thus, the current text-centric dominance may gradually give way to balanced multimodal ecosystems, powered by improved dataset curation and efficient alignment techniques.

The base-model landscape is now characterized by the primacy of open-source, general-purpose families, with LLaMA~\cite{touvron2023llama,touvron2023llama2,grattafiori2024llama} constituting the largest share and Qwen~\cite{bai2023qwen,team2024qwen2,yang2025qwen3} close behind, complemented by instruction-tuned derivatives (\eg, Vicuna~\cite{chiang2023vicuna}) and a thinner tail of encoder-style models (\eg, BioBERT~\cite{biobert}, ESM‑2~\cite{esm2}) that persist primarily in legacy or narrow-domain pipelines (Fig.~\ref{fig:family_dist}). Their dominance is explained by mature tooling, stable alignment recipes, scalable parameter ranges, and ultra-large pretraining corpora, which jointly enable low-cost adaptation and strong zero-/few-shot performance. In practice, open-source base models further facilitate rapid adaptation to emerging application scenarios by leveraging newly collected data via supervised fine-tuning (SFT), lightweight parameter-efficient methods, or modest instruction refinement. More broadly, progress is shaped by advances in training data curation and systems integrations, including retrieval-augmented workflows for maintaining up-to-date knowledge, high-quality expert QA and protocol-style instruction sets (\eg, DoctorGLM~\cite{xiong2023doctorglm}, MedAlpaca~\cite{han2023medalpaca}), targeted generation of challenging examples to improve coverage of rare cases, and the use of structured, tool-supported reasoning with simulators, analysis libraries, or code execution to support verifiable complex reasoning.

Across recent public tallies and our own tabulated statistics of released scientific models, parameter sizes in practice skew strongly toward smaller scales: 7B models constitute the largest share, 13B models are also frequent, while 70B-and-above models remain comparatively rare (Fig.~\ref{fig:param_size_dist}). This distribution reflects multiple deployment constraints, including privacy and compliance requirements (\eg, HIPAA, GDPR)~\cite{english2004hipaa}, inference latency and operational cost, the need for determinism and reproducibility, as well as on-premise, air-gapped, or data-sovereign environments. 
Many scientific tasks, such as protein folding, gene expression modeling, and materials discovery, are knowledge-dense yet narrow in scope, where small-to-mid sized models (7B–13B), when paired with retrieval augmentation or fine-tuning on scientific corpora, often achieve competitive performance relative to much larger counterparts~\cite{beltagy2019scibert}. Preferences for such models also mirror practical considerations: limited compute and memory in academic/lab settings, energy constraints, restricted access to sensitive datasets, and the complexity of deploying very large systems in regulated domains~\cite{reichstein2019deep}.
Looking ahead, as hardware efficiency (\eg, distributed training, mixed precision, memory optimization) and privacy/governance tooling advance, very large models are expected to play a greater role on the centralized training side, serving as knowledge sources, data generators, and evaluators. Nevertheless, distilled or quantized 7B–13B models are likely to remain the backbone for local and resource-constrained deployments, including in hospitals, laboratories, and field-deployed systems (\eg, satellites or environmental sensors)~\cite{he2024foundation}. These trends and drivers may vary across disciplines and institutions, and shares should always be interpreted with respect to the specific datasets and benchmarks at hand.

From a data–task interface perspective, several emerging themes are shaping the design and application of Sci-LLMs.
One promising direction is evidence-grounded generation with traceable provenance, which is essential for credible scientific outputs. Unlike general-purpose LLMs prone to hallucinations, Sci-LLMs are expected to produce verifiable claims with transparent source attribution, with data cards, citations, spatial or experimental coordinates, and retrieval logs serving as key scaffolds for trust and reproducibility~\cite{beltagy2019scibert,xiong2023doctorglm}.
Another challenge and opportunity lies in cross-modal alignment. High-quality supervision (\eg, region-level grounding in remote sensing) consistently yields better results than weakly aligned approaches where images are abstracted into generic embeddings~\cite{kuckreja2024geochat,wang2025geollava}.
A notable trend is the move toward agentic Sci-LLMs that integrate with scientific tools and workflows. Instead of static question-answering, these models are increasingly capable of retrieving literature, querying databases, running molecular or geospatial simulations, executing code for statistical analyses, and orchestrating lab or field data pipelines. This agentic behavior enables more reproducible and actionable scientific discoveries~\cite{xiong2023doctorglm,han2023medalpaca}.
Finally, temporal awareness and continual adaptation are becoming increasingly important, since scientific knowledge evolves rapidly. Versioned corpora~\cite{wang2020cord}, adaptive retrieval windows, and uncertainty calibration~\cite{liu2025chemauharnessreasoningllms} help models remain aligned with the current state of knowledge~\cite{reichstein2019deep}.
These patterns should be viewed not as fixed principles but as recurring observations that point to current bottlenecks and promising frontiers for Sci-LLM research.

Overall, the current landscape of Sci-LLMs is characterized less by architectural innovation and more by strategic adaptations of general-purpose foundations to domain-specific needs. The field remains heavily influenced by open-source base models, notably the LLaMA and Qwen families, which dominate due to their scalability, robust tooling, and strong zero-shot generalization. Model size skews toward the 6B–13B parameter range, reflecting pragmatic constraints around deployment costs, privacy-compliant inference, and operational efficiency in resource-limited environments such as clinics, labs, and edge devices. Performance gains are increasingly driven by sophisticated data pipelines and workflow integrations rather than pure scaling: knowledge-grounded generation provides verifiable outputs and supports hallucination mitigation~\cite{algaba2025deep,guan2024mitigating}, tool-assisted reasoning enables executable simulations and code, and high-quality cross-modal alignment supports meaningfully integrated understanding of text, images, structures, and geospatial data. Looking ahead, progress will hinge on improving verifiability and temporal adaptability—embedding provenance tracking, supporting continuous knowledge updates~\cite{algaba2025deep}, and refining agentic capabilities for real-world scientific tasks. As multimodal and tool-using paradigms mature, Sci-LLMs are poised to evolve from passive question-answering systems into active participants in the scientific process, facilitating discovery across biomedical, chemical, material, and environmental sciences~\cite{zhang2025evolving}.

\section{Scientific Data for Pre-training}
\label{sec:pre-training_data}



\begin{figure*}[t!]
  \centering
  \begin{subfigure}[t]{\textwidth}
    \centering
    \includegraphics[width=\linewidth]{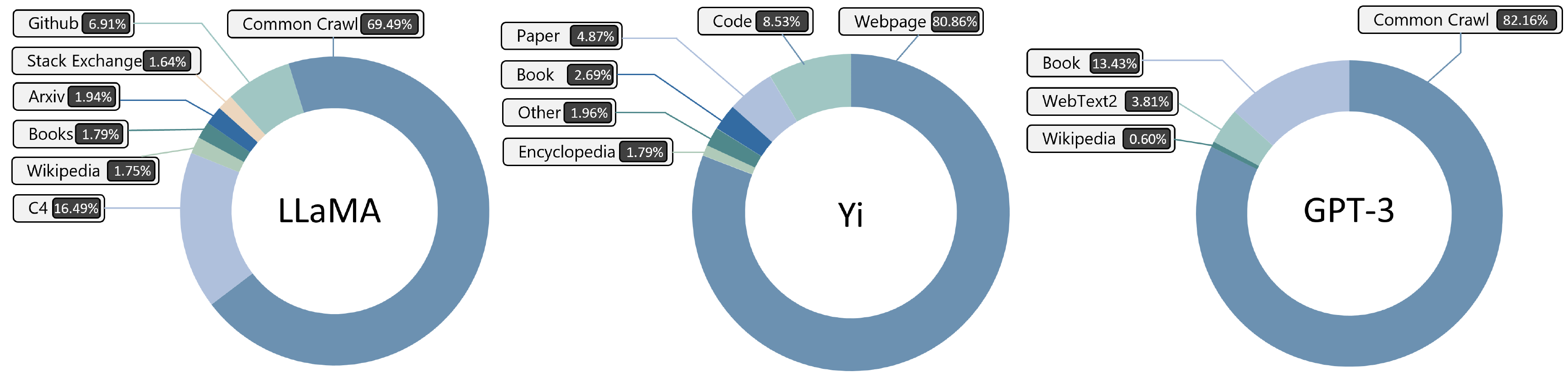}
    \caption{Pre-training dataset mixture of LLaMA~\cite{touvron2023llama}, Yi~\cite{young2024yi} and GPT-3~\cite{brown2020language}.}
    \label{fig:open_dataset_source}
  \end{subfigure}
  \vspace{1cm}
  \begin{subfigure}[t]{\textwidth}
    \centering
    \includegraphics[width=0.9\linewidth]{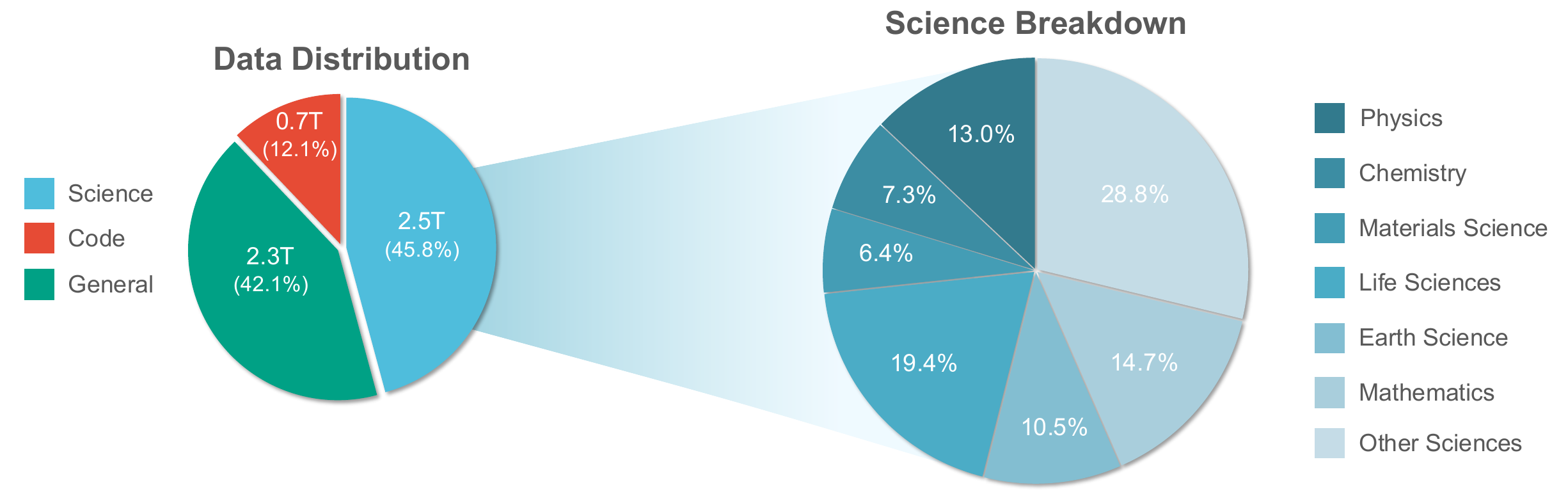}
    \caption{Distribution of continual pre-training data for Intern-S1~\cite{bai2025interns1scientificmultimodalfoundation}, involving 5.5T high-quality textual tokens with 2.5T scientific tokens across over six domains. Adapted from~\cite{bai2025interns1scientificmultimodalfoundation}.}
    \label{fig:interns1-pretrain-data}
  \end{subfigure}
  \vspace{-1cm}
  \caption{
    Pre-training dataset distributions for different language models.
    (a) Dataset mixture comparison across GPT-3, LLaMA, and Yi models.
    (b) Detailed distribution of Intern-S1's continual pre-training data with emphasis on scientific domains.
  }
  \label{fig:pretrain_data_overview}
\end{figure*}

Pre-training forms the foundation of LLMs and MLLMs by equipping them with broad, domain-relevant knowledge before task-specific fine-tuning. These models are typically pre-trained on massive and diverse datasets - for example, Yi~\cite{young2024yi} utilizes data from multiple sources including webpages, code, papers, and Q\&A content, while LLaMA~\cite{touvron2023llama}'s pre-training corpus spans approximately 1.4 TB across various domains such as CommonCrawl~\cite{commoncrawl}, GitHub, Wikipedia, and academic sources (Fig.~\ref{fig:open_dataset_source}). This extensive scale and broad coverage ensure models acquire comprehensive knowledge across different domains and languages. In the scientific domain, pre-training datasets must capture both the scale and diversity of knowledge, from symbolic laws of physics to complex biological systems and planetary processes. Unlike general web corpora, scientific datasets often combine structured experimental results, simulation outputs, specialized notations, and multimodal formats. The breadth and fidelity of these datasets directly influence a model’s ability to understand, reason, and generate within a specific scientific context. Looking at the scientific pre-training landscape, Intern-S1 exemplifies this specialized approach by dedicating 2.5T tokens (45.8\% of its total corpus) specifically to scientific content across six domains (Fig.~\ref{fig:interns1-pretrain-data}), providing the deep domain knowledge essential for superior performance on complex scientific tasks.
In the following subsections, we move from the smallest building blocks of matter (molecules and atoms) through complex biological systems and planetary-scale phenomena, concluding with interdisciplinary datasets that bridge multiple domains. The details of the pre-training datasets are summarized in Tab.~\ref{tab:training_datasets}.



\subsection{Physics, Chemistry and Material Sciences: the Foundation for Understanding the Material World}

Pre-training in physics, chemistry, and materials science focuses on representing the structure, dynamics, and properties of matter. These domains benefit from a combination of high-fidelity simulations, experimental measurements, and textual corpora that encode formal theories and experimental procedures. The challenge is balancing the precision of synthetic data with the complexity of real-world measurements, while integrating symbolic, numerical, and visual modalities.

\subsubsection{Physics}
In physics, the pre-training landscape is dominated by large-scale synthetic datasets derived from computational frameworks such as molecular dynamics (LAMMPS~\cite{thompson2022lammps}), 
finite-element methods, and cosmological simulations (Illustris~\cite{vogelsberger2014introducing}, Bolshoi~\cite{klypin2011dark}, as well as grid-based hydrodynamics with Enzo~\cite{bryan2014enzo}). These provide high-resolution spatiotemporal fields, wavefunctions, potentials, and other symbolic outputs that are invaluable for surrogate modeling and embedding physics-informed inductive biases. However, their controlled and often idealized nature makes it challenging for models to generalize to noisy or chaotic real-world conditions.

Experimental and observational datasets in physics, such as those from particle physics experiments including the European Organization for Nuclear Research (CERN)~\cite{CERN_OpenData} and the Large Hadron Collider beauty (LHCb)~\cite{lhcbexp}, condensed matter platforms including STM~\cite{Binnig1982_STM} and angle-resolved photoemission spectroscopy~\cite{damascelli2003angle}, or large astronomical observatories including Hubble Space Telescope (HST)~\cite{HST2025} and Atacama Large Millimeter/submillimeter Array~\cite{alma-science-archive}, are comparatively scarce in formats readily consumable by machine learning pipelines. The data are often fragmented across specialized repositories, use inconsistent formats, and may be restricted by access policies. Structured, standardized collections remain rare, limiting their use for large-scale pre-training.

Efforts such as the Galactica simulation database~\cite{chapon2024galactica} llustrate a move toward open, FAIR-compliant dissemination of astrophysics data. By centralizing metadata and reducing datasets from diverse simulation projects, Galactica covers cosmology, galaxy formation, and the interstellar medium, and supports generation of standardized, API-accessible high-level products (\eg, 1D profiles, 2D maps, 3D cubes). Although not yet matching the sheer scale of Illustris or Bolshoi, it contributes critical data diversity for building more generalizable physical science foundation models.

On the textual side, corpora such as SciBERT~\cite{beltagy2019scibert}, pre-trained on 1.14M full-text scientific papers (including physics literature), underscore the importance of domain-relevant language pre-training. Multimodal datasets like Multimodal ArXiv~\cite{li2024multimodal} which comprises 6.4M figure-caption pairs and 100K figure-based QA pairs, bridge visual and symbolic scientific reasoning. These complement simulation-heavy datasets by incorporating authentic visuals, diagrams, and plots, thus enriching models’ capacity for both symbolic reasoning and real-world data interpretation.

\subsubsection{Chemistry} 

Chemistry builds directly on physical principles to describe the structures, transformations, and properties of molecules and compounds. Pre-training datasets in chemistry reflect the field’s diversity of representations, including SMILES~\cite{zhang2024chemllm,tan2025chemmllm} and SELFIES strings for linear encodings, molecular graphs~\cite{cao2023instructmol} for connectivity patterns, and 3D coordinate formats (SDF, PDB) for spatial conformation~\cite{jiang2025chem3dllm}. These enable models to learn relationships between topology, stereochemistry, and molecular function.

Large, curated molecular libraries form the backbone of chemical pre-training. ZINC~\cite{irwin2005zinc} offers millions of commercially available drug-like compounds, ChEMBL~\cite{gaulton2012chembl} aggregates bioactive molecules with activity annotations, and MOSES~\cite{polykovskiy2020molecular} provides a standardized benchmark set for generative modeling. Early pre-trained models such as SMILES-BERT~\cite{smiles-bert} and more recent architectures like SMILES-Mamba~\cite{xu2024smiles} demonstrate how sequence-based learning can support tasks ranging from \emph{de novo} molecular generation~\cite{dst,mimosa} to property prediction~\cite{huang2020deeppurpose} and structure-based drug design~\cite{rga}.

Chemical reaction datasets expand the scope of pre-training to transformation pathways. The USPTO dataset~\cite{marco2015uspto}, containing million-scale reactions annotated with reactants, products, catalysts, temperatures, and other conditions, supports retrosynthesis planning, reaction outcome prediction, yield estimation, and catalyst selection~\cite{huang2022artificial,huang2021therapeutics}. Together, these datasets enable LLMs/MLLMs to model both static chemical structures and dynamic processes.

\subsubsection{Materials Science}
Materials science extends chemistry into the design, synthesis, and characterization of substances with tailored properties. Pre-training datasets in this field span multiple modalities: crystallographic structure files, chemical notation datasets, property-specific compilations, and large textual corpora.

Crystallographic datasets, encoded in CIF formats, are central for learning structural-property relationships. The Materials Project~\cite{jain2013commentary} offers over half a million entries covering known and predicted materials, while OQMD~\cite{Saal2013_OQMD} contains more than a million calculated electronic property records. ICSD~\cite{Zagorac:in5024} curates inorganic crystal structures, and specialized datasets such as CoRE MOF~\cite{chung2019advances}, QMOF~\cite{Rosen2021} and DigiMOF~\cite{Gubsch2022DigiMOF} target metal-organic frameworks. NOMAD~\cite{Scheidgen2023} and Materials Project Trajectory~\cite{Deng2023} scale up to millions of entries, incorporating dynamic simulation data.

Sequence representation datasets like USPTO~\cite{Lowe2017} and JARVIS-DFT~\cite{Choudhary2020} provide alternative chemical encodings (InChI, IUPAC, SELFIES), while large chemical libraries such as ZINC~\cite{Irwin2020} overlap with both chemistry and materials applications. Property-specific datasets~\cite{Pfeiffer2022} focus on targeted physical or mechanical attributes, enabling specialized pre-training for predictive modeling. Textual datasets like MatScholar~\cite{tshitoyan2019unsupervised}, with millions of publications, complement structured data by providing unstructured knowledge on material-property relationships.

Across physics, chemistry, and material sciences, pre-training datasets evolve from highly idealized simulations to richly annotated experimental corpora, and from symbolic equations to multimodal figure-caption pairs. The scale and diversity of these resources are critical: physics simulations anchor models in governing laws, chemical libraries teach molecular diversity and reactivity, and material databases bridge microscopic structures with macroscopic properties. Future progress will hinge on integrating these modalities by combining simulation outputs, experimental measurements, and literature-derived knowledge, to build foundation models capable of reasoning seamlessly from atomic-scale phenomena to engineered material systems.

\subsection{Life Sciences: Complexity from Molecules to Systems }
\subsubsection{Molecular and Cell Biology}
At the molecular scale, the central dogma, \ie, DNA-to-RNA-to-protein, shapes the data landscape for pre-training. Sequence-based datasets dominate, with different corpora focusing on small molecules, nucleic acids, or proteins.

Pre-training in the life sciences aims to equip LLMs and MLLMs with the ability to represent, reason about, and generate knowledge across the intricate hierarchy of living systems. This hierarchy begins at the smallest biological units (\eg, genes, proteins, and metabolites), progresses through cellular and tissue-scale processes, and culminates in organismal, clinical, and ecological contexts. Biological data is inherently heterogeneous: sequence strings, structural models, expression matrices, microscopy images, clinical narratives, and more. Effective pre-training datasets must therefore capture both the fine-grained molecular details and the higher-order interactions that emerge across scales, while aligning multimodal inputs into unified representations.

Current biology pre-training datasets for LLMs span multiple molecular modalities, with molecular, protein, and nucleic acid sequences constituting the primary data types. For molecular representations, several notable datasets have emerged: SPICE~\cite{eastman2023spice}, PCdes~\cite{zeng2022pcdes}, PubChemSTM~\cite{liu2023pubchemstm}, and MoMu~\cite{su2022momu} utilize SMILES strings or molecular graphs for pre-training, while TCPA~\cite{li2013tcpa} focuses on protein sequences. In the protein domain, UniRef~\cite{suzek2015uniref} databases serve as the foundational resource, with UniRef50 and UniRef90 containing approximately 49 million protein sequences after clustering at 50\% and 90\% sequence identity, respectively. For nucleic acid sequences, DNABERT~\cite{ji2021dnabert} utilized the human reference genome (Hg38.p13) for pre-training, while DNABERT-2~\cite{zhou2023dnabert} expanded to multi-species genomes from 135 species, creating a dataset 12 times larger than its predecessor. RNA pre-training has leveraged RNAcentral~\cite{rnacentral2019rnacentral} database with million-scale non-coding RNA sequences.

The evolution of these datasets reflects a clear trend toward multi-species, multimodal approaches and increased scale. Recent advances include sophisticated tokenization strategies, such as DNABERT-2's~\cite{zhou2023dnabert} Byte Pair Encoding (BPE) replacing traditional k-mer tokenization, and the incorporation of structural and functional annotations beyond raw sequences. Cross-modal pre-training has gained traction, with an increasing number of datasets~\cite{liu2023pubchemstm,su2022momu} bridging molecular structures with natural language descriptions, enabling more comprehensive molecular understanding. Future directions point toward larger-scale datasets that incorporate 3D structures, epigenetic modifications, and cross-species evolutionary relationships, as evidenced by emerging comprehensive benchmarks~\cite{zhou2023dnabert} for systematic model evaluation across diverse genomic tasks.

\subsubsection{Multi-Omics}
Multi-omics pre-training aims to unify genomics, transcriptomics, proteomics, and beyond into integrated representations. 

At the genomic level, pre-training corpora often start with the complete human reference genome (GRCh37~\cite{cole2008finishing}) and population-scale sequences from projects like the 1000 Genomes Project~\cite{siva20081000}. To enhance generalization and cross-species utilization, pretraining corpora are often further expanded to encompass genomic sequences from multiple species, such as archaea, fungi, and vertebrate mammalian, collected from scientific repositories such as NCBI GenBank~\cite{benson2012genbank}. Omni-DNA~\cite{omni-dna} constructs a 30B-token corpus by sampling and deduplicating genomic fragments from NCBI’s multi-species genome database, covering bacteria, archaea, fungi, plants, and vertebrates. GeneChat~\cite{genechat} focuses on encoding human genomic syntax and semantics by extracting 1024 bp fragments from the GRCh38 reference genome. DNAHLM~\cite{dnahlm} adopts a hybrid corpus of equal-size genomic and textual data, drawing DNA sequences from the GRCh38 human genome and papers from Wikipedia. More recently, BioReason~\cite{fallahpour2025bioreason} extends beyond sequence modeling by incorporating a dual-channel corpus consisting of PubMed and Wikipedia texts alongside a large-scale gene-gene interaction graph built from sources like STRING, Reactome, and Gene Ontology, enabling joint pretraining across natural language and biological relational structures.

In transcriptomics, early large-scale pretraining efforts have focused on gene expression matrices derived from single-cell RNA sequencing (scRNA-seq) data. Foundation models~\cite{yang2022scbert,cui2024scgpt} are typically trained on datasets including HCA and Tabula Muris, where expression profiles are represented as gene tokens or gene-expression pairs. Moving beyond unimodal expression, scMMGPT~\cite{scmmgpt} demonstrates a large-scale dataset with natural language data, involving over six million single cells across three modalities: scRNA-seq, scATAC-seq, and RNA-protein measurements from CITE-seq. RNA-GPT~\cite{xiao2024rna} develops a training corpus with 130,102 full-length transcripts from the GENCODE v38 reference, boosting the unification of transcript-level RNA understanding and generation with language-level reasoning.

In proteomics, UniProtKB (Swiss-Prot and TrEMBL) serves as the foundational pretraining resource~\cite{xtrimopglm}. For example, ProteinLMDataset~\cite{proteinlm} is built by SIFTS-mediated mapping of protein data bank~\cite{berman2000protein} entries to UniProt, integrating billions of tokens from PubMed abstracts, Swiss-Prot and PMC full texts; Evolla\cite{evolla} extracts 14 M expert-curated Information Points from Swiss-Prot and clustered TrEMBL entries, which are then transformed into high-confidence question-answer pairs via an LLM-driven augmentation pipeline.

Emerging multi-omics corpora begin to unify diverse biological modalities, integrating sequence-level data with biomedical text. NatureLM~\cite{naturelm} assembles over 3.27 trillion tokens from 35 biomedical corpora encompassing molecular sequences, clinical narratives, literature, and imaging-derived captions. This massive collection incorporates structured omics repositories such as UniProt, GENCODE, and the Human Protein Atlas alongside unstructured text from medical corpora like PubMed, enabling alignment between textual semantics and molecular features across scales. LLaMA-Gene~\cite{llama-gene} curates a multimodal biomedical instruction corpus by aligning 6.2 million natural language queries with structured molecular knowledge graphs derived from GeneCards~\cite{stelzer2016genecards}, OMIM~\cite{amberger2015omim}, and Ensembl~\cite{harrison2024ensembl}. This results in paired representations of gene-level annotations, phenotypes, diseases, and variant consequences, supporting instruction-tuned pretraining for gene-centric biomedical reasoning. ChatNT~\cite{chatnt} constructs a fully multimodal instruction dataset comprising 605 million DNA tokens and 273 million English tokens, covering 27 downstream tasks involving DNA, RNA, and protein processes. Together, these works exemplify a paradigm shift toward integrative instruction datasets that fuse omics, clinical, and textual domains into unified token spaces for large-scale pretraining.

\subsubsection{Neuroscience}
In the field of neuroscience, pretraining primarily entails two components: extensive text corpora of neuroscience literature and modality‐specific encoders pretrained on brain signals such as EEG, fMRI, and MEG. The literature corpus, exemplified by the BrainGPT~\cite{luo2025large} comprises approximately 1.3 billion words drawn from 332,807 abstracts and 123,085 full-text articles in the PubMed Central Open Access Subset, covering 100 high-impact journals (\eg, \textit{Nature}, \textit{Cell}, \textit{Neuron}, \textit{PNAS}) published between 2002 and 2022. The LaBraM~\cite{jiang2024large} framework integrates over 2534.78 hours of EEG data from about 20 public and proprietary datasets, encompassing motor imagery, emotion recognition, grasp-and-lift tasks, P300 spelling paradigms, epilepsy detection, and resting-state recordings, with channel counts of 19-64 and sampling rates of 160-2048 Hz. 

\subsubsection{Healthcare and Medical Science}
Depending on the model type, pre-training strategies for medical models vary: LLMs are primarily trained on large-scale clinical and biomedical texts to acquire medical language understanding. However, when translated to MLLMs, they require another multimodal pre-training stage that aligns visual and textual modalities to develop image-grounded understanding. Accordingly, the pre-training datasets can be broadly categorized into text-only corpora for LLMs and image-text pairs for MLLMs.

Medical textual data contains essential domain knowledge. The textual corpora are dominated by conversational clinical dialogues~\cite{chen2023huatuogptii,han2023medalpaca,li2023chatdoctor,abacha2017overview,chen2020meddialog}. Clinical dialogues cover a wide range of outpatient scenarios, but their level of expertise and reliability is difficult to guarantee due to the absence of follow-up examinations for verification. 
Medical textbooks and research papers~\cite{luo2022biored,han2023medalpaca,wang2020cord,lozano2025biomedica,gao2020pile} help address this issue, serving as critical sources of knowledge in the medical domain. 
Electronic Health Records (EHR)~\cite{wang2020mimic,li2024mediq,smith1988using,zheng2024efficiently} include basic demographic data, summaries of major diseases and health issues, and key healthcare service records, providing longitudinal health information of patients over time. However, EHR datasets suitable for reasoning over temporal patient trajectories are still scarce. 
Clinical reports~\cite{johnson2019mimic,chambon2024chexpertplus,wu2023medmd,huang2021deepeyenet,bai2024m3d} document the entire patient journey, ranging from admission and examination to diagnosis, treatment, and follow-up. However, access to such reports typically requires strict ethical review and entails potential privacy risks, which limit their overall availability and scale.

For MLLMs, image-text pre-training datasets play a central role. Large-scale corpora such as PMC-OA~\cite{lin2023pmcoa}, ROCOv2~\cite{ruckert2024rocov2}, MedICaT~\cite{subramanian2020medicat}, and Open-PMC-18M contain millions of biomedical figures and their associated captions, largely sourced from academic literature. Datasets like MIMIC-CXR~\cite{johnson2019mimic}, CheXpertPlus~\cite{chambon2024chexpertplus}, and PMC-CaseReport~\cite{wu2023medmd}, on the other hand, provide detailed diagnostic reports with finer-grained information derived from the corresponding medical images. 
These datasets cover a wide range of modalities, including CT, MRI, X-ray, ultrasound, PET, endoscopy, and histopathology, offering diverse supervision signals for learning visual-semantic correspondence. 
Domain-specialized image-text corpora also exist to target specific medical subfields. For example, MM-Retinal~\cite{wu2025mmretinalv2} focuses on ophthalmology, while Quilt-1M~\cite{ikezogwo2023quilt} concentrates on histopathological imagery with expert-vetted captions. These datasets serve to refine model understanding in narrowly scoped visual domains where general medical datasets may lack coverage.

Beyond static medical images, medical videos also encapsulate essential domain knowledge, including educational content for clinical training, patient simulation~\cite{wang2025medgen}, surgical procedures~\cite{hu2024ophclip}, and other clinically relevant scenarios. 
Models can learn comprehensive diagnostic and therapeutic knowledge from such videos. However, there remains a significant gap in scale between medical videos and medical images.

Despite their scale and variety, existing datasets in the healthcare and medical sciences domain show a striking modality imbalance, where medical image data occupies a significant position among all datasets, with the majority centered around radiological imaging. Further, for multimodal pre-training data, the annotation quality remains variable, ranging from noisy figure caption to partially validated annotations, which can affect model reliability.

\subsubsection{Agriculture}
In the agricultural domain, LLMs are generally pre-trained using corpora compiled from millions of multilingual agronomy journal articles, tens of thousands of professional textbooks, and genomic sequence databases.
The construction of such pre-training datasets typically involves a labor-intensive pipeline including OCR processing, deduplication, and filtering of low-quality content.
Although several agricultural LLMs have been introduced~\cite{yang2025seedllm,yang2024pllama}, none of their domain-specific pre-training datasets have been publicly released, hindering reproducibility and further research.

\subsection{Astronomy and Earth Science: Understanding Our Planet}
Astronomy and Earth science datasets expand scientific LLM/MLLM pre-training into domains where spatial, temporal, and spectral diversity is immense. They provide observational records, simulation outputs, and literature that span cosmic scales and Earth’s interconnected physical systems. For LLMs, pre-training relies heavily on textual resources derived from research publications, mission archives, and observational metadata. For MLLMs, multimodal corpora integrate high-resolution imagery, time-series data, maps, and spectra with descriptive text, enabling models to connect visual and quantitative patterns with domain-specific narratives.

\subsubsection{Astronomy}

Astronomy is among the most data-intensive scientific fields, yet large-scale, open, and multimodal pre-training datasets remain rare. Existing resources are fragmented across text, image, and spectral modalities, each with distinct acquisition challenges. While simulation-heavy domains like physics can generate abundant synthetic corpora, astronomical data acquisition depends on long-term sky surveys with large telescopes, such as LAMOST~\cite{LAMOST} and Gaia~\cite{Gaia}, making large-scale datasets costly and slow to compile. Moreover, observational modalities like images, spectra, and time-series differ in wavelength coverage, resolution, and signal-to-noise ratios, and are stored in heterogeneous formats with inconsistent calibration standards. Core physical parameters (\eg, stellar mass, metallicity) are often inferred indirectly via modeling rather than directly observed, limiting the availability of high-quality, labeled examples for supervised pre-training.

Among existing text-based datasets, resources like NASA ADS~\cite{ADSAPI} provide extensive corpora of astronomical research papers, abstracts, and technical documents. These have supported the construction of domain language models such as AstroBERT~\cite{astrobert}, trained for semantic understanding and entity recognition in astronomical contexts. SpecCLIP model~\cite{SpecCLIP} using LAMOST~\cite{LAMOST} and Gaia XP spectral data~\cite{Gaia}, aligns and reconstructs different spectral modalities through comparative learning. AstroPT~\cite{DESI-LS}, an image model built based on Dark Energy Spectroscopic Instrument Legacy Survey images, uses an autoregressive generative model to learn the potential distribution structure of galaxy images. However, such datasets typically focus on single modalities with narrow coverage, preventing the formation of a general-purpose astronomical foundation dataset. At present, text data remains the most tractable and widely used modality for pre-training in astronomy, while comprehensive multimodal datasets that integrate images, spectra, and time series are still largely absent.

\subsubsection{Earth Science}
Earth science remain less explored in pretraining dataset construction; most existing corpora in this field are derived from academic papers, textbooks, and similar sources. The scarcity is due in part to the dispersed and heterogeneous nature of Earth science data. Textual information is often embedded in PDFs of academic papers and textbooks, requiring complex parsing, while visual data (\eg, atmospheric variable fields, remote sensing imagery, and geological cross-sections) lacks the readily captionable semantic features found in natural images, making text–image alignment particularly challenging. 

Despite the scarcity of public pretraining datasets, several approaches to data construction offer valuable insights. For instance, EarthSE~\cite{xu2025earthse} leverages approximately 100,000 Earth science-related academic papers as its corpus. By employing advanced PDF parsing tools, these papers are converted into text, followed by automated annotation and data cleaning processes to produce high-quality datasets. Similarly, studies like ClimaQA~\cite{manivannan2024climaqa} extract structured corpora from Earth science textbooks. K2~\cite{deng2024k2}, on the other hand, gathers substantial textual data from internet sources, such as Wikipedia, relevant to Earth sciences. 

Although limited in scale and diversity for pretraining LLMs, these resources show that scholarly literature and curated web content remain the primary sources for Earth science textual data. Moving forward, integrating multi-source data, improving parsing techniques, and developing algorithms tailored for aligning Earth science images with text will advance pretraining dataset development in this field.

\subsection{Pre-training Data Analysis}

Across domains, current scientific pre-training corpora show clear strengths and equally clear gaps. 

\begin{figure}[tp]
    \centering
    \includegraphics[width=0.99\linewidth]{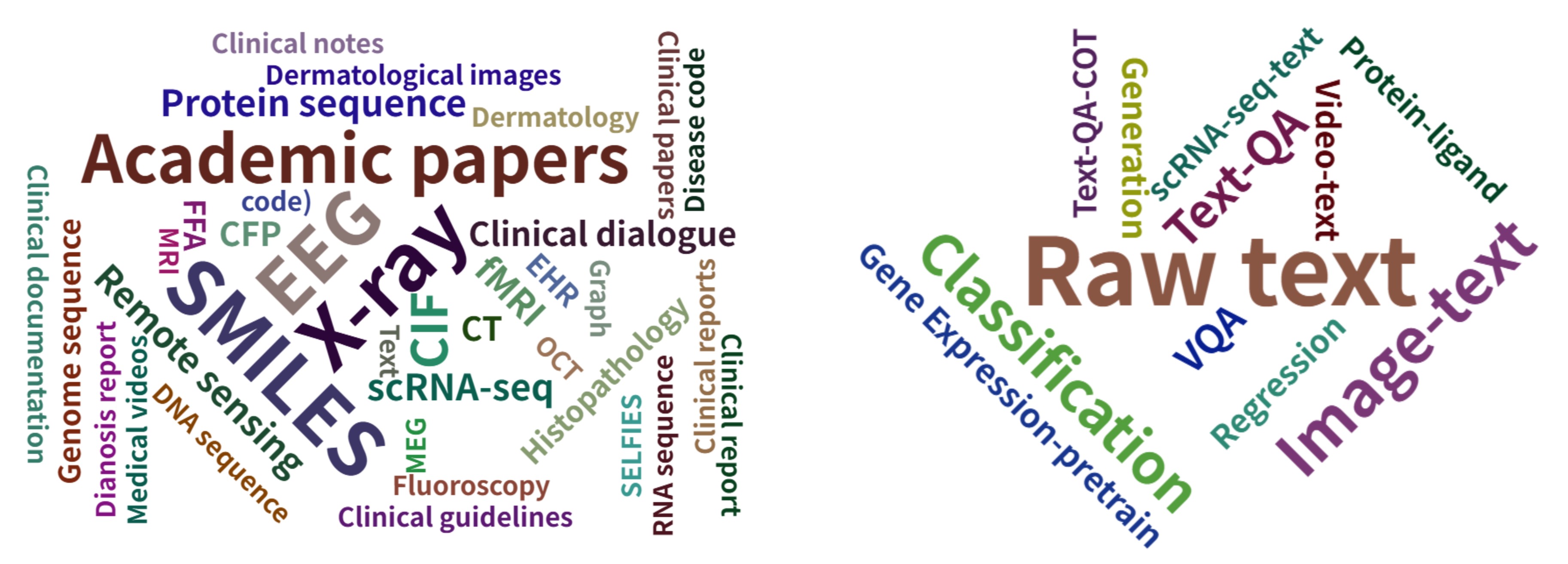}
    \caption{Word clouds of the pre-training dataset. The plots show the relative distributions of modalities (left) and types (right), with word size proportional to frequency.}
    \label{fig:pretrain_wordcloud_pt}
\end{figure}

Throughout the scientific landscape, the dataset ecosystem is both broad and heterogeneous, spanning text (papers, guidelines, EHR), symbolic structures (SMILES strings, CIF, gene and protein sequences), and multimodal pairings (figures, radiology, microscopy, spectra, videos). This diversity is illustrated in Fig.~\ref{fig:pretrain_wordcloud_pt}, which visualizes the relative distributions of pre-training data modalities (left) and task types (right). As shown, certain modalities such as academic papers, SMILES strings, and radiology images dominate, while others remain underrepresented; similarly, task types are heavily skewed toward raw text and classification. Such uneven coverage underscores both the breadth and imbalance of current scientific corpora, leading to several problems:

First, modality imbalance persists: physics remains dominated by idealized simulations~\cite{vogelsberger2014introducing,klypin2011dark}, which transfer poorly to noisy, instrument-specific observations, underscoring the simulation-to-observation gap. 
Second, many MLLM datasets rely on captions or rule-based labelers, yielding weakly grounded semantics~\cite{lin2023pmcoa,ruckert2024rocov2}, while even higher-quality radiology resources still depend on automatic pipelines that propagate labeling bias~\cite{chambon2024chexpertplus}. 
Third, heterogeneity and poor standardization impede cross-source fusion. For example, materials repositories (Materials Project~\cite{jain2013commentary}, NOMAD~\cite{Scheidgen2023}, OQMD~\cite{Saal2013_OQMD}) expose inconsistent metadata and calculation settings, complicating integrated pre-training and evaluation. Similar issues appear in astronomy, where multi-instrument spectra~\cite{SpecCLIP,LAMOST,Gaia} differ in bandpass, resolution, and calibration, challenging multimodal alignment. 
Fourth, some fields lack truly open, large-scale pre-training corpora: Earth science efforts~\cite{xu2025earthse,deng2024k2} remain text-centric and modest in scale, limiting broad generalization. 
Fifth, data governance constrains clinical/EHR corpora~\cite{johnson2023mimiciv,guevara2024large}, yielding smaller or temporally stale distributions relative to real-world care. 
Finally, scale–quality trade-offs are unresolved: massive chemical/molecular pools~\cite{Irwin2020,Pfeiffer2022} offer breadth but limited property curation, whereas targeted materials sets emphasize fidelity at the expense of coverage.

Such uneven landscape gives rise to a fundamental tension: scientific LLMs/MLLMs require rich, multimodal pretraining to support domain-aware reasoning, but collecting such corpora is often expensive and sparse. Therefore, classical large‑scale scaling for training general-domain models, which throws ever‑more tokens and parameters at the problem, is much less feasible for the development of scientific models. 

Efficient pretraining thus emerges as a critical design principle. Leveraging compute‑optimal scaling laws~\cite{jordan2022chinchilla,porian2024resolving} (\eg, models should balance parameters and tokens for optimal compute efficiency) offers a roadmap for budget‑aware model design. Techniques such as carefully curated data mixtures~\cite{xie2023doremi}, high‑quality subset selection~\cite{lee2021deduplicating}, and continual pretraining~\cite{shi2024continual,guo2024efficient} further promise to stretch domain‑limited scientific resources effectively.

\section{Scientific Data for Post-training}
\label{sec:post-training_data}

\begin{figure}[t!]
    \centering
    \includegraphics[width=0.80\linewidth]{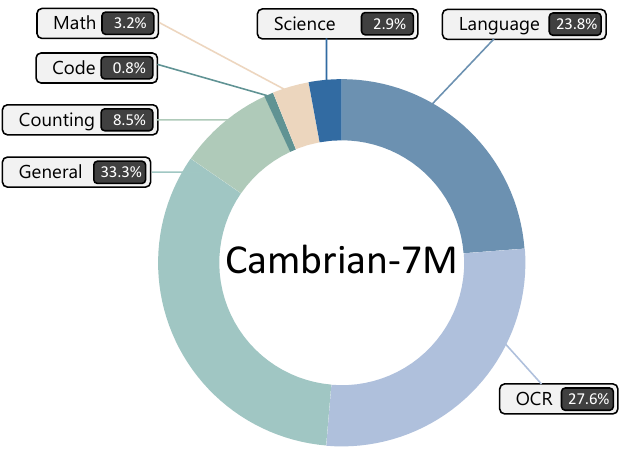}
    \caption{Composition of the Cambrian-7M~\cite{tong2024cambrian} instruction tuning dataset. 
    }
    \label{fig:cambrian10m}
\end{figure}

Post-training in scientific LLMs/MLLMs aims to align a pre-trained backbone, which is already equipped with broad factual knowledge, with the specific problem-solving styles and interactive workflows of scientific practice. Unlike pre-training which focuses on coverage and scale, post-training curates domain-specific, high-quality, and task-oriented datasets that teach models to solve problems, follow instructions, and explain their reasoning in ways aligned with disciplinary norms, moving beyond simple factual memory.

Across the sciences, post-training datasets have evolved from small, text-only instruction corpora toward large, multimodal, and reasoning-rich collections. However, these datasets vary greatly in sources, size, supervision type, and modality, reflecting differences in data availability, curation cost, and the maturity of AI adoption in each domain.

Small proportion of scientific data in current multimodal instruction tuning is exemplified by the Cambrian-7M dataset \cite{tong2024cambrian} (Fig.~\ref{fig:cambrian10m}), where science-specific content comprises only 2.9\% of the total training corpus, with the majority dominated by OCR (27.6\%), general knowledge, and language tasks.

\subsection{Current Landscape Across Scientific Domains}
The details of the post-training datasets are summarized in Tab. \ref{tab:training_datasets}.

\subsubsection{Physics}
Physics post-training datasets aim to move beyond fact recall toward the procedural and conceptual mastery that physicists use in practice. The scope spans multi-step derivation, formula reconstruction, unit consistency checks, experimental interpretation, and numerical estimation. These tasks demand both symbolic fluency and the ability to reason under physical constraints, which are often absent from generic LLM training corpora.

Existing open resources remain dominated by text-based QA formats, often adapted from educational or competition contexts. PIQA~\cite{bisk2020piqa} captures physical commonsense, including tool use and intuitive actions, though it stops short of formal derivations. SciBench~\cite{scibench} and the physics problems within the PhysicsArena~\cite{Dai2025PhysicsArena} benchamrk introduce computational questions with numeric computation and formula application, making them suitable for fine-tuning unit handling and basic symbolic manipulation. MATH500~\cite{lightman2023math500} is a curated 500-problem subset of the MATH~\cite{hendrycks2021measuring} benchmark spanning seven competition-style mathematics subjects; while it does not include a physics category, its algebraic and symbolic problems can help evaluate skills that are often prerequisite for physics problem solving.

Beyond direct extraction from exams and textbooks, synthetic or semi-synthetic resources increasingly scale coverage. Nemotron-Science~\cite{Bercovich2025LlamaNemotronER} subset contains teacher-generated reasoning traces across scientific domains including physics; NaturalReasoning~\cite{Yuan2025NaturalReasoningRI} contributes 2.8M challenging questions with reference answers and is widely used to distill long CoT from stronger models; and SCP-116K~\cite{lu2025scp116k} offers 116k automatically extracted problem–solution pairs in higher-education science (including physics), providing step-wise solutions without relying on LLM-generated CoT.

Overall, physics post-training datasets today provide a strong base for short-form problem-solving, with growing use of synthetic CoT corpora~\cite{Yuan2025NaturalReasoningRI,Bercovich2025LlamaNemotronER} to extend reasoning depth. However, most of them are text-only without figures or symbolic markup, failing to represent the dual textual-symbolic nature of physics reasoning. Further, post-training datasets still rarely capture the multimodal richness of real-world tasks, such as interpreting force diagrams, circuit schematics, or motion graphs, despite such modalities being central to the discipline.

\subsubsection{Chemistry}
Chemistry post-training relies on high-quality, task-specific datasets to fine-tune models for molecular property prediction, structure-based reasoning, and generative chemistry. Unlike pre-training corpora that may contain millions of weakly labeled compounds, post-training data is limited in scale due to the high cost of wet-lab experiments and structural determinations.

For example, drug-discovery ADMET datasets~\cite{huang2021therapeutics} are often limited to hundreds to thousands of entries because measuring absorption, distribution, metabolism, excretion, and toxicity requires time-intensive experiments. 
The Cross-Docked dataset~\cite{crossdock} contains 22.5M estimated 3D protein-ligand binding poses generated by molecular docking into multiple protein binding pockets, providing a large-scale resource for training and benchmarking structure-based drug discovery models. 
PDBBind~\cite{wang2004pdbbind} database stands out as a high-quality, manually curated resource that extracts experimentally validated protein-ligand complexes from the Protein Data Bank, each annotated with quantitatively measured binding affinity data, supporting both structural analysis and predictive modeling of binding strength. 

Chemistry datasets increasingly combine molecular formats (SMILES, InChI, 3D coordinates) with textual annotations~\cite{yu2024llasmol}, allowing LLMs to align symbolic chemistry representations with natural language descriptions. This multimodal pairing is key to enabling cross-format translation, \eg, predicting a compound’s IUPAC name from structure or vice versa.

\subsubsection{Materials Science}
Materials science post-training datasets are scarce and often repurposed from pretraining corpora. Molecular generation benchmarks like MOSES\cite{10.3389/fphar.2020.565644} and ChEBI-20\cite{edwards-etal-2021-text2mol} pair SMILES with text descriptions, supporting tasks from generation to captioning. ChEBI-20-MM~\cite{liu2025quantitativeanalysisknowledgelearningpreferences} extends these with richer metadata (InChI, IUPAC, polar area), enabling cross-format translation.
Apart from text and SMILE modalities, there are visual datasets from high-resolution characterization resources such as the Warwick Electron Microscopy Datasets~\cite{Ede_2020}, containing tens of thousands of STEM/TEM images and simulated wavefunctions. These enable image captioning, defect identification, and property inference when paired with textual descriptions. However, such visual data are limited. Most datasets lack multi-step reasoning traces, multimodal integration, or workflows that combine molecular design with property calculation and analysis.

\subsubsection{Life Sciences}

Life sciences post-training data spans diverse subdomains, each with distinct data modalities, supervision formats, and reasoning demands.

\emph{Molecular and cell biology} datasets include three main groups. First, sequence-to-function datasets such as PEER~\cite{xu2022peer} and BEACON~\cite{ren2024beacon} focus on protein and RNA property prediction. Second, large-scale instruction corpora like Mol-Instructions~\cite{fang2023mol}, OPI~\cite{xiao2024opi}, and PubChemSTM~\cite{liu2023pubchemstm} translate biochemical facts into conversational form, covering protein, nucleic acid, and small molecule entities, moving supervised fine-tuning beyond factual recall toward interactive QA. The third stream, still emerging, involves reasoning-focused datasets that pair each biology QA with an explicit chain-of-thought, such as ProCoT~\cite{jin2024prollm} for pathway reasoning and ToT-Biology~\cite{moremilk2025totbiology} for mechanistic explanations.

For \emph{multi-omics} post-training, DNA-focused datasets like Omni-DNA~\cite{omni-dna}, GeneChat~\cite{genechat}, and DNAHLM~\cite{dnahlm} frame genomics tasks (\eg, promoter detection, variant interpretation) as instruction-response pairs. RNA post-training includes single-cell and bulk expression modeling, as in scMM-GPT~\cite{scmmgpt}, which aligns scRNA-seq, scATAC-seq, and CITE-seq modalities with prompts describing biological contexts. Proteomics leverages UniProt-derived resources such as ProteinLMDataset~\cite{proteinlm} and Evolla~\cite{evolla}, creating hundreds of thousands to millions of protein-centric QA pairs. Multi-omics instruction sets like Biology-Instructions~\cite{Biology-Instructions} extend post-training to integrated DNA, RNA, and proteins, typically by synthesizing instruction-response pairs from reference databases and combining them with curated variant interpretation and functional annotation tasks.

In \emph{healthcare and medicine}, post-training data support a wide range of tasks with the most mature ecosystems: clinical dialogues (MedDialog~\cite{he2020meddialog}, ChatDoctor~\cite{li2023chatdoctor}, NoteChat~\cite{wang2023notechat}) for medical chatbots, medical image report generation (PMC-CaseReport~\cite{wu2023medmd}, MIMIC-CXR~\cite{johnson2019mimic}, CheXpertPlus~\cite{chambon2024chexpertplus}) for structured documentation, multimodal question-answering (EHRXQA~\cite{bae2023ehrxqa}, PubMedVision~\cite{huatuogptvision}, VQA-RAD~\cite{lau2018vqarad}, GMAI-VL-5.5M~\cite{li2025gmaivlgmaivl55mlarge}) for textual or visual comprehension, with chain-of-thought data (GMAI-Reasoning-10K~\cite{su2025gmai}) for step-by-step diagnostic reasoning on medical images. 

Post-training in \emph{neuroscience} refers to the alignment of measured neural signals, EEG, MEG, and fMRI, with the text embedding space of large language models to enable decoding of related semantics. The experimental tasks fall into several broad categories, including visual decoding, text decoding, sleep classification, clinical abnormality detection, motor imagery, emotion recognition, and workload assessment. In visual decoding, several rich benchmark datasets have been collected. Things-EEG1~\cite{grootswagers2022human} comprises EEG recordings from 50 participants responding to rapid serial visual presentation of 22,248 images covering 1,854 object concepts. Things-EEG2~\cite{gifford2022large} provides high temporal resolution EEG from 10 subjects over 82,160 image presentation trials drawn from 16,740 conditions selected from the THINGS database. The Natural Scenes Dataset (NSD)~\cite{allen2022massive} contains roughly 213,000 trials from eight subjects viewing 70,566 natural images, with blood oxygen level dependent responses captured using 7 Tesla fMRI at 1.8 millimeter resolution. Things-fMRI~\cite{hebart2023things} includes denoised responses from three participants to 8,740 images representing 720 objects, collected across 12 independent scanning sessions. To extend visual decoding into the realm of imagined content, NSD-Imagery~\cite{kneeland2025nsd} offers a benchmark with 2,304 mental imagery trials collected from NSD, with stimuli spanning simple shapes, complex natural scenes, and conceptual words. Complementing the fMRI-based work, Things-MEG~\cite{hebart2023things} records neural responses from four participants to the same 22,448 images (1,854 objects) with millisecond-level temporal precision. Neuro-3D~\cite{guo2025neuro} constructed the EEG-3D dataset, which contains EEG signals collected from 12 participants while they viewed 72 categories of 3D objects (both images and rotating videos).
For text decoding, the ZuCo collections capture EEG and eye-tracking data during natural reading and semantic annotation. ZuCo1~\cite{hollenstein2018zuco} recorded data from 12 native English adults reading over 21,000 words in 1,107 sentences across tasks such as sentiment judgment, entity relation recognition, and extraction of targeted relations like nationality, occupation, or employer. ZuCo2~\cite{hollenstein2019zuco} refines the experimental design by gathering EEG and eye movement data from 18 participants during both free reading and annotation specific to semantic relations, using 739 English sentences to better isolate cognitive differences between conditions.
Beyond decoding of visual and linguistic content, other neural domains contribute complementary signals. Sleep stage classification is supported by datasets such as HMC~\cite{alvarez2021inter}, SleepEDF~\cite{aboalayon2016sleep}, and SHHS~\cite{quan1997sleep}. Clinical abnormality detection focuses on disorders such as epilepsy, with datasets including TUEV~\cite{harati2015improved}, TUAB~\cite{harati2015improved}, and TUSL~\cite{von2017electroencephalographic}. Motor imagery is studied using the SHU~\cite{ma2022large} dataset. Emotion recognition draws on SEED~\cite{zheng2015investigating} and SEED-IV~\cite{zheng2018emotionmeter} to characterize affective states from neural activity. Cognitive Workload~\cite{zyma2019electroencephalograms} has been probed by collecting EEG from 36 healthy university students engaged in continuous mental arithmetic through serial subtraction, contrasting resting state with task periods to reveal neural correlates of load. Together, these datasets form a diverse and multi-task foundation for grounding brain activity in language model spaces and decoding semantics relevant to perception, cognition, clinical assessment, and internal mental states.

\emph{Agriculture} uses domain-specific instruction corpora (\eg, CROP~\cite{zhang2024empowering}) and multimodal VQA datasets (\eg, AgroInstruct~\cite{awais2025agrogpt}, MIRAGE~\cite{dongre2025mirage}) to adapt LLMs/MLLMs to crop health assessment, pest identification, and farm management.

Together, life sciences post-training data covers a broad modality spectrum from sequences and molecular graphs to clinical images and neural recordings, requiring models to unify understanding across vastly different biological scales.

%

\subsubsection{Astronomy}
Astronomy post-training data has evolved from pure-text corpora to rich multimodal resources. Early efforts collected hundreds of thousands of arXiv astronomy paper abstracts~\cite{astrollama}, embedding field-specific terminology and style. Later expansions included texts from introductions and conclusions~\cite{astrollama-chat}, as well as LLM-generated QA pairs from arXiv content, shifting toward interactive tasks.

To support more complex joint vision-language understanding tasks, post-training data construction incorporated multimodality. For instance, AstroLLaVA~\cite{astrollava} integrates NASA's ``Astronomy Picture of the Day'' and HST observation data~\cite{HST}, generating tens of thousands of image-caption pairs. Additionally, large-scale synthetic pipelines now leverage arXiv, astronomy Wikipedia, and textbooks to produce millions of domain-specific question-answer pairs~\cite{AstroMCQ,MCQ3,MCQ4}. For fine-grained tasks, such as named entity recognition in astronomy literature, manual curation remains essential, as seen in Astro-NER~\cite{Astro-NER}.

These datasets collectively enable models to handle domain knowledge understanding and multimodal image-text grounding for astronomical observation.

\subsubsection{Earth Science}
Earth science post-training datasets now span atmospheric, oceanic, terrestrial, and ecological domains.
Early examples like FloodNet~\cite{rahnemoonfar2021floodnet} paired remote-sensing images with templated questions. Automated pipelines such as EarthVQA~\cite{wang2024earthvqa} and TEOChatlas\cite{irvin2024teochat} expanded to hundreds of thousands of GIS-derived visual QA pairs. WeatherQA~\cite{ma2024weatherqa} introduced reasoning over weather composites, and SeafloorAI~\cite{nguyen2024seafloorai} scaled to millions of sonar-QA pairs.

Cross-sphere datasets also appeared, like GeoLLaVA-8K~\cite{wang2025geollava}, the highest-resolution vision-language datasets in remote sensing field to date, covering 22 real-world dialogue tasks. Supporting corpora like RS5M~\cite{zhang2024rs5m} and SkyScript~\cite{wang2024skyscript} offer millions of image-caption pairs across optical, Synthetic‑Aperture Radar, and Infrared (IR) modalities.

With increasingly automated annotation via advanced MLLMs like GPT-4 or Gemini-Vision, Earth science post-training data now enables not only scene captioning but also multi-step reasoning over complex Earth-system interactions.




\subsection{Post-training Data Analysis}

Existing post-training datasets share the following patterns and trends across domains.  

First, instruction-based corpora dominate, converting structured domain knowledge (\eg, databases, ontologies, benchmark tasks) into prompt-response pairs. These range from molecular biology and chemistry’s SMILES-language instruction sets~\cite{fang2023mol,yu2024llasmol} to astronomy’s literature-derived QA~\cite{AstroMCQ}, and from clinical dialogue datasets~\cite{wang2023notechat} to Geographic Information System (GIS)-to-question pipelines~\cite{irvin2024teochat} in Earth science.  

\begin{figure}[t!]
    \centering
    \includegraphics[width=\linewidth]{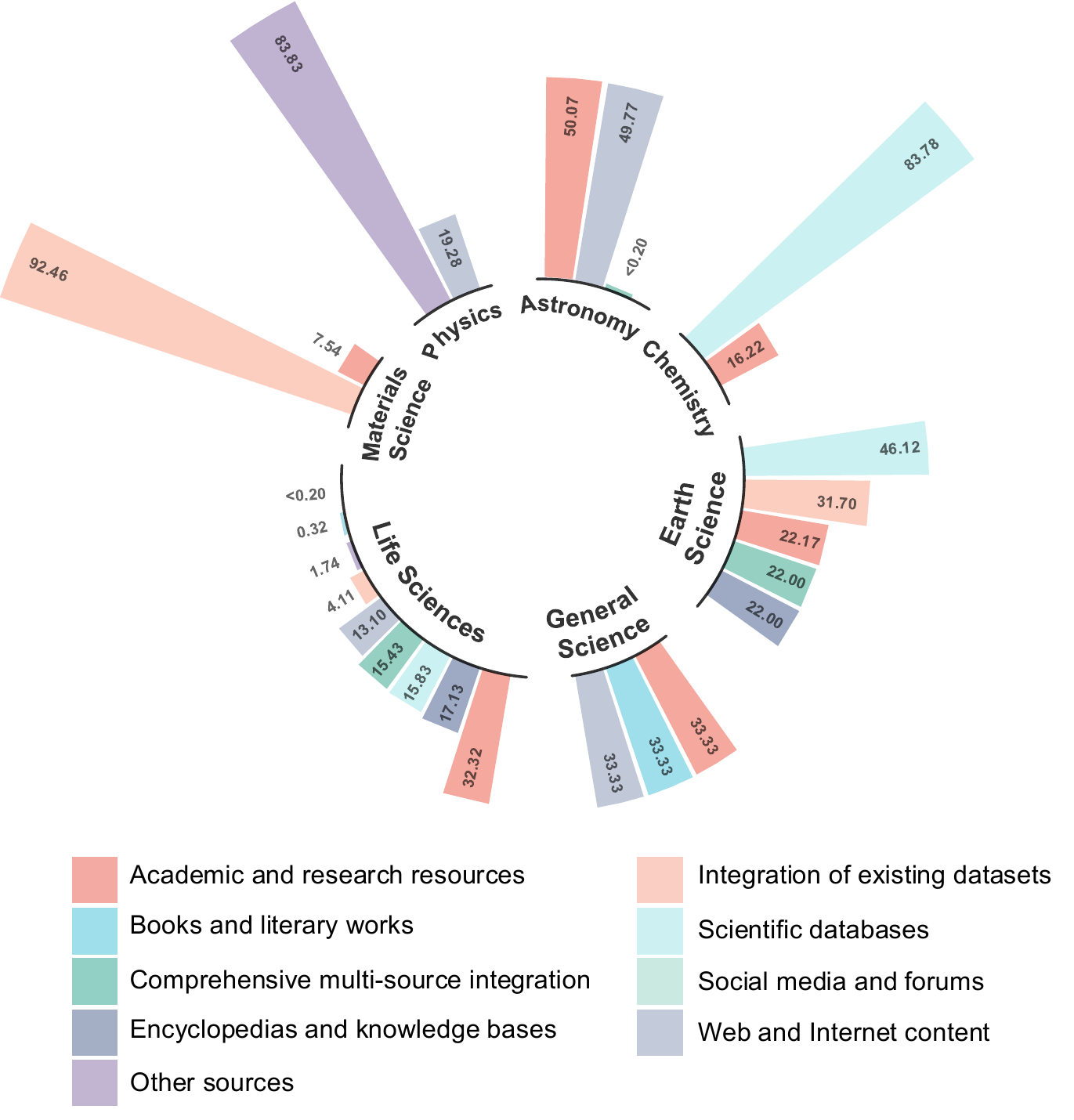}
    \caption{Source distribution of existing post-training corpora for scientific LLMs/MLLMs, normalized within each domain, showing significant domain-specific biases and cross-domain imbalance. These skews highlight where future corpus building could diversify inputs to reduce training bias and improve model generalization across disciplines.}
    \label{fig:posttraining_source}
\end{figure}

Another trend to be noted is the increasing importance of  multimodal and multi-domain corpora. Domains with rich data modalities (\eg, images), such as healthcare~\cite{johnson2019mimic,he2020pathvqa,chen2025slidechat}, astronomy~\cite{astrollava}, and Earth science~\cite{wang2025geollava}, now build VQA datasets or image-caption pairs to bridge visual and textual reasoning. Further, the multi-omics domain in life sciences typically require analysis across genomics, proteomics, and transcriptomics~\cite{llama-gene,naturelm}. In chemistry and materials science, SMILES strings, 3D molecular coordinates, microscopy images, and textual descriptions are increasingly co-annotated. This multimodal shift is crucial for teaching models to interpret data in heterogeneous forms and perform fluidly across related scientific subfields. As shown in Fig. \ref{fig:posttraining_source}, the source distribution of existing post-training corpora for scientific LLMs/MLLMs reveals significant domain-specific biases and cross-domain imbalances across different scientific fields. These skews highlight opportunities for future corpus building to diversify inputs, reduce training bias, and improve model generalization across disciplines.

\begin{figure}[tp]
    \centering
    \includegraphics[width=0.99\linewidth]{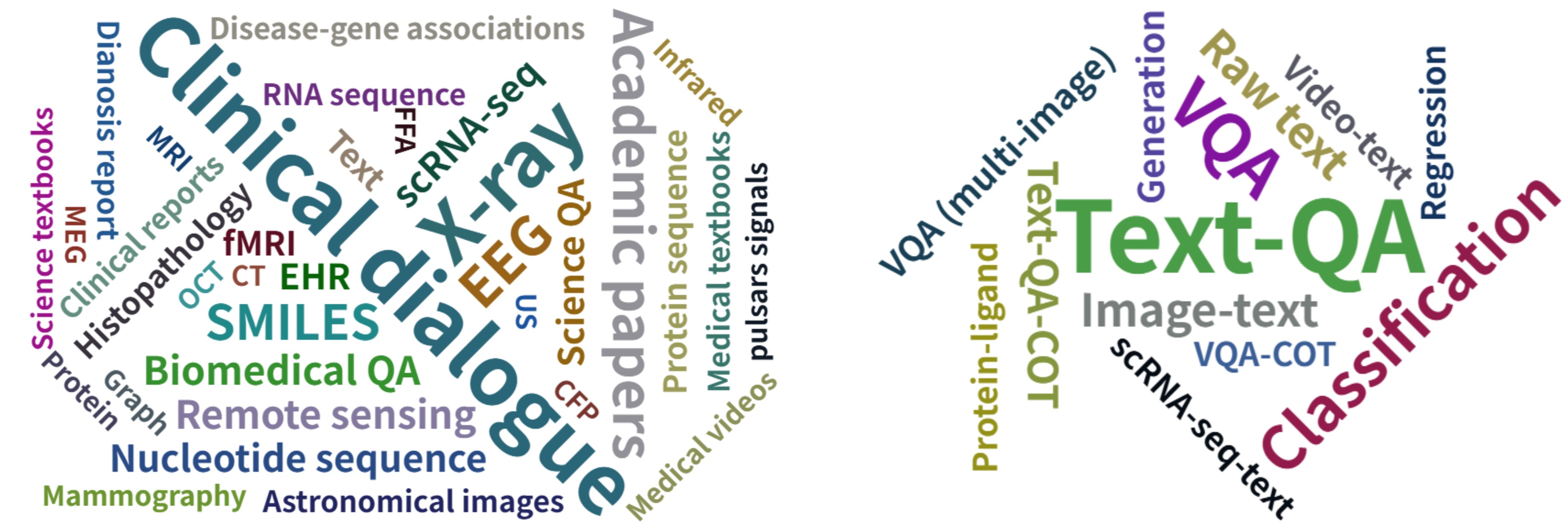}
    \caption{Word clouds of the post-training dataset. The plots show the relative distributions of modalities (left) and types (right), with word size proportional to frequency.}
    \label{fig:wordcloud_sft}
\end{figure}

Further, across domains, there is a clear trend toward explicit reasoning supervision beyond simple QA, driven by the need for models to handle complex, multi-step decision-making. However, these reasoning-oriented datasets are unevenly distributed. Biomedical sciences have begun producing chain-of-thought datasets for molecular pathways~\cite{jin2024prollm,moremilk2025totbiology} or multi-step diagnosis~\cite{wu2025medreason,sun2025reasonmed}, even for multimodal tasks~\cite{su2025gmai}; but large-scale, publicly available CoT corpora are relatively scarce in other domains.

Finally, scalable data synthesis has emerged as a practical solution to annotation bottlenecks. High-quality literature corpora, simulation outputs, and curated databases are now mined by LLMs to produce domain-specific instruction-response pairs~\cite{chen2025slidechat,yu2024llasmol,AstroMCQ} and reasoning traces~\cite{Bercovich2025LlamaNemotronER,chen2024huatuogpt} at scale, employing advanced techniques like multi-agent validation~\cite{sun2025reasonmed} to maintain fidelity, enabling the production of millions of domain-relevant samples that would be infeasible to curate manually. As illustrated in the word clouds of Fig. \ref{fig:wordcloud_sft}, post-training datasets encompass diverse modalities (left) and types (right), ranging from scientific representations like SMILES and nucleotide sequences to text-QA, image-text, and VQA content, reflecting the field's shift toward multimodal and integrated approaches.

In combination, these trends mark a shift from narrow, text-bound, single-domain resources toward broad, richly annotated, and operationally relevant datasets. This evolution positions post-training not merely as a final polishing step, but as a critical stage where scientific LLMs acquire the multimodal fluency, interdisciplinary reasoning, and tool integration skills necessary for real-world research environments.

Despite these advances, significant gaps remain. Datasets with multi-step reasoning traces tied to real experimental or computational workflows are still scarce in most domains. Some of existing CoT datasets~\cite{chen2024huatuogpt,sun2025reasonmed,Bercovich2025LlamaNemotronER} are distilled from existing reasoning models~\cite{o1,guo2025deepseek} without extensive expert validation. 
Moreover, multimodal coverage is uneven: while medicine, Earth science, and astronomy have rich image-text corpora, physics still lacks large-scale datasets that pair problems with diagrams or simulations. Licensing, privacy, and standardization also hinder dataset reuse, especially in healthcare and proprietary industrial research. 

Future efforts should prioritize \emph{integrated multimodal corpora}; process-aware datasets with explicit \emph{reasoning traces}, experiment design steps, and intermediate analyses; and \emph{tool-grounded examples} showing models how to invoke simulations, parse outputs, and iterate on hypotheses. Continuous post-training pipelines will be needed to keep pace with fast-evolving scientific data, blending automated ingestion with expert oversight. Synthetic data generation will remain essential, but should follow hybrid pipelines that combine automated scaling with human validation to maintain fidelity.

Ultimately, the goal is to move from LLMs that recall scientific facts to models that can operate as collaborative research assistants: reasoning across disciplines, working with tools, and adapting to new knowledge in real time.

\section{Evaluation of Sci-LLMs}
\label{sec:evaluation_data}

The evaluation of Sci-LLMs has increasingly gained attention as AI-for-Science (AI4Science) becomes integral to contemporary research. Recent developments in this area highlight the critical need for comprehensive assessment frameworks that evaluate model performance across diverse scientific disciplines, addressing multiple dimensions such as knowledge retention, understanding, reasoning, multimodality, and adherence to scientific values. Platforms such as SciHorizon~\cite{qin2025scihorizon} exemplify this trend, offering holistic benchmarking solutions that assess both AI-readiness of scientific datasets and fine-grained capabilities of LLMs across domains. In the following, we will explore the evolution and current status of scientific benchmark datasets, outlining their role in driving further advancements in AI4Science evaluation methodologies.

\subsection{Current Landscape Across Scientific Domains}

The evaluation of scientific foundation models across diverse disciplines has led to the development of specialized benchmarks that assess both domain-specific knowledge and reasoning capabilities. These benchmarks span from fundamental physics problems to complex biological systems, each differing in sources and targeted problems, designed to capture the unique challenges within their respective fields. The details of the evaluation datasets are summarized in Tab. \ref{tab:eval_datasets}.

\subsubsection{Physics}
In physics, evaluation benchmarks have evolved to test models across educational, competitive, and research-oriented tasks. At the foundational level, MM-PhyQA~\cite{Anand2024MMPhyQA} targets high-school physics via multimodal questions with explicit multi-image chain-of-thought prompting, while OlympiadBench~\cite{He2024OlympiadBencha} stresses bilingual, Olympiad-grade mathematics-and-physics problems with expert step annotations. PIQA~\cite{bisk2020piqa} and PROST~\cite{Aroca-Ouellette2021PROST} earlier emphasized physical commonsense through multiple-choice plausibility tasks, establishing a bridge from general commonsense QA into domain-specific physics. 

The progression continues through undergraduate-level challenges with PhysUniBench~\cite{Wang2025PhysUniBench}, PhysReason~\cite{Zhang2025PhysReasona}, and PhysicsArena~\cite{Dai2025PhysicsArena}, which systematically probe deeper physics reasoning through variable identification, process formulation, and solution derivation. UGPhysics~\cite{xu2025ugphysics} expands this scope by compiling bilingual undergraduate physics resources across mechanics, thermodynamics, and electromagnetism, while PhyX~\cite{shen2025phyx}, PHYSICS~\cite{feng2025physics}, and SeePhys~\cite{Xiang2025SeePhys} integrate text with diagrams and experimental setups to test multimodal reasoning in diverse physics domains. Complementing these, TPBench~\cite{chung2025theoretical} introduces advanced theoretical physics tasks spanning cosmology, relativity, and quantum mechanics, while PHYBench~\cite{Qiu2025PHYBencha} targets physical perception more broadly, introducing metrics like Expression Edit Distance to distinguish genuine reasoning from shortcuts.

Beyond problem-solving, physics benchmarks extend to equation discovery and symbolic regression. FSReD / AI Feynman~\cite{Feynman} supplies physics-grounded targets for symbolic regression, while SRBench~\cite{Cava2021Contemporary} establishes a living benchmark suite for comparing symbolic regression methods. LLM-SRBench~\cite{Shojaee2025LLMSRBencha} specifically targets scientific equation discovery with large language models, carefully designing problem splits to avoid trivial memorization.

Physical intuition in video is covered by IntPhys 2~\cite{Bordes2025IntPhys}, which presents synthetic scenarios requiring models to distinguish possible from impossible events, MVP-Bench~\cite{Krojer2025Shortcutaware} which constructs minimal video pairs to force true physical understanding, and MVBench~\cite{Li2024MVBench} which offers broad temporal multimodal video understanding tasks.

\subsubsection{Chemistry}
Chemistry benchmarks have similarly evolved to encompass both knowledge assessment and practical applications. ChemBench~\cite{zhang2024chemllm} and ChemEval~\cite{huang2024chemeval} provide comprehensive coverage of nine and 42 core chemistry tasks, respectively, while ChemMLLM~\cite{tan2025chemmllm} extends evaluation to multimodal chemistry research, including image-to-image translation for molecule optimization and text-to-image translation for molecular design. Specialized benchmarks target specific aspects: ChemSafetyBench~\cite{zhao2024chemsafetybench} focuses on safety issues of LLM responses in chemical experiments;  TrialBench~\cite{chen2024trialbench} focuses on clinical trial problems relevant to drug development, QCBench~\cite{xie2025qcbench} evaluates quantitative chemistry problem-solving across seven subfields from analytical to quantum chemistry, and PMO (practical molecule optimization)~\cite{gao2022samples} addresses molecular optimization with 23 objectives covering diversity, synthetic accessibility, and optimization ability. The critical role of spectroscopic data is captured by SpectrumWorld~\cite{yang2025spectrumworld}, which introduces 14 multimodal tasks spanning over 10 major spectroscopic techniques and 1.2 million distinct chemical compounds, evaluating models on spectrum-to-structure reasoning and spectral prediction from SMILES.

\subsubsection{Materials Science}
The intersection of chemistry and materials leads naturally to materials science benchmarks, which have evolved from traditional machine learning evaluations to LLM-specific assessments. MoleculeNet~\cite{wu2018moleculenetbenchmarkmolecularmachine} established early standards with over 700,000 compounds for molecular property prediction, while MatBench~\cite{dunn2020benchmarking} introduced specialized tasks for inorganic materials focusing on electronic structure and mechanical characteristics. Modern benchmarks like LLM4Mat-Bench~\cite{rubungo2024llm4matbenchbenchmarkinglargelanguage} advance the field with 1.9 million crystal structures supporting multiple input modalities, revealing important limitations of general-purpose LLMs in handling specialized representations like CIF files. Question answering capabilities are assessed through MaScQA~\cite{zaki2023mascqaquestionansweringdataset} and MatBookQA~\cite{mishra2024llamat}, which evaluate conceptual understanding and numerical reasoning in materials science. Generative capabilities are tested by GuacaMol~\cite{brown2019guacamol} and MOSES~\cite{10.3389/fphar.2020.565644} for molecular design tasks, while multimodal understanding is challenged by MMSci~\cite{li2025mmscidatasetgraduatelevelmultidiscipline} and MaCBench~\cite{alampara2024macbench}, mirroring real-world materials characterization workflows.

\subsubsection{Life Sciences}
Life sciences present particularly diverse evaluation challenges spanning molecular biology, healthcare, agriculture, and neuroscience. At the molecular level, benchmarks progress from DNA sequence understanding through DeepSEA~\cite{zhou2015predicting}, Ensembl collections~\cite{howe2021ensembl}, and Genomics-Long-Range~\cite{trop2024genomics} to small-molecule tasks with TOMG-Bench~\cite{li2024tomg} and MoleculeQA~\cite{lu-etal-2024-moleculeqa}. Higher-level biological reasoning is assessed by LAB-Bench~\cite{laurent2024lab} for wet-lab competence, GeneTuring~\cite{hou2023geneturing} for genomic knowledge retrieval, and Genome-Bench~\cite{yin2025genome} for multi-step CRISPR reasoning. MicroVQA~\cite{burgess2025microvqa} bridges microscopy and molecular function through expert-verified visual question answering. In video domain, SCIVID \cite{hasson2025scivid} is a cross-domain scientific video benchmark comprising five tasks across animal behavior, medical imaging, and weather forecasting. It includes diverse modalities (grayscale, RGB, multi-channel meteorological data), varying temporal scales, and tasks such as classification, point tracking, and spatiotemporal forecasting.

Healthcare evaluation emphasizes clinical knowledge through both text and visual modalities. Text-based benchmarks include BioASQ~\cite{krithara2023bioasq}, PubMedQA~\cite{jin2019pubmedqa}, and recent comprehensive efforts like MedBench~\cite{liu2024medbench}, MedXpertQA~\cite{zuoMedXpertQABenchmarkingExpertLevel2025b}, and HealthBench~\cite{arora2025healthbench} that approach board-exam rigor. Visual question answering progresses from focused datasets like VQA-RAD~\cite{lau2018vqarad}, PathVQA~\cite{he2020pathvqa}, and SLAKE~\cite{liu2021slake} to more comprehensive multimodal assessments in AMOS-MM~\cite{ji2022amos} and RP3D-DiagDS~\cite{zheng2024large}. More recently, with the rapid progress in Sci-LLMs and scientific agents, 
some challenging benchmarks have emerged for evaluating these advanced models. RareBench~\cite{chen2024rarebench} targets rare-disease diagnosis, compiling the largest open-source rare-patient dataset and assessing LLMs across tasks such as phenotype extraction and differential/disease screening. MedAgentBench~\cite{jiang2025medagentbench} provides a virtual EHR environment with 100 realistic patients and 300 clinician-authored tasks across 10 categories to benchmark medical LLM agents. 
AgentClinic~\cite{schmidgall2024agentclinic} evaluates multimodal agents by simulating clinical environments that require history taking, clinical interviewing, and sequential decision making. Agents will need to use tools and actively gather useful information through doctor–patient interactions for accurate diagnosis. 
These suites push evaluation beyond static QA toward interactive, end-to-end decision making aligned with real-world practice.

Agricultural applications are evaluated through SeedBench~\cite{ying-etal-2025-seedbench} for seed breeding capabilities, AgXQA~\cite{kpodo2024agxqa} for extension services, and AgEval~\cite{arbab2024ageval} for plant stress phenotyping. Neuroscience assessment combines knowledge-based evaluation through BrainBench~\cite{luo2025large} with semantic decoding tasks spanning visual decoding~\cite{grootswagers2022human,gifford2022large}, text decoding~\cite{hollenstein2018zuco}, and clinical applications including sleep classification~\cite{alvarez2021inter} and emotion recognition~\cite{zheng2015investigating}.

Multi-omics modeling has driven unified benchmark development across biological scales. RNA-specific evaluations have evolved from expression matrix tasks in scBERT~\cite{yang2022scbert} and scGPT~\cite{cui2024scgpt} to multimodal assessments in scMMGPT~\cite{scmmgpt} and comprehensive QA in RNA-GPT~\cite{xiao2024rna}'s RNA-QA dataset with over 400K entries. Cross-modal integration is exemplified by LLaMA-Gene~\cite{llama-gene}'s gene-centric instruction-following, NatureLM~\cite{naturelm}'s 50+ biomedical dataset evaluation, and ChatNT's 18-task genomics instruction suite covering processes from RNA degradation to protein stability.

\subsubsection{Astronomy}
Astronomy benchmarks utilize diverse data sources, ranging from scientific literature to observational data. AstroLLaMA~\cite{astrollama} and AstroMLab~\cite{MCQ1,MCQ3,MCQ4} utilize arXiv's astro-ph category for training and evaluation, while specialized tasks include Astro-NER~\cite{Astro-NER} for entity recognition and Astro-QA~\cite{Astro-QA} for question answering. Observational data processing is addressed through Starwhisper-pulsar~\cite{PCD} for pulsar classification, AstroPT~\cite{DESI-LS} for physical simulation acceleration, and visualization tools like ASTROVISBENCH. PAPERCLIP~\cite{PAPERCLIP} combines text and image data for literature analysis, while Pathfinder~\cite{pathfinder} provides efficient navigation of large-scale astronomical observations.

\begin{figure}[tp]
    \centering
    \includegraphics[width=0.95\linewidth]{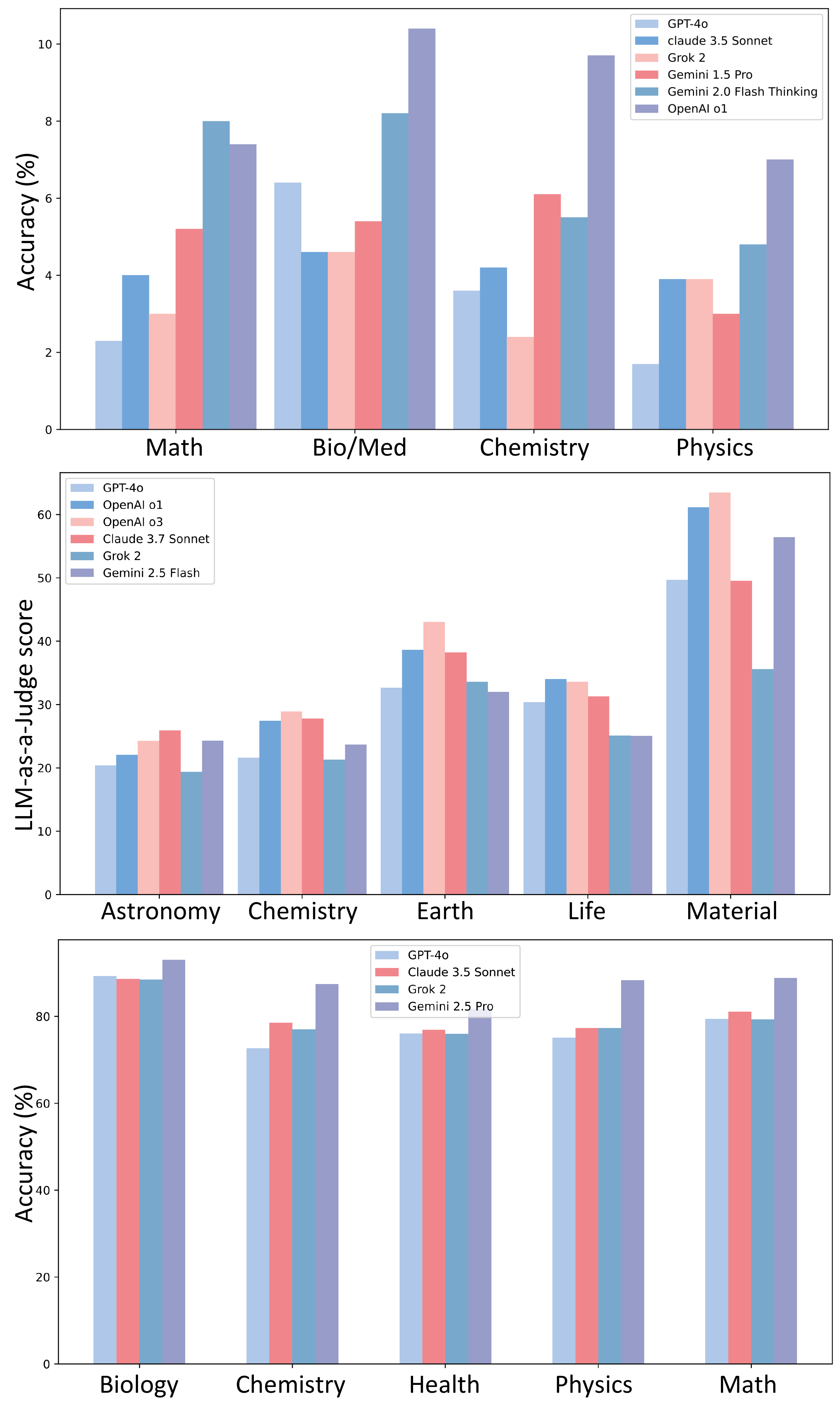}
    \caption{Performance of leading closed-source models drops significantly on challenging scientific benchmarks (HLE~\cite{phan2025humanity}, SFE~\cite{zhou2025scientists}) compared to MMLU-Pro~\cite{wang2024mmlu} across multiple domains. Top to bottom: HLE, SFE (en), MMLU-Pro.}
    \label{fig:evaluate_example}
\end{figure}

\subsubsection{Earth Science}
Earth science benchmarks focus on atmospheric studies and remote sensing applications. Atmospheric evaluation includes text-based ClimaQA~\cite{manivannan2024climaqa} and ClimateBERT~\cite{webersinke2021climatebert}, alongside multimodal WeatherQA~\cite{ma2024weatherqa}. Remote sensing benchmarks such as OceanBench~\cite{bi2023oceangpt}, RSIEval~\cite{hu2025rsgpt}, and XLRS-Bench~\cite{wang2025xlrs} emphasize satellite imagery interpretation through tasks including image captioning, visual question answer and visual grounding in ultra-high-resolution RS scenarios.. Interdisciplinary efforts like OmniEarth-Bench~\cite{wang2025omniearth}, EarthSE~\cite{xu2025earthse}, and MSEarth~\cite{zhao2025msearth} integrate data across hydrosphere, biosphere, lithosphere, and cryosphere, challenging models with complex cross-domain reasoning.

Across all these scientific domains, evaluation metrics have evolved beyond simple accuracy to include domain-specific measures: AUROC (Area Under the Receiver Operating Characteristic curve) and MCC for imbalanced biological data, exact match and MSE for symbolic regression, Expression Edit Distance for physical reasoning, validity and synthetic accessibility for molecular generation, and multimodal metrics like IoU (Intersection over Union) for visual grounding tasks. These benchmarks collectively reveal that while foundation models show promise in scientific applications, significant gaps remain in handling specialized representations, cross-modal reasoning, and the integration of domain expertise with general language understanding.

\subsubsection{General Science}
General-purpose science benchmarks have coalesced into three major strands: exam-style text QA that samples broadly across disciplines~\cite{wang2024mmlu,Huang2023CEval,zhong2023agieval}, multimodal figure/image QA that reflects the visual nature of scientific communication~\cite{lu2022learn,li2024multimodal,Yue2024MMMU,yueMMMUProMoreRobust2025}, and specialized formats that probe symbolic or programmatic reasoning beyond free-form answers~\cite{Shojaee2025LLMSRBencha}. 

Exam-style suites such as MMLU~\cite{wang2024mmlu}, C-Eval~\cite{Huang2023CEval}, and AGIEval~\cite{zhong2023agieval} provide wide coverage from secondary to undergraduate levels in both English and Chinese, enabling coarse-grained cross-lingual comparisons but often emphasizing short multiple-choice formats. Yet, as models saturated these leaderboards, newer variants emphasized robustness, harder distractors, and reasoning-heavy prompts (\eg, MMLU-Pro~\cite{wang2024mmlu}). 
In parallel, multimodal suites such as ScienceQA~\cite{lu2022learn} and MMMU~\cite{li2024multimodal} advanced beyond text by combining images, diagrams, tables, and interleaved text; MMMU-Pro~\cite{Yue2024MMMU} further filters out items answerable by text-only models and embeds questions in images to enforce genuine visual-linguistic integration, yielding substantially lower accuracies than on the original set. Graduate-level sets like GPQA~\cite{Rein2023GPQAa} and SuperGPQA~\cite{du2025supergpqa} target expert-authored, ``Google-proof'' scientific reasoning across biology, physics, and chemistry (and hundreds of graduate disciplines in SuperGPQA), helping to expose reasoning gaps that remain hidden on easier general-purpose tests.
%

These methodological choices clarify what is being measured—fact recall, modality integration, or multi-step reasoning; they help explain why success on broad academic exams does not automatically translate to scientific cognition under stricter evidence conditions.
Notably, there is a significant performance gap between general academic benchmarks and domain-specific scientific challenges. As shown in Fig. \ref{fig:evaluate_example}, while leading closed-source models achieve 80-95\% accuracy on MMLU-Pro~\cite{wang2024mmlu}, their performance drops dramatically on frontier scientific ``stress tests'' like Humanity's Last Exam (HLE)~\cite{phan2025humanity} and Scientists' First Exam (SFE)~\cite{zhou2025scientists}. Specifically, most models score only 2-10\% on HLE across various domains, with chemistry showing the best but still poor results. On SFE, despite relatively better performance in materials science, accuracy remains low at 20-40\% in other scientific domains. This stark contrast reveals that current LLMs, despite excelling at general knowledge tasks, struggle significantly with tasks requiring deep scientific reasoning and domain expertise.

Consequently, evaluation methodology in general science is pivoting toward designs that make reasoning requirements explicit and verifiable. 
Fixed-choice protocols report accuracy but implicitly test calibration via distractor design, making them sensitive to ambiguity and annotation artifacts; MMLU-Pro’s ten-option format and curated hard negatives reduce chance performance and inflate the penalty for shallow heuristic. MMMU-Pro’s vision-only setting removes textual crutches, isolating visual understanding from language priors and better reflecting figure-centric scientific communication.
SFE formalizes multimodal scoring with IoU, BERTScore, and LLM-as-a-Judge for structured visual tasks, while HLE introduces calibration error alongside accuracy to quantify overconfidence on hard scientific questions. Programmatic tasks like LLM-SRBench~\cite{Shojaee2025LLMSRBencha} enable exact-match and MSE for equations, and broad suites such as SciEval~\cite{scieval} and SciKnowEval~\cite{sciknoweval} aggregate multiple task families with diverse metrics to reflect the varied outputs typical in science. Together, these evaluations complement broad academic tests by injecting domain-shaped modalities, harder question design, and metric pluralism, thereby offering a more faithful picture of scientific reasoning than can be obtained from general benchmarks alone.

\subsection{Evaluation Data Analysis}

\begin{figure}[t!]
    \centering
    \includegraphics[width=\linewidth]{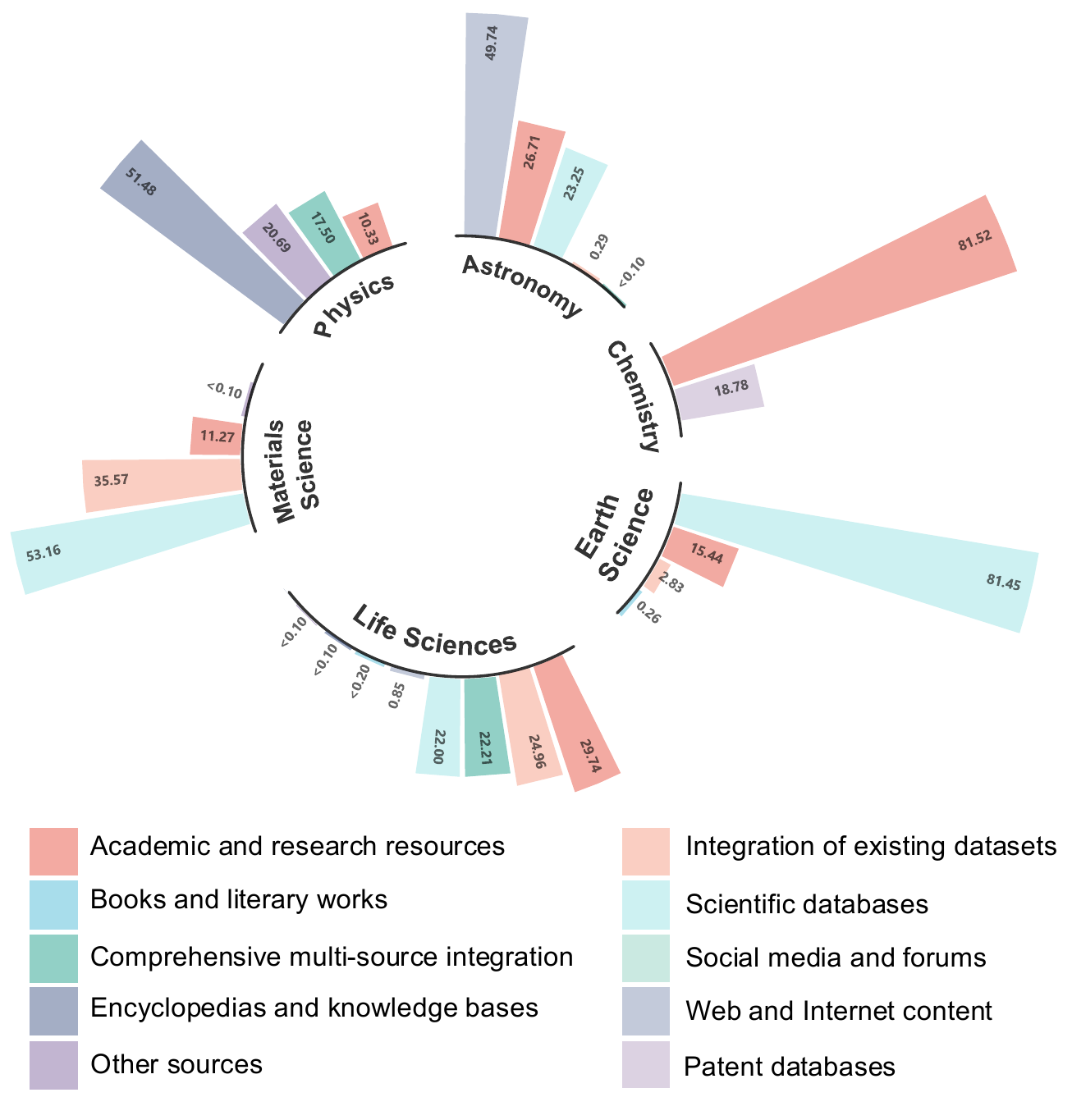}
    \caption{Source distribution of existing evaluation corpora for scientific LLMs/MLLMs, normalized within each domain. Most domains rely on a single dominant source type, showing today’s headline scores often reflect proficiency with one writing style or data type rather than robust, cross-domain scientific reasoning, highlighting the need for broader, more heterogeneous evaluation suites.}
    \label{fig:eval_sources}
\end{figure}

\begin{figure}[t!]
    \centering
\includegraphics[width=0.99\linewidth]{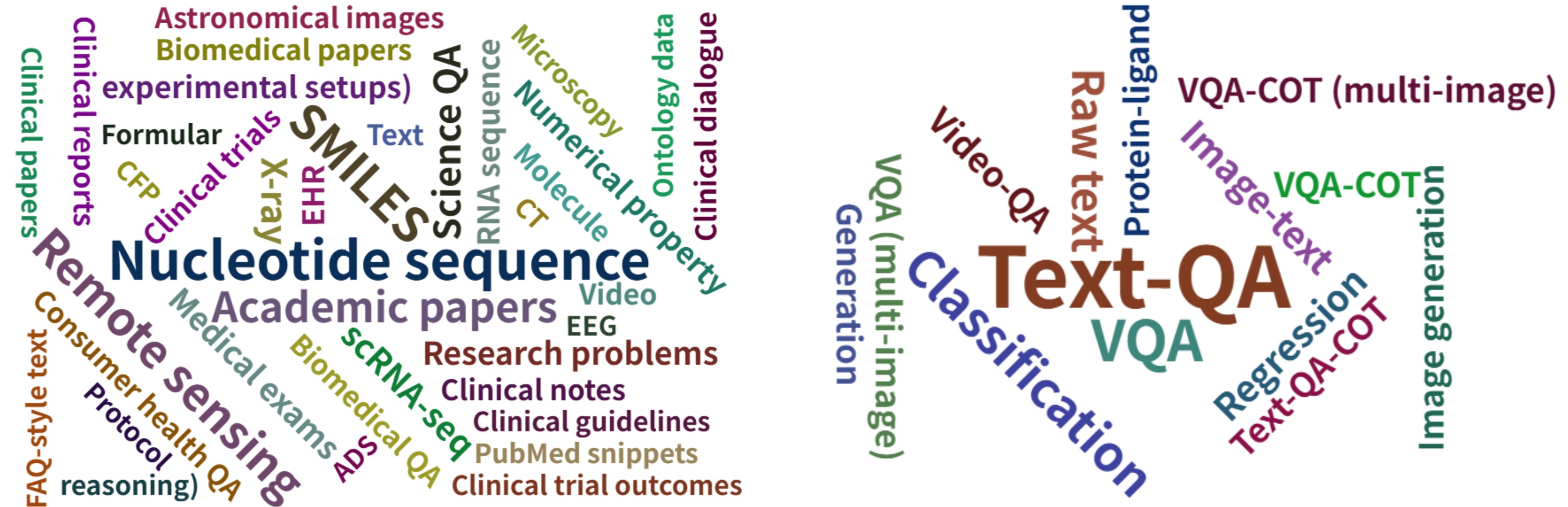}
    \caption{Word clouds of the scientific benchmarks. The plots show the relative distributions of modalities (left) and types (right), with word size proportional to frequency.}
    \label{fig:wordcloud_eval}
\end{figure}

To understand the landscape of scientific evaluation benchmarks, we first examine the distribution of data sources and benchmark characteristics across domains. Fig.~\ref{fig:eval_sources} reveals a striking pattern: most scientific domains rely heavily on a single dominant source type, with academic and research resources dominating in Physics and Chemistry, while Life Sciences shows slightly more diversity. This homogeneity in source materials raises concerns about the robustness and generalizability of current evaluation suites, as models may overfit to specific data types rather than developing broad scientific reasoning capabilities. Fig.~\ref{fig:wordcloud_eval} further illustrates the composition of these benchmarks through word clouds, where the prevalence of text-based QA formats and specific modalities like ``VQA'' and ``Text-QA'' highlights the current emphasis on question-answering paradigms, while revealing gaps in coverage of other important scientific tasks such as hypothesis generation, experimental design, or cross-domain reasoning.

These visualization patterns motivate a deeper analysis of how scientific benchmarks are constructed and what they actually measure. Across recent benchmarks for evaluating Sci-LLMs/MLLMs, we observe several patterns.

\subsubsection{Tiered Regime in Data Generation and Annotation}
Annotation in scientific benchmarks shows a tiered, hybrid regime: manual expert curation anchors quality in hard domains, semi-automated human-in-the-loop pipelines deliver scale with control, and fully automated systems enable extreme throughput where labels can be programmatically derived. Each of them trades off quality, scalability, and resource requirements.


Manual annotation remains prevalent in specialized scientific domains where expert knowledge is crucial~\cite{krithara2023bioasq,He2024OlympiadBencha}. For instance, MicroVQA~\cite{burgess2025microvqa} employs 12 human annotators for microscopy image question-answering, while OmniEarth-Bench~\cite{wang2025omniearth} utilizes over 40 annotators to ensure comprehensive coverage of Earth science domains. 

Semi-automated pipelines balance speed and fidelity by pairing LLM/tooling with expert review: Genome-Bench drafts with GPT-4 models before human checks ~\cite{yin2025genome}; MM-PhyQA blends ChatGPT and scripts with over eight reviewers ~\cite{MMPhyQA}; RP3D-DiagDS couples custom crawlers and GPT-4 with specialist adjudication ~\cite{zheng2024large}. However, recommended practices (\eg, annotator training, pilot studies, iterative refinement) are still too rarely documented.


Fully automated pipelines achieves efficient annotation using established computational frameworks: the Genomics Long Range benchmark synthesize targets from experimental/computational protocols ~\cite{zhou2015predicting,trop2024genomics}; USPTO mines 1.9M patents programmatically~\cite{marco2015uspto}; RSVQA-LRBEN generates million-scale remote-sensing QAs by rule-based analysis of satellite imagery ~\cite{lobry2020rsvqa}. This maximizes coverage and efficiency. However, benchmarks that involve LLMs as auto-annotation tools raise risks of \textit{(i)} circularity and contamination when the same or closely related LLMs are later evaluated on LLM-labeled data, and \textit{(ii)} propagation of potential inaccuracies and biases in LLM-based annotation. Such problems are even harder to review, because LLMs can produce nuanced, plausible, yet erroneous answers \emph{at scale}, which are often difficult to validate without high-level expertise. This highlights the need for careful validation even in automated pipelines~\cite{Qiu2025PHYBencha,hu2023mimicdiffvqa}.

\subsubsection{Skewed Knowledge Level with Increasing Difficulty}

The knowledge level required by the evaluation datasets, \ie, difficulty, is under-specified and skewed. A large fraction of recent datasets do not provide information about their difficulty entirely, typical in integrated or web-mined corpora where provenance is diffuse (\eg, OmniMedVQA~\cite{huOmniMedVQANewLargeScale2024a}, VRSBench~\cite{li2024vrsbench}, SRBench~\cite{Cava2021Contemporary}). Among those that do specify, there is a polarization: high-stakes or research-level resources tag themselves ``Expert'' (\eg, PhysicsArena~\cite{Dai2025PhysicsArena}, LLM4MatBench~\cite{rubungo2024llm4matbenchbenchmarkinglargelanguage}, RP3D-DiagDS~\cite{zheng2024large}, MedXpertQA~\cite{zuoMedXpertQABenchmarkingExpertLevel2025b}) while exam/education-style benchmarks cluster at ``Undergraduate'' (\eg, UGPhysics~\cite{xu2025ugphysics}; PHYSICS~\cite{feng2025physics}, MEDIQA-AnS~\cite{savery2020question}) with very few ``Intermediate'' slices to chart capability boundaries across a continuum~\cite{saikh2022scienceqa}. Cross-sectioning by release date suggests the skew is increasing: 2024–2025 saw a wave of expert-labeled clinical and science sets (MedXpertQA 2025.01; PhysicsArena 2025.05) alongside new Undergraduate exam corpora (UGPhysics 2025.01; PHYSICS 2025.03).


These expert-level benchmarks demand not only deep domain knowledge but also the ability to synthesize information from and reason on multimodal and cross-domain cues. Medical benchmarks particularly exemplify this with requirements of complex reasoning on rare diseases~\cite{chen2024rarebench} and dubious cases~\cite{arora2025healthbench}. Questions in these benchmarks are typically designed to be ``Google-proof''~\cite{Rein2023GPQAa} and entangled, requiring genuine understanding and multi-step thinking~\cite{Zhang2025PhysReasona} rather than simple memorization, setting a particularly high bar for model evaluation. 
The emergence of expert-level benchmarks could be attributed to the need for testing the limits of capability of frontier LLMs, and also reflects growing recognition that scientific reasoning requires not just factual knowledge but the ability to apply, analyze, and create new understanding~\cite{bloom1956taxonomy}.

\subsubsection{Shift towards Domain-Specific Metrics}

In terms of evaluation methods and metrics, question-answering form is the prevailing evaluation, but the metrics are evolving from simple accuracy measurements to sophisticated multi-faceted, domain-specific assessment frameworks, reflecting the heterogeneity of scientific problems. 

Scientific benchmark datasets designed for modern Sci-LLMs typically focus on closed-ended questions (\eg, multiple-choice questions, ``True/False'' problems), where the exact answers can be easily extracted from the outputs of Sci-LLMs using regular expressions; the dominant evaluation metrics are simple and objective: exact match and accuracy. Such a single universal score, however, provides limited insights on the capability of Sci-LLMs, and is difficult to employ in open-ended questions. 
Benchmarks that require natural language generation frequently adopt n-gram overlap (BLEU/ROUGE) to compare free-form outputs against references~\cite{savery2020question,he2020pathvqa}. However, these surface-form metrics do not consider semantic correctness. BERTscore~\cite{zhang2020bertscore}, as employed in some benchmarks~\cite{zhao2025msearth,yan2024clinicallab}, mitigates this problem by comparing the embedding similarity between Sci-LLM's responses and gold answers, yet the semantic similarity still does not guarantee factual correctness and underweights negation and nuanced meanings.

Domain-anchored measures are strongest where the science supplies mature targets: in genomics and multi-omics, AUROC/AUPRC are standard for association and retrieval (\eg, DISEASES~\cite{pletscher2015diseases}, repoDB~\cite{brown2017standard}), while regression tasks~\cite{bogard2019deep} adopt R\textsuperscript{2}, RMSE, or Pearson's correlation coefficients (PCC) to quantify effect-size prediction rather than linguistic plausibility. Chemistry emphasizes chemical validity and drug-likeness for molecular generation, rightly scoring whether molecules are synthesizable and pharmacologically plausible~\cite{irwin2012zinc,gaulton2012chembl}. Physics benchmarks illustrate metric specialization along two axes: exact string/structure match for symbolic regression~\cite{Feynman,Cava2021Contemporary}, which verifies whether a discovered closed-form is the same function, and step-wise or explanation-sensitive grading~\cite{Qiu2025PHYBencha} that penalizes reasoning drift even when final answers coincide. 

The merit of this trend is clear: metrics are increasingly aligned with the scientific target, enabling faithful model selection and revealing failure modes that generic QA accuracy would hide. But there are risks. First, narrow metrics can be gamed (\eg, maximizing BLEU without factual grounding, or optimizing AUROC under pathological class priors). Also, portfolios are inconsistent across datasets, impeding cross-domain comparison. Furthermore, many QA/VQA sets still rely on overlap-based or single-number accuracy for open-ended tasks, under-measuring calibration, citation faithfulness, and harm~\cite{arora2025healthbench}. Looking forward, future scientific benchmarking should \textit{(i)} pair task-native objectives with calibration and uncertainty reporting (\eg, ECE/Brier alongside AUROC for DISEASES~\cite{pletscher2015diseases}); \textit{(ii)} add process-aware scoring that evaluates intermediate steps and evidence use~\cite{Qiu2025PHYBencha}; \textit{(iii)} incorporate reference-grounded factuality/citation checks for text outputs so a model must justify answers beyond n-grams~\cite{savery2020question,he2020pathvqa}; and \textit{(iv)} standardize multi-metric dashboards per domain to avoid metric gaming and improve comparability across releases~\cite{zhan2023rsvg,wang2025xlrs,li2024vrsbench}.

\begin{figure*}[t!]
    \centering
    \includegraphics[width=0.95\linewidth]{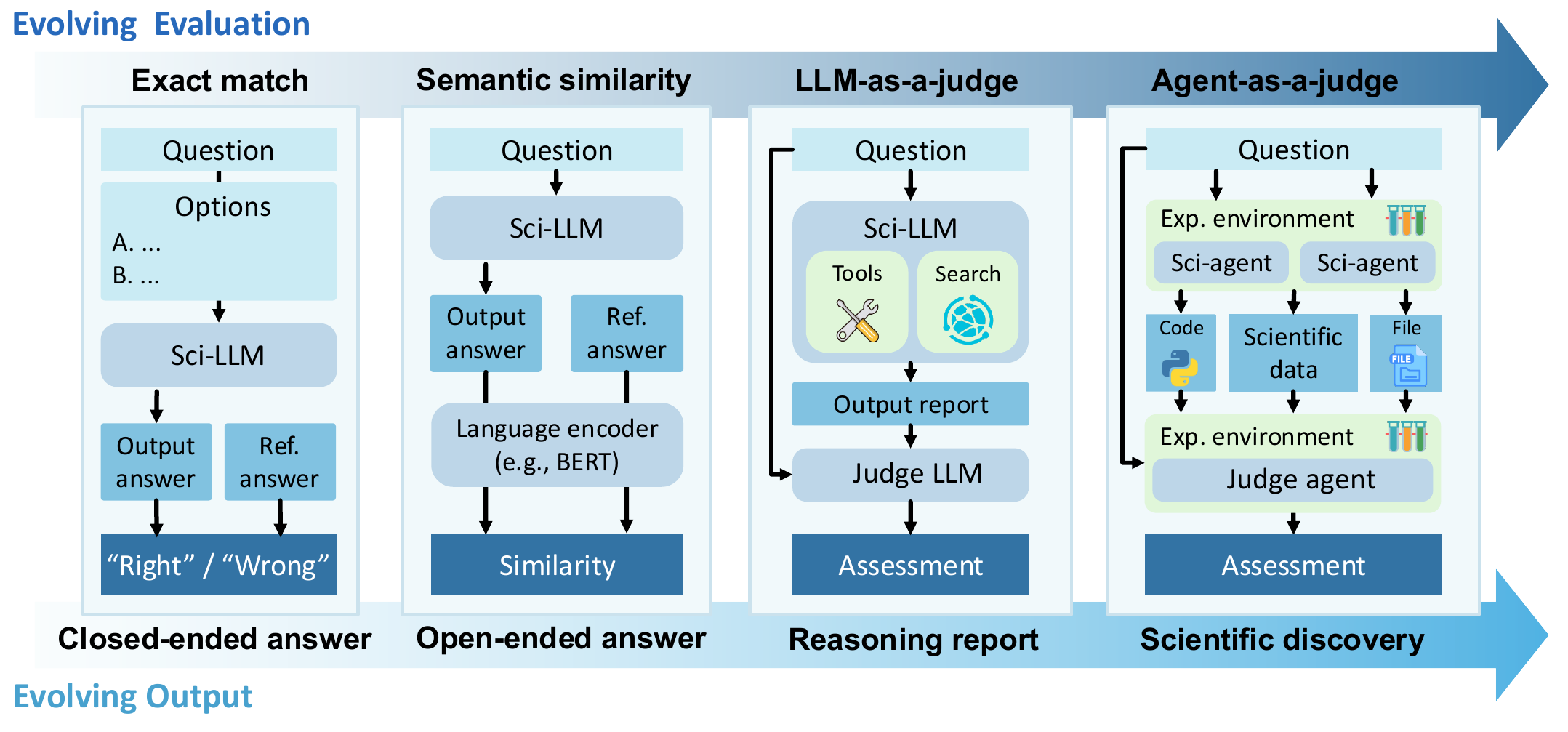}
    \caption{The evolution of evaluation methods for LLMs, starting from simple ``Right or wrong'' exact matches and progressing to semantic similarity comparisons for open-ended answers with metrics like BERT-Score~\cite{zhang2020bertscore}. More advanced methods include using an LLM as a judge to generate reasoning reports, culminating in the use of multiple agents and tools within an experimental environment for scientific discovery to provide a comprehensive model assessment.}
    \label{fig:evaluate_develop}
\end{figure*}



\subsection{LLM / Agent as a Judge}

With the rapid advancements in LLMs and multimodal generative models, traditional evaluation methods, which often rely on a single numerical score (\eg, accuracy) or require extensive manual labor, have become inadequate. To address this challenge, an emerging trend is to use agentic systems to evaluate other agents or models. This ``Agent-as-a-Judge'' paradigm is a natural extension of ``LLM-as-a-Judge''~\cite{gu2024survey} and provides richer, more reliable evaluations by incorporating agentic features like dynamic planning and intermediate feedback (Fig. \ref{fig:evaluate_develop}).

The primary advantages of the ``Agent-as-a-Judge'' framework are its flexibility, efficiency, and explainability. It typically employs a multi-round, dynamically adjusting evaluation process that mimics the strategies of human experts. During this process, the agent judge can dynamically adapt its evaluation direction and test cases based on intermediate results and observed feedback. This approach moves away from a reliance on fixed benchmarks and large sample sizes, significantly reducing the time and computational cost required for evaluation.

For instance, in code generation~\cite{zhuge2024agent}, the agent judge can evaluate a developer agent's performance on multi-step tasks, not just the final outcome. In the domain of visual generation~\cite{zhang2024evaluation}, an evaluation agent can conduct multi-round assessments based on an open-ended user query, ultimately providing a detailed natural language analysis and summary rather than just a simple numerical score. This provides deeper insights into a model's strengths and weaknesses. In the hypothesis generation task~\cite{yang2024moose,yang2025moose}, a judge agent evaluates the novelty, validity, and coverage of key points in the proposed hypotheses, which is well-suited to their inherently flexible and open-ended nature.

This trend has profound implications for future evaluation in the scientific domain, particularly for automated scientific discovery. Automated scientific discovery often involves complex, multi-step tasks where the outcome cannot be easily quantified with a single metric. Traditional evaluation methods are ineffective at capturing the intermediate processes and pinpointing failures within these tasks. The "Agent-as-a-Judge" framework addresses this by providing rich intermediate feedback and a comprehensive analysis of the entire process.

\subsection{Inspiration from Test-Time Learning}

Test-Time Learning (TTL) is gaining significant traction in the natural sciences due to its unique value proposition. First, scientific benchmark evaluation inherently involves working with test sets that lack ground-truth answers, which perfectly fits TTL’s paradigm of adaptation at inference time without requiring labeled data~\cite{hu2025test}. On the other hand, datasets in the natural sciences exhibit strong heterogeneity and distribution shift. For example, Earth sciences encompass atmospheric, oceanic, remote sensing imagery, and textbook text, with large differences in data structure and semantics within each subdomain. Conventional, statically pretrained LLMs often underperform when confronted with data distributions markedly different from their training corpus, whereas TTL enables immediate adaptation by dynamically updating parameters or reasoning strategies using currently observed, unlabeled test samples.

TTL’s practical application in the natural sciences manifests in several technical pathways. MedAdapter~\cite{shi2024medadapter} employs post-hoc adapters for TTL in biomedical applications. Across four biomedical reasoning tasks and eight datasets, the performance of white-box LLMs improved by 18.24\%, while the performance of black-box LLMs improved by 10.96\%. In the field of chemistry, \cite{thomas2025test} proposes scaling test-time training with reinforcement learning for chemical language models to improve chemical space exploration on their proposed benchmark, MolExp, which focuses on discovering structurally diverse molecules with similar bioactivity. Evaluation results on MolExp reveal that extending increasing the TTL will improve model performance, but the performance gains will diminish if the TTL time is too long. In theoretical physics, Gao \etal~\cite{gao2025test} proposed a symbolic weak-verifier framework in TTL to enhance performance on the TPBench~\cite{chung2025theoretical} physics dataset.

\section{Scientific Data Development}
\label{sec:analysis}

This section examines how scientific data influences model development across various stages including data collection, training, and evaluation, highlighting systemic limitations and emerging opportunities. We begin by analyzing the methodologies in scientific data construction (Sec.~\ref{sec:analysis_data}), and then point out critical limitations of current datasets (Sec.~\ref{sec:analysis_char}). Finally, we identify deeper structural issues that hinder the usability of scientific data for LLM development (Sec.~\ref{sec:analysis_system}).

\begin{figure*}[h!]
    \centering
    \includegraphics[width=0.95\linewidth]{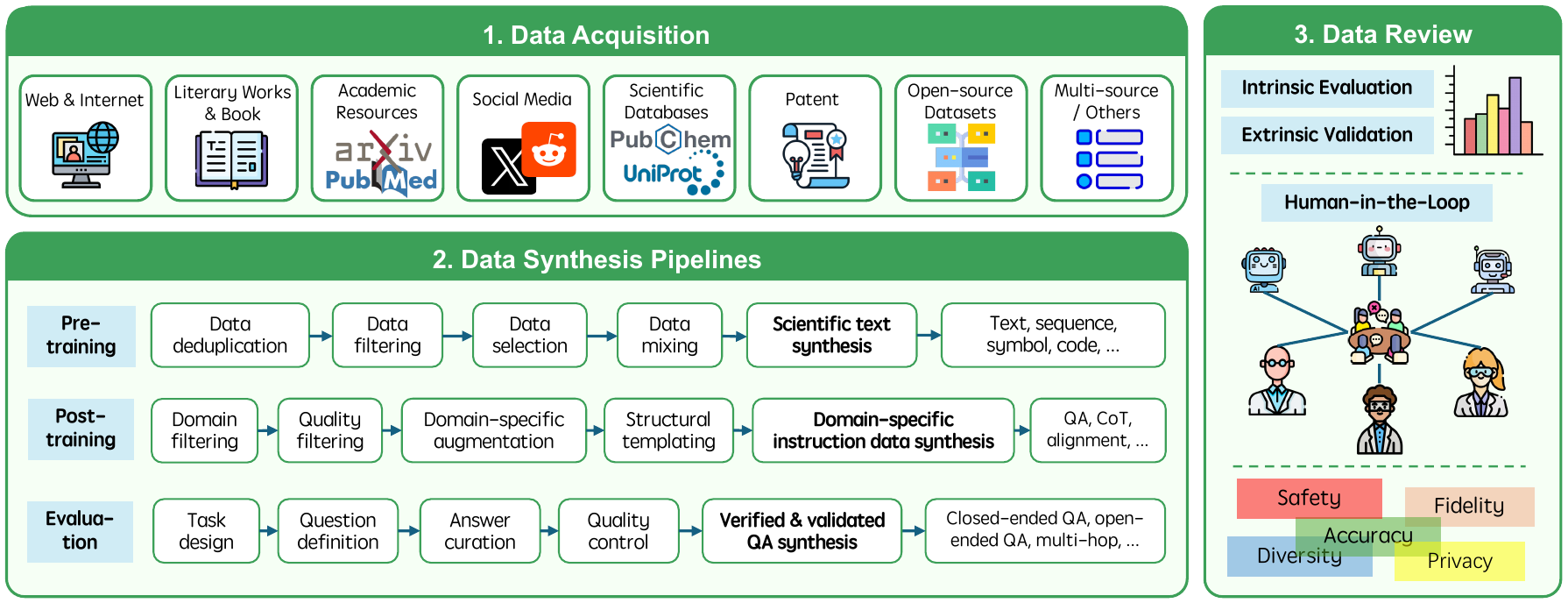}
    \caption{Scientific data construction pipeline: multi-source data acquisition, data synthesis pipelines for pre-training, post-training and evaluation stages, and comprehensive review framework incorporating intrinsic evaluation, extrinsic validation and human-in-the-loop feedback with five quality criteria (safety, fidelity, accuracy, diversity, privacy).}
    \label{fig:data_pipeline}
\end{figure*}

\subsection{Data Collection and Labeling}
\label{sec:analysis_data}

The development of Sci-LLMs fundamentally depends on the quality of their training data; our analysis of existing datasets reveals a complex landscape of acquisition and annotation practices that vary across domains, reflecting both the heterogeneous nature of scientific knowledge and the practical constraints of dataset construction. 
This subsection discusses three aspects that outline the key factors shaping how scientific datasets are constructed and curated for LLM development, including: \emph{(i)} data source heterogeneity and acquisition strategies (Sec.~\ref{sec:analysis_data_heterogeneity}), which describe the diversity of infrastructures and repositories that supply scientific data; \emph{(ii)} annotation methodologies and quality control (Sec.~\ref{sec:analysis_data_annotation}), which address the pipelines and validation processes used to ensure data reliability; and \emph{(iii)} cross-domain patterns and domain-specific considerations (Sec.~\ref{sec:analysis_data_cross}), which highlight recurring challenges such as bias, ethical constraints, and disciplinary practices.

\subsubsection{Data Source Heterogeneity and Acquisition Strategies}
\label{sec:analysis_data_heterogeneity}

The scientific data ecosystem exhibits remarkable diversity in its sources, with each domain developing distinct acquisition strategies tailored to its knowledge infrastructure. Academic and research resources constitute the primary foundation (Figs.~\ref{fig:posttraining_source} and~\ref{fig:eval_sources}), accounting for the majority of datasets across all disciplines. In life sciences, repositories like PubMed Central and specialized databases such as MIMIC-CXR~\cite{johnson2019mimic} provide structured access to millions of medical images and clinical reports. The astronomy domain leverages arXiv extensively, with datasets like AstroLLaMA~\cite{astrollama} utilizing over 300,000 abstracts, while materials science relies heavily on computational databases like the Materials Project and experimental repositories such as USPTO~\cite{marco2015uspto} patents.

This reliance on established scientific infrastructure presents both advantages and limitations. While peer-reviewed sources ensure data quality and scientific validity, they introduce significant temporal delays—publications typically lag behind actual discoveries by months or years, creating what the paper identifies as a "data latency" problem. Moreover, the dominance of English-language sources creates linguistic bias, with Chinese-language datasets primarily confined to healthcare applications like CMB-Exam~\cite{wang2023cmb} and agricultural resources like CROP, despite substantial scientific contributions from non-English speaking regions.

Web-scraped content emerges as a secondary but increasingly important source, particularly for multimodal data. Remote sensing datasets like RS5M~\cite{zhang2024rs5m} aggregate millions of satellite images from online repositories, while medical education platforms contribute to datasets like MedDialog~\cite{he2020meddialog}. However, the quality and reliability of web-sourced data vary considerably, necessitating sophisticated filtering mechanisms. Patent databases represent a unique intersection of scientific and commercial knowledge, particularly valuable in chemistry and materials science, where USPTO provides access to nearly 2 million chemical reactions with detailed experimental procedures often absent from academic publications.

\subsubsection{Annotation Methodologies and Quality Control}
\label{sec:analysis_data_annotation}

The scientific data synthesis employs a sophisticated multi-track pipeline architecture designed to address the distinct requirements of pre-training, post-training, and evaluation phases (Fig.~\ref{fig:data_pipeline}). The pre-training synthesis pipeline begins with data deduplication to eliminate redundancy across heterogeneous sources, followed by quality-based filtering that removes low-value content. Selected data undergoes strategic mixing to ensure balanced representation across scientific domains, creating a diverse foundation for initial model training. This relatively straightforward process prioritizes scale and coverage over precision, establishing broad scientific knowledge bases.

In contrast, the post-training synthesis pipeline implements more stringent quality controls tailored for instruction-following capabilities. Domain-specific filters first categorize content by scientific subdisciplines, after which quality filters apply elevated standards including factual verification and citation validation. The pipeline then enhances underrepresented domains through targeted synthesis and implements structural templates to standardize instruction-response formats. This refined approach ensures that post-training data not only maintains scientific accuracy but also follows consistent patterns that facilitate effective fine-tuning.

The evaluation data synthesis pipeline represents the most rigorous track, beginning with careful task design that spans multiple cognitive levels from basic factual recall to complex multi-step reasoning. Question creation generates diverse query types including multiple-choice questions with scientifically plausible distractors, open-ended problems requiring detailed explanations, and multi-hop challenges that test reasoning capabilities. Each answer undergoes meticulous construction with step-by-step derivations and comprehensive explanations, followed by multi-round quality assurance to validate both scientific accuracy and logical coherence.

These pipelines produce three distinct categories of synthesized data. Instruction-response pairs encompass sequence-based formats for procedural knowledge, symbol-based representations for mathematical and chemical notations, and code implementations for computational tasks. Knowledge and QA pairs include both alignment data for factual grounding and chain-of-thought examples that demonstrate explicit reasoning processes. Open-ended QA pairs, primarily used for evaluation, feature both multiple-choice questions and complex problems requiring detailed explanations.

The synthesized evaluation data undergoes comprehensive human-in-the-loop review across six critical dimensions. Safety checks ensure no harmful scientific misinformation, while accuracy validation verifies factual correctness against authoritative sources. Diversity assessment confirms broad coverage across subdomains and question types, and fidelity review maintains consistency with established scientific principles. Privacy screening removes any personally identifiable information, and throughout this process, domain experts provide iterative feedback to refine data quality. This rigorous validation framework proves essential for evaluation datasets, as they serve as definitive benchmarks for assessing model capabilities in scientific reasoning and knowledge application.

\subsubsection{Cross-Domain Patterns and Domain-Specific Considerations}
\label{sec:analysis_data_cross}

Despite domain-specific variations, several patterns emerge across scientific data collection efforts. The transition from individual datasets to integrated ecosystems characterizes modern approaches, with initiatives like GMAI-VL~\cite{li2025gmaivlgmaivl55mlarge} in healthcare aggregating 5.5 million multimodal examples across institutions. This consolidation addresses fragmentation but introduces new challenges in maintaining provenance and ensuring consistent quality standards across heterogeneous sources.

Domain expertise requirements create natural barriers to cross-disciplinary data sharing. Medical datasets require understanding of clinical workflows and regulatory constraints, while astronomical data demands familiarity with coordinate systems and instrumental calibrations. Agriculture occupies a unique position, requiring integration of biological knowledge with environmental monitoring, resulting in datasets like MIRAGE~\cite{dongre2025mirage} that combine expert agricultural consultations with field imagery.

There is a concerning trend toward annotation convenience rather than scientific completeness. Datasets often reflect what is easily accessible rather than what is scientifically important—published positive results dominate while negative findings remain largely absent. This bias extends to experimental conditions, with datasets capturing idealized scenarios rather than the messy reality of scientific practice. Materials science datasets focus on computationally generated structures while experimental synthesis failures go unrecorded, creating an incomplete picture of the scientific process.

Privacy and ethical considerations impose additional constraints, particularly in life sciences. While physics and astronomy data are generally open, medical datasets require extensive de-identification and access controls. This creates a fundamental tension between data availability and patient protection, resulting in geographic and demographic biases as datasets predominantly originate from well-resourced institutions in developed countries. Agricultural datasets face similar challenges with proprietary farming data, limiting the diversity of crop varieties and growing conditions represented in publicly available resources.

\subsection{Limitations of Current Scientific Datasets}
\label{sec:analysis_char}

Despite rapid growth in scientific corpora, current datasets exhibit significant limitations in scope, granularity, and modality coverage. This subsection characterizes fundamental challenges that constrain the training and evaluation of Sci-LLMs, including: \emph{(i)} the scarcity of experimental data (Sec.~\ref{sec:analysis_char_scarcity}), which arises from the high cost of data acquisition and the rarity of scientific phenomena; \emph{(ii)} the over-reliance on text modality data (Sec.~\ref{sec:analysis_char_reliance}), which limits multimodal reasoning and reduces empirical grounding; \emph{(iii)} the representation gap between static knowledge and dynamic processes (Sec.~\ref{sec:analysis_char_representation}), showing how current datasets fail to capture the evolving nature of scientific inquiry; and finally, \emph{(iv)} the multi-level biases (Sec.~\ref{sec:analysis_char_bias}) that stem from publication practices, language dominance, and domain skew, all of which impact the fairness and generalizability of Sci-LLMs. 

\subsubsection{Scarcity of Experimental Data}
\label{sec:analysis_char_scarcity}
The scarcity of experimental data in scientific domains stems from several inherent characteristics of scientific data. These factors collectively hinder the development of data-intensive scientific LLMs and MLLMs. The first characteristic is the high acquisition cost in experimental data generation.
Scientific experimentation is often extraordinarily expensive and time-consuming. Experimental research frequently faces significant financial constraints that cause sufficient experiments to yield statistically reliable results. For instance, in drug discovery, obtaining accurate protein structures is essential for understanding molecular interactions, but it requires costly wet-lab experiments and specialized equipment like cryo-electron microscopes, X-ray crystallography. 
Similarly, generating high-fidelity simulation data~\cite{vogelsberger2014introducing, sliwoski2014computational, karplus2002molecular}, which can serve as a proxy for experimental data in scientific machine learning, typically  demands substantial computational resources and long processing time to generate datasets of adequate size. This inherent financial and temporal burden directly restricts the scale and diversity of   experimental datasets. The traditional pace of scientific investigation, constrained by these resource limitations, often struggles to match the  data demands of modern AI models, creating a fundamental bottleneck. In healthcare, access to clinical data usually requires rigorous ethical review and carries privacy risks, constraining their widespread availability and scalability.
Another inherent unique challenge causing the data scarcity is the rarity of specific scientific phenomena. 
Unlike other forms of scarcity that might be mitigated through increased resources or improved collection methods, this type of scarcity is intrinsic to the natural world or specific experimental conditions. For example, in healthcare, research into rare diseases is perpetually hampered by the limited availability of patient data, directly impeding the development of effective treatments and diagnostic tools. This means that AI models designed for these domains must be capable of learning effectively from extremely limited examples, as the underlying phenomena themselves are inherently infrequent.
The lack of AI-ready experimental data is another key challenge in building effective scientific LLM models. Experimental data in the natural sciences suffer from heterogeneity and the lack of standardization~\cite{griffiths2021dataset}, as they come from diverse instruments, protocols, and domains, each with its own formats, units, and conventions. Without community-adopted standards for data schemas and metadata fields, integrating datasets across labs or domains becomes a labor-intensive and error-prone task. As a result, crucial contextual information (\eg, experimental conditions, calibration details) are often omitted or encoded inconsistently, forcing AI practitioners to spend disproportionate effort on data preprocessing rather than model development.

\subsubsection{Over-reliance on Text Modality Data}
\label{sec:analysis_char_reliance}

Current scientific corpora for LLMs and MLLMs rely heavily on published articles, patents, and reviews, which are rich in descriptive content but poor in raw experimental detail~\cite{taylor2022galactica, prabhakar2025omniscience, zhang2024sciglm, muennighoff2025s1}. This over-reliance on the text modality introduces several issues.
First, scientific datasets tend to prioritize aggregated summaries over raw measurements, leading to limited quantitative depth. Textual reports often present averaged results without revealing underlying data distributions. Consequently, models are never exposed to the full variability of experimental outcomes, limiting their capacity to reason about uncertainty or discern fine-grained trends.
Second, text-based scientific literature often exhibits selection and reporting bias. Authors typically highlight statistically significant or positive findings, while omitting negative results or methodological failures. This causes a skewed perception of science as a linear and uniformly successful process.
Beyond textual limitations, current scientific datasets suffer from a scarcity of structured experimental data, as detailed in Sec.~\ref{sec:analysis_char_scarcity}. Machine-readable protocols, equipment settings, and raw time-series measurements are rarely shared in standardized formats~\cite{huang2025perceptual, wilkinson2016fair}. Without detailed reagent tables, step-by-step procedures, or high-resolution simulation outputs, models cannot infer the precise cause-effect relationships that drive scientific discovery.
Moreover, many key scientific modalities are either excluded or available only as low-resolution figures embedded in PDFs. These include spectra, microscopy images, chromatography traces, and raw sensor streams. Without high-quality multimodal signals, MLLMs lack the empirical grounding to connect textual hypotheses with experimental evidence.
Overall, the imbalance between descriptive text and scientific modality data severely limits a model’s ability to generalize from narrative summaries to the rigorous, data-driven reasoning required in cutting-edge research. Bridging this gap will require more complete, structured, and multimodal experimental datasets.

\subsubsection{Representation Gap between Static Knowledge and Dynamic Processes}
\label{sec:analysis_char_representation}

Scientific datasets usually provide static snapshots of knowledge at the time of collection, which fails to reflect the continuously evolving nature of scientific discovery. In contrast, scientific progress is a iterative cycle of formulating hypotheses, testing them against emerging data, and refining through continuous experimentation and analysis. This mismatch between static data and the dynamic research process creates a significant representation gap: models trained on these one-off datasets struggle to make reliable predictions or conduct meaningful reasoning about evolving phenomena.
The gap is particularly pronounced in observational records, experimental results, and scientific QA benchmarks that often rely on predetermined question–answer pairs   from published sources. The static nature of these collections leads to ``knowledge expiration'' as new findings emerge, thereby undermining their relevance and validity. As facts change, models trained on these snapshots may yield outdated or even contradictory conclusions, which impedes their utility for real-time reasoning and hypothesis generation that requires up-to-date evidence and iterative feedback.


\subsubsection{Multi-level Biases in Scientific Datasets}
\label{sec:analysis_char_bias}

Scientific datasets contain systematic biases that embed skewed perspectives into the training of LLMs. 
These biases arise when data deviates from a comprehensive scientific reality,  including publication bias, domain bias, author and institutional biases. Understanding these biases is the crucial step toward building fairer and more accurate AI models.
%
%
Publication bias leads to an overabundance of positive results, as studies with statistically significant findings are up to three times more likely to be published than those with null results ~\cite{dwan2013systematic}. This dismissal of negative or inconclusive data distorts the available evidence. Language bias reinforces the dominance of English, as English-language publications make up the vast majority of accessible scientific literature~\cite{tardy2004role}. This causes models to misrepresent or underperform on scientific work from other languages and cultural contexts.
Pervasive domain bias exists in repositories such as PubMed, which disproportionately focus on the life sciences and biomedicine, while underrepresenting disciplines like physics, chemistry, and social sciences.  This impairs the ability of LLMs to generalize across scientific domains. 
Finally, author and institutional biases emerge when a small number of prolific researchers or elite institutions contribute disproportionately. This phenomenon imprints specific writing styles and thematic focuses, causing models to mirror dominant voices rather than reflect the full diversity of scientific discourse.
Addressing these systematic biases through corpus diversification, targeted augmentation of underrepresented domains, and bias-aware sampling is essential for building fairer and more reliable scientific LLMs.




\subsection{Systematic Issues in Data Quality}
\label{sec:analysis_system}

Beyond surface-level limitations, the scientific data ecosystem suffers from systemic issues that undermine the data-driven scientific AI. This subsection highlights three critical areas that must be addressed to support robust Sci-LLM development. First, we describe the data traceability crisis (Sec.~\ref{sec:analysis_system_trace}), where missing provenance and undocumented preprocessing hinder reproducibility and trust. Next, we explore scientific data latency (Sec.~\ref{sec:analysis_system_latency}), which delays the incorporation of recent discoveries into model training and limits real-time scientific reasoning. Finally, we focus on the lack of AI-readiness (Sec.~\ref{sec:analysis_system_readiness}), emphasizing how poor formatting, missing metadata, and domain-specific heterogeneity prevent many datasets from being directly used in LLM pipelines. These structural deficiencies highlight the need for end-to-end redesign of scientific data practices, enabling continuous, traceable, and AI-compatible knowledge integration.

\subsubsection{Data Traceability Crisis}
\label{sec:analysis_system_trace}

The scientific data traceability crisis in building LLMs and MLLMs for various science domains poses a significant challenge to the integrity and utility of AI-driven scientific discovery. 
The data traceability crisis stems from inconsistent, incomplete, and often undocumented management of the diverse scientific datasets used to train these complex models.
The metadata of scientific datasets describing sample provenance, processing details and versioning information are often sparse or missing.
Fundamentally, this deficiency in transparency and auditability may undermine scientific rigor and reproducibility. Subsequent researchers struggle to reconstruct how scientific data are generated and transformed. 
It exacerbates existing problems such as bias propagation, introduces considerable legal and ethical liabilities, and complicates the crucial process of validating AI-generated scientific hypotheses. 
Also, there is increasing difficulty in distinguishing synthetic from real experimental data. 
Recent analyses show systematic under-utilization of roughly three-quarters of online data repositories, largely due to insufficient data traceability~\cite{graziani2025we}. 
The cumulative effect could diminish the trust in AI systems, particularly within high-stakes scientific applications ranging from novel drug discovery to precise medical diagnostics. 
Addressing this issue necessitates a comprehensive strategy that integrates advanced technological solutions with robust data governance frameworks, clear regulatory guidelines, and a sustained commitment to fostering greater transparency and accountability throughout the AI development lifecycle.

\subsubsection{Scientific Data Latency}
\label{sec:analysis_system_latency}

Scientific data latency refers to the delay between when new experimental results, publications, or datasets are generated and when they become available for a scientific LLM to ingest.
This latency issue undermines model accuracy, reliability, and relevance, particularly in fast-evolving fields such as biomedicine, climate science, and materials science, where new discoveries can quickly render older information obsolete. 
The data latency issue arises from several aspects.
First, many scientific findings appear only after lengthy peer‐review and publication processes with datasets remain inaccessible, delaying their inclusion in model training. 
Second, even publicly released data often lack standardized metadata or real‐time update mechanisms, causing models to train on out‐of‐date versions of datasets. 
Third, high‐throughput instruments and simulation platforms can produce terabytes of data daily, but bandwidth constraints, quality‐control pipelines, and manual curation introduce additional lags before data are transformed into machine‐readable formats. 
As a result, scientific AI models may perpetuate outdated knowledge, overlook the latest experimental protocols or discoveries, leading to increased risk of hallucination when faced with unfamiliar recent developments. 
Addressing data latency requires the adoption of open‐access policies and development metadata standards to enable automatic updates to training corpora.


\subsubsection{The Lack of AI-readiness} 
\label{sec:analysis_system_readiness}

In the era of scientific AI, scientific data needs to be readily consumable by AI models, seamlessly integrating into their training and inference processes to support automated and scalable scientific discovery.
Despite their immense potential, many scientific datasets are underutilized due to their lack of AI-readiness, posing significant challenges for scientific LLM development. This incompatibility issue stems from incomplete essential metadata, insufficient preprocessing, mismatched structures, and the inherent complexities of diverse scientific information, making direct utilization for model training difficult. Such limitations impede immediate usability, forcing researchers to invest substantial effort in data adaptation rather than accelerating LLM-driven scientific discovery. The majority of published scientific data require extensive preprocessing, curation and enrichment before they become AI‑ready, significantly slowing down progress in building domain‑specialized LLMs and other data‑driven scientific tools. To bridge this gap, the scientific community must shift from simply making data available to ensuring it is truly actionable.

\section{New Paradigms for Data-Driven Sci-LLMs}
\label{sec:future}

New paradigms are emerging that reimagines Sci-LLMs not just as passive predictors but as active, goal-directed systems, \ie, agents, capable of autonomy, interactivity, and orchestration across tools and tasks~\cite{gao2024empowering}.
This section explores two major shifts shaping the future of Sci-LLMs. First, we examine the emergence of scientific agents (Sec.~\ref{sec:future_agent}), which transform Sci-LLMs into autonomous entities that emphasize planning, experimenting, and self-improving. Then, we analyze how data ecosystems for Sci-LLMs must be redesigned to support these agents (Sec.~\ref{sec:future_ecosys}). 

\subsection{Scientific Agent}
\label{sec:future_agent}

A key paradigm shift is treating LLMs as scientific agents that can plan and execute research tasks with a degree of autonomy. This subsection introduces key developments in this direction, beginning with a brief introduction on the transition from Sci-LLMs to scientific agents (Sec.~\ref{sec:future_agent_llm}), followed by the concept of multi-agent collaboration (Sec.\ref{sec:future_agent_multiagent}). Next, we explore the integration of external tools (Sec.~\ref{sec:future_agent_tool}), which enable agents to interact with databases, software, and real-world systems. We also discuss self-evolving agents (Sec.~\ref{sec:future_agent_evolve}) that refine their skills, prompts, and tool usage through iterative feedback. Then, we highlight emerging evaluation frameworks and benchmarks (Sec.~\ref{sec:future_agent_eval}) that rigorously assess agents on end-to-end workflows, collaboration, and safety in scientific tasks. Finally, we introduce the application of scientific agents on autonomous scientific discovery (Sec.~\ref{sec:future_agent_eval}).

\subsubsection{LLMs as Scientific Agents}
\label{sec:future_agent_llm}

Rather than simple question-answering, a scientific LLM agent is given high-level goals (\eg, ``discover potential drug candidates for disease X'') and autonomously decomposes the task, gathers information, performs experiments (virtually), and synthesizes results~\cite{ren2025towards}. These agents maintain structured, hypothesis-driven workflows that echo the scientific method: defining hypotheses, selecting experimental methods, and validating results before drawing conclusions. Crucially, they emphasize reproducibility and scientific rigor, incorporating domain-specific constraints and verification steps that generic AI assistants often lack. Studies have highlighted that accelerating discovery requires capabilities beyond generic chatbots – for instance, generating novel hypotheses, designing and running experiments, and interpreting complex data in context~\cite{huang2025biomni,boiko2023autonomous}. By building these capabilities, LLM-based scientific agents aim to serve as AI co-researchers that can handle tedious or complex aspects of research, allowing human scientists to focus on creativity and high-level decisions.

\subsubsection{Multi-Agent Collaboration}
\label{sec:future_agent_multiagent}

Recent scientific agents have shifted from single monolithic planners to structured teams that reflect real laboratory roles and social dynamics~\cite{luo2025largeagent}. The Virtual Lab~\cite{swanson2025virtual} organizes a principal-investigator agent and specialist scientist agents into recurring ``research meetings,'' demonstrating end-to-end design of SARS-CoV-2 nanobodies and validating wet-lab outcomes; the setting formalizes division of labor, critique, and iteration, and reports meaningful human-in-the-loop oversight while preserving agent autonomy. VIRSCI~\cite{su2024many} models team formation explicitly for idea generation, showing that diversified agent roles and controlled disagreement increase novelty without sacrificing feasibility. PiFlow~\cite{pu2025piflow} adds principle-aware collaboration for hypothesis refinement by constraining agent proposals with physical/biological priors to reduce aimless exploration, a common failure mode in free-form multi-agent pipelines. At the system level, Agent Laboratory~\cite{schmidgall2025agent} frames an entire ``paper-production'' pipeline—problem scoping, method selection, execution, analysis, and writing—via cooperating agents with persistent artifacts and audit trails. For embodied science, ChemAgents~\cite{song2025multiagent} deploy a hierarchical multi-agent controller onboard a robotic chemist to coordinate experiment planning, execution, and self-correction across hardware and simulation. Beyond homogeneous LLM teams, hybrid collectives of agents and humans (\eg, steering committees) have become standard, with explicit critique-and-revision loops and role-switching when agents detect stale priors or tool failures~\cite{huang2025deepresearchagents}. 

Empirically, multi-agent settings yield the largest gains when: (1) roles are capability-aligned (planner, critic, executor); (2) communication channels are structured (RFA templates, meeting minutes, explicit ``claim–evidence'' schemas); and (3) conflict resolution is formalized (voting, debate, or auctioning). Science-centric multi-agent benchmarks, \eg, MultiAgentBench~\cite{zhu2025multiagentbench} for coordination/competition and communicative multimodal tasks, now make such interaction skills measurable.

\subsubsection{Tool Use}
\label{sec:future_agent_tool}

A defining feature of scientific LLM agents is their heavy integration with external tools and data resources~\cite{ren2025towards}. SciToolAgent~\cite{ding2025scitoolagentknowledgegraphdrivenscientific} organizes hundreds of domain tools via a knowledge-graph of capabilities, preconditions, and I/O signatures; the graph enables retrieval-augmented tool selection, multi-hop sequencing, and fault-aware backoff across several domains. It reports consistent gains over vanilla tool-calling baselines on curated scientific workflows and adds policy checks for responsible use. Biomni~\cite{huang2025biomni} exemplifies a domain-scale agent that interfaces with 150 tools, 59 databases, and 105 software packages to automate biomedical analyses end-to-end, emphasizing reproducibility and provenance. Under the hood, modern stacks increasingly adopt the Model Context Protocol (MCP)~\cite{anthropic2024mcp} to standardize tool discovery, authorization, and invocation, reducing ``glue code'' and enabling safer cross-vendor orchestration; MCP also clarifies user consent and credentials for tools that execute code or reach sensitive data. For web-facing evidence gathering, computer-using and browser-control agents~\cite{openai2025cua} have matured from ad hoc headless scripts to trained GUI/web agents that read, click, and upload files, with reinforcement learning on screen traces; these unlock literature mining, data extraction, and online lab logistics but raise security issues (\eg, prompt-injection, DOM-mismatch), motivating sandboxes and allowlists~\cite{mudryi2025hidden}. For workflow synthesis, WorkflowLLM and WorkflowBench~\cite{fan2024workflowllm} explicitly evaluate whether an agent can translate natural-language protocols into executable API graphs and recover from tool failures; results indicate that specialized workflow-tuned models can outperform general LLMs even with in-context learning. Overall, the state of the art combines: symbolic resource models (capability graphs, ontologies), standardized tool transport, execution sandboxes (containers, rate caps), and reflective monitors that detect hallucinated tools or unsafe parameterizations before launch.

\subsubsection{Self-evolving Agents}
\label{sec:future_agent_evolve}

Self-evolving agents extend scientific LLM agents by adding continual adaptation loops to the standard plan–experiment–verify workflow, so the agent improves itself over time, not just the artifact. Intra-test-time, agents externalize feedback and update episodic memory or prompts to correct future trials, boosting sequential decision making and coding without weight updates~\cite{shinn2023reflexion,madaan2023selfrefine}. Agents also accumulate executable skills and even create tools: Voyager~\cite{wang2024voyager} builds a growing library of programs plus an automatic curriculum that transfers to new worlds. 
Inter-test-time, agents update their models via self-generated supervision~\cite{zelikman2022star,yuan2024selfrewarding}. Further, prompts and tool-use policies can be evolved automatically~\cite{guo2024evoprompt}. For example, Toolformer~\cite{schick2023toolformer} demonstrates self-supervised acquisition of API-calling skills that persist across tasks. Together, these mechanisms instantiate agents that learn from experience, expand capabilities, and reduce brittleness over long horizons.

In scientific fields, self-evolving agents hold the great potential to continually refine hypotheses, protocols, and tool-use policies from experimental and literature feedback, rather than remaining fixed. 
In biomedicine, STELLA~\cite{jin2025stella} couples an evolving ``Template Library'' with a dynamic ``Tool Ocean,'' where a Tool-Creation agent autonomously discovers and integrates new bioinformatics tools; the system’s accuracy on biomedical benchmarks rises as it accumulates trials, evidencing intra- and inter-task self-improvement. OriGene~\cite{zhang2025origene} instantiates a self-evolving virtual disease biologist: specialized agents refine thinking templates, tool composition, and analytic protocols using human and wet-lab feedback, and the framework generated targets (\eg, GPR160 for liver cancer) that were experimentally validated in patient-derived models. In chemistry, ChemAgent~\cite{tang2025chemagent} maintains a self-updating library that decomposes problems and reuses refined solutions, yielding large gains on SciBench~\cite{scibench} and pointing to drug- and materials-discovery use cases. 
However, scientific agents that reliably self-evolve across long horizons with closed-loop laboratory validation remain rare today and an important next step in AI-driven scientific discovery~\cite{gottweis2025towards}.
%


\subsubsection{Evaluation Frameworks and Benchmarking}
\label{sec:future_agent_eval}

Evaluation has shifted from single-turn QA to long-horizon scientific workflows with verifiable endpoints. ScienceAgentBench~\cite{chen2024scienceagentbench} decomposes 102 real tasks from peer-reviewed papers across four disciplines into executable subtasks with gold pipelines, expert validation, and containerized harnesses; despite multiple attempts, the best agents solved only about a third of tasks, highlighting large headroom and the need for tool mastery and code debugging. CURIE~\cite{cui2025curie} stresses long-context scientific reasoning and information extraction across six domains with expert-curated problems, pushing agents to manage citations, units, experimental conditions, and cross-figure synthesis. DiscoveryWorld~\cite{jansen2024discoveryworld} provides a simulated environment that supports end-to-end discovery, including hypothesis formation, experiment design, measurement, and model revision, while automatically scoring task completion, action relevance, and discovered knowledge to enable repeatable testing without wet-lab costs. Auto-Bench~\cite{chen2025auto} targets causal discovery and hypothesis testing, rewarding agents for uncovering latent structure and justifying interventions. WorkflowBench~\cite{fan2024workflowllm} measures orchestration quality using code-level metrics (\eg, CodeBLEU, pass rates) for converting natural instructions into robust API workflows. For collaboration, MultiAgentBench~\cite{zhu2025multiagentbench} and communicative multimodal suites~\cite{ossowski2024comma} quantify coordination, negotiation, and information-sharing when agents have asymmetric views. Cross-cutting surveys~\cite{mohammadi2025evaluation} now standardize taxonomies of what-to-evaluate (capability, reliability, safety) and how-to-evaluate (interaction modes, datasets, metrics, tooling), and call for third-party harnesses, leakage controls, and safety red-teaming specific to agents with execution privileges. Emerging best practices include: containerized runners; seeded randomness and pass@k for robustness; provenance logging; leakage audits for data-contaminated facts; and safety checks for tool scopes, credentials, and network access.

\subsubsection{Autonomous Scientific Discovery}
\label{sec:future_agent_app}

Autonomous scientific discovery represents a transformative paradigm using LLMs~\cite{bai2025qwen2, wang2024qwen2, guo2025deepseek, luo2025llm4sr, zheng2025automation} and robotics to conduct scientific research independently without direct human intervention~\cite{yuan2025dolphin, yan2025surveyforge, gottweis2025towards, yamada2025ai}. By automating critical research tasks including data analysis, hypothesis generation, experiment design, and result interpretation, these automated systems efficiently process vast amounts of information and uncover patterns that elude human researchers~\cite{yuan2025dolphin,lu2024ai}.

Chemistry has seen rapid progress with LLM-tool hybrids that couple symbolic planners with domain utilities. A representative milestone is \emph{Coscientist}~\cite{boiko2023autonomous}, which combined GPT-4 planning with code execution and instrument control in a cloud laboratory to autonomously design, run, and analyze multistep chemistry experiments, including protocol synthesis, hardware documentation navigation, liquid-handling control, and data-driven optimization. 
\emph{ChemCrow}~\cite{bran2023chemcrow} integrated GPT-4 with expert-designed chemistry tools, demonstrating end-to-end tasking from retrosynthesis and catalyst design to guiding discovery of new chromophores, with expert evaluation showing substantial gains over base models. In the life sciences, agentic LLMs are beginning to automate experimental design logic. \emph{CRISPR-GPT}~\cite{qu2025crisprgpt} illustrates how domain knowledge and tool use can turn free-form language reasoning into executable gene-editing workflows, chaining literature-grounded analysis with constraint-aware proposal, delivery recommendations, and validation planning. 

Materials discovery provides a complementary proving ground where scientific agents orchestrate in silico design loops and prepare hand-offs to self-driving labs. \emph{LLMatDesign}~\cite{jia2024llmatdesign} shows that reflective agentic loops can translate high-level targets into candidate materials, invoke calculators for property estimation, and iteratively refine compositions in low-data regimes. At the systems level, emerging frameworks~\cite{singh2025generalized} aim to standardize the interface between agentic planning and autonomous experimentation platforms, highlighting patterns for task specification, data management, and safety interlocks that generalize across domains. 

Despite these promising results, autonomous scientific discovery faces significant challenges in two aspects. (i) Generating proposals that balance scientific validity with genuine novelty requires systems to identify research gaps and formulate innovative hypotheses while maintaining scientific rigor, a task complicated by AI models' reliance on existing data patterns. (ii) Implementing closed-loop feedback for end-to-end experimental validation demands seamless integration across multiple domains, from robotics for experiment execution to advanced analytics for result interpretation, while adapting to real-world experimental uncertainties. Recent developments such as InternAgent~\cite{team2025novelseek} demonstrate progress in addressing these challenges through integrated pipelines that span from idea generation to experimental validation, achieving notable improvements in tasks like reaction yield prediction and enhancer activity prediction within significantly reduced timeframes compared to traditional human-led research.

\begin{figure*}[t!]
    \centering
    \includegraphics[width=0.95\linewidth]{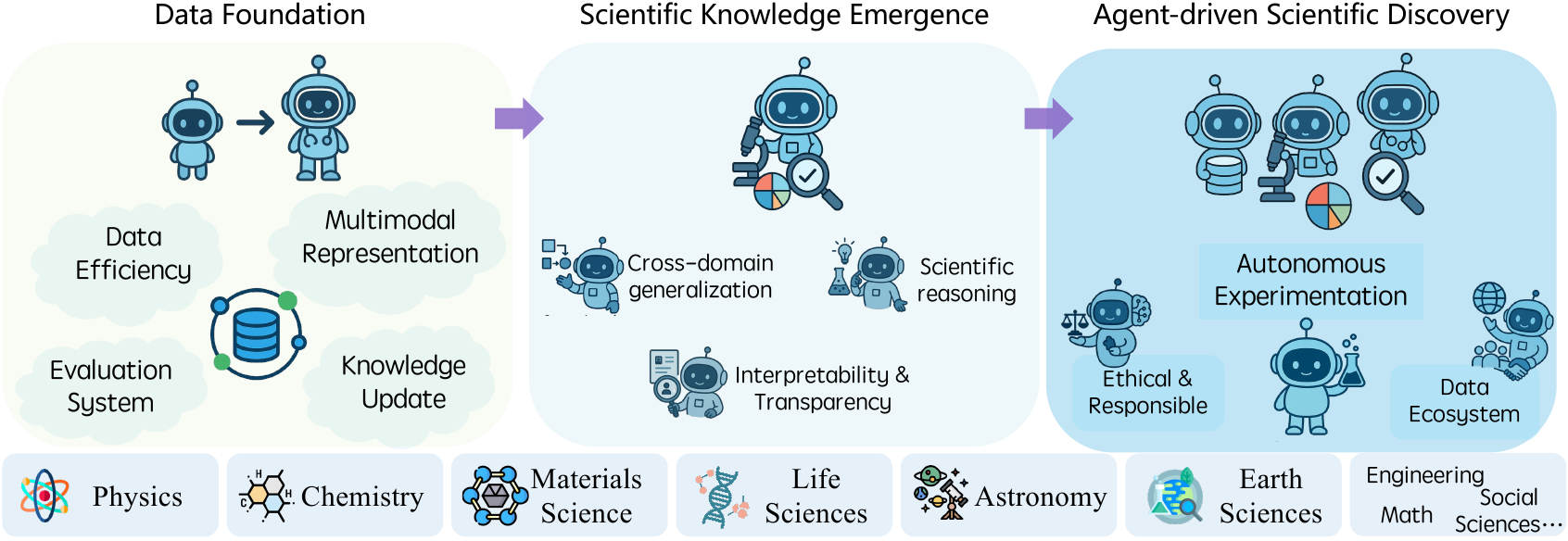}
    \caption{From data infrastructure to agent-assisted discovery: A three-stage evolution of AI in scientific research. This figure delineates the incremental evolution of data-driven Sci-LLMs: (i) Stage I establishes foundational data infrastructure with capabilities in efficiency, multimodal representation, and knowledge updating; (ii) Stage II demonstrates the emergence of scientific capabilities in LLMs driven by mature data ecosystems, enabling cross-domain generalization and scientific reasoning; (iii) Stage III envisions autonomous AI agents that assist scientific discovery while creating closed-loop feedback with data ecosystems, a prospective paradigm for self-evolving discovery systems. This evolution, currently manifesting across physics, chemistry, life sciences, and other domains, illustrates both realized achievements and the expanding potential for AI-driven research as these technologies proliferate into broader scientific disciplines.
    }
    \label{fig:data_agent_future}
\end{figure*}

\subsection{Data Ecosystems for Sci-LLMs}
\label{sec:future_ecosys}

While scientific agents, exemplified by systems like ChemCrow~\cite{m2024augmenting} and Biomni~\cite{huang2025biomni}, independently perform complex scientific tasks, they require an equally advanced \emph{data ecosystem} to truly thrive. 
This subsection outlines how data ecosystems must evolve to support autonomous, tool-using Sci-LLMs. Fig.~\ref{fig:data_agent_future} depicts this evolution: current data foundations have enabled the emergence of scientific knowledge capabilities in LLMs (Stages I-II), while the transition to agent-driven discovery (Stage III) necessitates reciprocal development of data ecosystems to establish closed-loop feedback between autonomous experimentation and data infrastructure. We first provide an analysis of the bottlenecks behind the rise of scientific agents (Sec.~\ref{sec:future_ecosys_bottleneck}), and then introduce the concept of an operating system-level interaction protocol (Sec.~\ref{sec:future_ecosys_os}). We propose design principles for next-generation scientific data architecture (Sec.~\ref{sec:future_ecosys_design}), laying the foundation for a closed-loop system of machine-led scientific inquiry. Finally, we discuss a sustainable data sharing protocols that may benefit the AI4Science community (Sec.~\ref{sec:future_ecosys_share}).
The ultimate vision is to develop comprehensive platforms like Intern-Discovery\cite{intern-discovery} and ScienceOne~\cite{scienceone}, which aim to support the entire research workflow through human-machine collaboration and the integration of ``dry'' computational analysis with ``wet'' lab experimentation, turning Sci-LLMs from ``knowledge processors'' to genuine ``reasoning engines'' for scientific discovery. 




\subsubsection{The Data Bottleneck Behind the Rise of Scientific Agents}
\label{sec:future_ecosys_bottleneck}

A primary bottleneck is the severe imbalance in data modalities available for training. The corpora for today's Sci-LLMs are overwhelmingly dominated by textual data, such as scientific papers and textbooks~\cite{taylor2022galactica, beltagy2019scibert}. While valuable, this creates a critical gap: there is a severe scarcity of high-quality, AI-ready experimental and observational data. This imbalance forces models to learn a description of science rather than the underlying principles from primary evidence. Consequently, their reasoning is often shallow, excelling at textual pattern matching but struggling with novel problems that require a deep, causal understanding of experimental phenomena. Efforts to bridge this gap, such as Biomni~\cite{huang2025biomni} which integrates heterogeneous biological data from genomics to proteomics, underscore both the necessity and the immense difficulty of creating such multimodal datasets at scale.

Compounding this issue is the disconnected nature of the scientific knowledge hierarchy within current datasets. Scientific knowledge is not a flat collection of facts but a structured hierarchy, and existing data fails to capture the rich connections between its layers (Sec.~\ref{sec:background_structure}). For instance, raw experimental data is often decoupled from its rich context, such as the specific instrumental settings and protocols used to generate it, making it nearly impossible for an agent to critically evaluate data quality. Furthermore, while scientific formulas are abundant in texts, the logical derivation processes and underlying assumptions are rarely encoded, limiting an agent's ability to perform rigorous, step-by-step symbolic reasoning. Most critically, the creative aspects of science, including failed experiments, serendipitous discoveries, and novel hypotheses, are almost entirely absent from training data, starving agents of the examples needed to learn genuine innovative thinking.

\subsubsection{Building an Operating System-level Interaction Protocol}
\label{sec:future_ecosys_os}

To transcend these limitations, the solution lies not merely in better datasets but in a fundamentally new architecture for how agents interact with the scientific world. This necessitates a shift from monolithic, self-contained models to dynamic, agent-based systems capable of wielding external tools for experimentation, simulation, and analysis~\cite{mialon2023augmented}. Such complex interaction demands an \emph{operating system-level interaction protocol}, which would serve as the standardized interface between the agent's core reasoning engine and the vast ecosystem of specialized scientific resources, including databases, computational simulators, data analysis packages, and even robotic wet-lab platforms.

This ``operating system'' would empower the scientific agent to autonomously manage a full research cycle. Upon receiving a high-level objective, the agent would first decompose the problem into a sequence of actionable steps. For each step, it would select and invoke the appropriate tool through the standardized protocol—be it querying the Materials Project database for candidate compounds, running a simulation in LAMMPS~\cite{thompson2022lammps} to test physical properties, or analyzing spectral data with a dedicated library. The protocol must also enable the agent to parse the diverse outputs from these tools, including numerical results, error codes, structured data files, while integrating this new information back into its reasoning context to inform its next action. By establishing this robust interaction framework, we can begin to address the core data bottlenecks directly. An agent equipped with such a protocol is no longer solely dependent on static, pre-existing datasets. Instead, it can actively generate and consume AI-ready data on the fly, bridging the chasm between textual knowledge and empirical evidence. This creates a closed-loop system where hypotheses are not just formulated based on past literature but are immediately tested through simulation or data retrieval, and the results iteratively refine the agent's understanding.

\subsubsection{Design Principles for Next-Generation Scientific Data Architecture}

\label{sec:future_ecosys_design}

Realizing the vision of autonomous scientific agents necessitates a fundamental rethinking of how scientific data is created, managed, and shared. Merely accumulating more data is insufficient; the next generation of scientific data infrastructure must be architected from the ground up to support agent-driven discovery. This requires a paradigm shift guided by a new set of design principles that prioritize the needs of intelligent systems, transforming data from a passive archive into an active, operational resource. These principles aim to resolve the systemic bottlenecks of traceability, latency, and AI-readiness that currently hinder progress.

The foremost principle is to ensure that all scientific data is actionable and AI-ready by design. This moves beyond the FAIR principles of Findability, Accessibility, Interoperability, and Reusability by demanding that data be immediately consumable by machine learning models with minimal preprocessing~\cite{wilkinson2016fair}. In practice, this means establishing and enforcing community-wide standards for rich, structured metadata that captures the full experimental context, from sample provenance and instrument calibration to software versions and processing parameters. Data should be published not as static, isolated files but as integrated packages that link raw outputs to their corresponding protocols and analyses, enabling an agent to understand not just what the data is, but how it was generated and why it is significant.

A second critical principle is the development of infrastructure for continuous integration and low-latency updates. The current lag between a scientific discovery and its incorporation into training corpora renders models perpetually out-of-date, a fatal flaw in fast-moving fields. Next-generation data architectures must implement automated pipelines that continuously ingest, validate, and structure new data from publications, preprints, and experimental platforms. Adopting open-access policies and version-controlled repositories with real-time API access will be crucial. This ensures that scientific agents can learn from the most current knowledge and experimental findings, reducing the risk of hallucination and enabling them to reason at the cutting edge of research.

Finally, the new architecture must be built upon a foundation of unambiguous traceability and comprehensive knowledge integration. To build trustworthy AI systems, every piece of data must be accompanied by an immutable record of its origin and transformation history, a ``chain of custody'' that allows for complete reproducibility and auditing~\cite{liang2017provchain}. This requires more than just metadata; it calls for the integration of data across different modalities and levels of the scientific knowledge hierarchy. The ideal data ecosystem would seamlessly link a theoretical concept in a textbook to the specific formulas that formalize it, which in turn connect to the experimental datasets that validate it, and the computational code used to analyze it. By architecting this deeply interconnected web of knowledge, we provide scientific agents with the rich, multi-faceted context they need to perform complex, verifiable, and truly insightful reasoning.

\subsubsection{Sustainable Data Sharing Mechanism}
\label{sec:future_ecosys_share}

Traditional models of data exchange, such as centralized repositories,  or closed-access publications, are proving insufficient for the scale, diversity, and adaptability required to support the development of cutting-edge scientific LLMs. As LLMs increasingly depend on vast, heterogeneous, and continuously evolving datasets, data sharing is being reconceptualized as a dynamic ecosystem rather than a static resource.

Emerging paradigms of sustainable data sharing center on principles of openness, fairness, and long-term viability. Decentralized architectures, often enabled by blockchain, create transparent systems for tracing provenance, attributing value, and rewarding contributions through automated contracts, which can foster trust and incentivize participation. Data ecosystems are shifting from static collections to automated curation pipelines that continuously integrate peer-reviewed publications, experimental outputs, and domain-specific repositories.
This ensures that scientific LLMs are not only comprehensive but also current and reliable. The establishment of community-governed data commons is also important, in which stakeholders across academia, industry, and public institutions collaborate to set standards for licensing and ethical use.
At the same time, recognizing data-sharing contributions within academic evaluation systems, similar to citation credit, could provide strong incentives for participation. The main challenge is to create fair and transparent rules for governance and benefit-sharing that balance the interests of institutions, companies, and individual researchers while ensuring legal, ethical, and reproducible practices. Ultimately, sustainable data sharing mechanisms represent not just a technical necessity but also a cultural and institutional shift, laying the foundation for scientific LLMs that can accelerate discovery while upholding the values of equity, transparency, and reproducibility.

\subsubsection{Data Safety and Privacy}

The transition to data-driven science is gated by a critical threshold of trust: to confidently leverage high-value datasets, researchers must be assured of their safety, ethical standing, and legal compliance. Creating this trust requires a comprehensive governance framework built on two core pillars: robust privacy protection and adherence to national data controls.

The first pillar is rigorous privacy protection, particularly for sensitive information in fields like medicine and social sciences. Data can be stratified by risk, from low-risk aggregated statistics to high-risk genomic or personal health records~\cite{ohm2009broken}. A primary challenge with high-risk data is preventing re-identification, where even anonymized datasets can be cross-referenced with public information to uncover individual identities~\cite{sweeney2002k,atreya2013reducing,narayanan2008robust}. This risk necessitates advanced de-identification techniques and strict access protocols to protect research participants.

The second pillar addresses data sovereignty and national controls. Scientific data is increasingly viewed as a strategic national asset, leading nations to implement regulations that govern its cross-border flow and use. Prominent examples include the European Union's General Data Protection Regulation (GDPR)~\cite{regulation2018general}, which imposes strict conditions on transferring personal data of EU citizens internationally~\cite{chassang2017impact}, and U.S. Export Administration Regulations (EAR)~\cite{jacobsen2019us}, which control the export of sensitive dual-use technologies. These legal frameworks require that international scientific collaborations could build compliance into their data management plans from the outset to avoid project-threatening legal and ethical conflicts.

\section{Challenges and Outlook}
\label{sec:outlook}


\subsection{Challenges}
\label{sec:challenge}
\subsubsection{Scientific Data Selection for Efficient Pretraining}

The sheer volume of scientific literature and data necessitates a strategic approach to data selection for pretraining Sci-LLMs. Naively ingesting all available information is not only computationally expensive but can also be detrimental to model performance due to the varying quality of data \cite{albalak2024survey}. The challenge, therefore, is to curate a high-quality, diverse, and representative dataset that enables the model to learn the fundamental principles of a scientific domain. A significant hurdle is the inherent noise and bias present in scientific datasets. Training data can contain everything from experimental artifacts and outdated information to systemic biases present in the research literature. Filtering out such low-quality or irrelevant data is crucial for improving training efficiency and the downstream performance of the model. Furthermore, ensuring broad coverage across different sub-domains, languages, and contexts is essential to prevent the model from becoming overly specialized and to foster interdisciplinary insights. Recent approaches to data selection are moving beyond simple heuristics. Model-based filtering techniques, which use a trained model to identify high-quality and diverse data samples, have shown promise in improving pretraining for multilingual datasets \cite{messmer2025enhancing}. Some methods even employ online batch selection, dynamically choosing the most informative data during the training process itself to adapt to the model's evolving understanding \cite{wang2024greats}, and thus create an efficient pretraining process by focusing on data that maximizes learning and generalization \cite{wettig2024qurating,xia2024less}.

\subsubsection{Optimizing Data Processing Pipelines}

Once a dataset has been selected, it must be transformed into a format that a large language model can understand. This involves developing robust and scalable data processing pipelines tailored to the unique characteristics of scientific information. Traditional data pipelines often struggle with the heterogeneity of scientific data, which can range from unstructured text and images to highly structured formats like tables and code. The tokenization process, which breaks down text into manageable units for the model, presents a significant challenge in scientific domains. General-purpose tokenizers, such as BPE (Byte Pair Encoding), frequently fail to capture the semantic meaning of specialized scientific terms, chemical formulas, or biological sequences, leading to fragmented representations. For instance, a complex molecule name might be broken into generic tokens that lose its specific chemical meaning. Consequently, specialized vocabularies and tokenization strategies are required to maintain domain fidelity. Additionally, data cleaning and normalization are crucial steps, particularly for unstructured formats like PDFs, which often contain formatting errors, figures, and tables that must be accurately extracted and converted to a uniform input format for efficient processing by the model~\cite{taylor2022galactica,zhang2024sciglm}.

\subsubsection{Representing Non-Sequential and Non-Textual Data}

Large language models are fundamentally designed to process sequential data, typically text. However, a significant portion of scientific knowledge is expressed in non-sequential and non-textual formats, presenting a profound challenge for Sci-LLMs. In chemistry, models must interpret 3D molecular structures, which are inherently graphical and non-sequential, alongside text-based representations like SMILES strings. Similarly, in biology, protein structures, gene regulatory networks, and genomic data are challenging to represent within a standard linear transformer architecture. Addressing this requires innovative approaches that bridge the gap between sequential language processing and complex data structures. This often involves multimodal or hybrid architectures. For example, some approaches utilize graph to encode structural information like molecular graphs and then project these embeddings into the Transformer's input space. Other methods rely on specialized encoding schemas, such as representing complex mathematical equations or tables as structured text sequences, while still preserving their logical and spatial relationships. The challenge lies in ensuring that these representations maintain semantic fidelity and allow the model to reason across different modalities, moving beyond simple text understanding to truly grasp the complex relationships embedded in scientific data.

\subsubsection{LLM Knowledge Update and Version Control}

Scientific research evolves rapidly, with constant influxes of new discoveries, datasets, and revised theories across disciplines. Yet, most LLMs are trained on static snapshots of the literature, rendering them quickly outdated, especially in fast-moving domains like biomedicine, healthcare, and atmospheric science, where recent findings can directly influence critical decisions. Retrieval-augmented approaches offer partial relief by accessing external sources at inference time, but often fall short in relevance filtering, source attribution, and resolving conflicting information. To develop truly current scientific LLMs, continuous and automated updating pipelines are essential, capable of regularly ingesting peer-reviewed publications, preprints, and curated datasets with built-in version control and traceability. 
Although tools like ChatGPT and DeepSeek integrate web search, they lack guarantees of relevance or reliability. A promising direction is to create collaborative platforms for dataset generation and distribution, leveraging adaptive strategies to ensure sustained LLM performance over time.

\subsection{Future Work}
\label{sec:futurework}


\subsubsection{Integrated Scientific Data Ecosystems}
The path forward requires fundamental reconceptualization of how we approach scientific AI, moving beyond incremental improvements to existing paradigms. Central to this transformation is the development of integrated scientific data ecosystems that transcend traditional repository models. These ecosystems must seamlessly connect experimental apparatus, computational simulations, theoretical frameworks, and published knowledge into living, evolving networks. Rather than static datasets, we envision active data streams where new experimental results automatically propagate through the system, updating model understanding while maintaining rigorous provenance tracking. This requires not only technical infrastructure but also new incentive structures within the scientific community that reward data curation and sharing as first-class research contributions.

\subsubsection{Automated Scientific Data Standardization Pipeline}

In the era of data-centric scientific AI, future work must prioritize the development of automated data standardization pipelines.  
These pipelines will serve as the foundational infrastructure for training robust and reproducible Sci-LLMs, with emphasis shifting from model architecture to data curation. More research work should focus on developing systems that can automatically clean, validate, and enrich raw scientific data in heterogeneous forms and modalities, ensuring high-fidelity inputs for AI models. The development of robust data versioning and reproducible preparation workflows will also be essential to make Sci-LLM development not just scalable but also transparent and reproducible. The ultimate goal is to move from manual, ad hoc data curation to a scalable, automated system that provides the scientific community with readily accessible, high-quality, and standardized data.

\subsubsection{Comprehensive Evaluation System}

Future directions for comprehensive evaluation should address challenges at both the model and data levels. From the perspective of Sci-LLMs, there is a growing need for standardized, domain-specific benchmarks that go beyond surface-level metrics to assess reasoning depth, factual accuracy, and scientific creativity across disciplines. Evaluations should incorporate multimodal and multistep scientific tasks to better reflect real-world research scenarios. 
On the data side, defining and measuring dataset quality remains a fundamental challenge, as current approaches often fail to capture how data supports model capabilities. Key criteria, such as AI-readiness, completeness, scientific relevance, timeliness, usability, and accessibility, must be integrated into data evaluation frameworks. 
A key direction for future research is to develop a systematic framework for data assessment, enabling more informed dataset selection and ultimately advancing model reliability and performance. 
Integrating these two perspectives will enable more robust, nuanced, and trustworthy evaluation frameworks that drive the development of truly capable scientific AI systems.

\subsubsection{Advanced Scientific Reasoning}
The evolution from current language models to genuine scientific reasoning systems demands architectural innovations that embed physical laws, causal structures, and domain-specific constraints directly into model design. Future architectures must move beyond pattern matching to incorporate symbolic reasoning capabilities, enabling manipulation of mathematical equations and chemical structures with the same fluency as natural language. These systems should exhibit compositional generalization—applying learned principles to novel combinations never seen during training—and maintain explicit representations of uncertainty that propagate through reasoning chains. The integration of neural and symbolic approaches, long pursued but never fully realized, becomes essential for scientific domains where interpretability and correctness are paramount.

\subsubsection{Autonomous Scientific Agents}
A paradigm shift from passive models to active scientific agents represents perhaps the most transformative direction for future research. These agents must possess capabilities beyond current systems: proposing testable hypotheses, designing experiments to resolve uncertainties, and iterating based on empirical results. This requires developing safe interaction protocols with laboratory equipment and simulation environments, creating standardized interfaces for scientific tools and databases, and establishing frameworks for multi-agent collaboration where specialized models contribute complementary expertise. The vision extends to AI systems that not only assist human scientists but also autonomously explore hypothesis spaces too vast for human investigation.

\subsubsection{From Sci-LLMs to Scientific Discovery}
The ultimate objective of Sci-LLMs extends beyond the automation of routine tasks to the acceleration of pivotal scientific breakthroughs. Sci-LLMs present unique potentials to identify subtle, non-obvious correlations and patterns within vast, multimodal datasets that would be impossible for human researchers to process in a short time. We are moving from a phase where Sci-LLMs are primarily used for literature review and synthesis to an advanced stage where these models can serve as powerful instruments for accelerated hypothesis generation, potentially contributing to Nobel Prize-worthy discoveries.
While human creativity and ethical oversight remain important, Sci-LLMs will act as collaborators to help significantly reduce the discovery cycle, allowing researchers to pursue more ambitious research. This integration has the potential to redefine the very nature of scientific method, pushing the boundaries of human knowledge in unprecedented ways.

\subsubsection{Ethical Governance for Responsible Scientific AI Innovation}
The responsible development of increasingly capable scientific AI systems necessitates robust ethical frameworks and governance structures~\cite{lab2025safework}. As these systems begin to influence research directions and resource allocation, ensuring equitable access becomes critical to prevent further concentration of scientific capabilities. Questions of attribution, accountability, and validation for AI-generated discoveries require careful consideration and community consensus. The environmental impact of training large-scale models shall be balanced against their potential contributions to sustainability science, demanding innovations in efficient training and model architectures.

\section{Conclusion}\label{sec:conclusion}
This survey systematically reviews the emerging field of scientific large language models from the perspectives of data, model architectures, and agent-based systems. By introducing a unified taxonomy of scientific data and analyzing more than 270 pre-training and post-training datasets as well as over 190 evaluation datasets, we highlight the distinctive multimodal, cross-scale, and domain-specific challenges that differentiate scientific AI from general-purpose LLMs. We summarize the evolution from transfer learning and large-scale foundation models to instruction-following and tool-augmented scientific agents, and examine current evaluation practices spanning static benchmarks, process-oriented assessments, and autonomous scientific discovery frameworks. We further discuss persistent issues in data quality, representation gaps, and knowledge updating, and outline future directions including operating-system–level data ecosystems and hybrid neural–symbolic architectures. Together, these insights provide a consolidated reference and a forward-looking roadmap for building trustworthy, continually evolving Sci-LLMs capable of advancing data-driven scientific discovery.

\newpage

\begin{table*}
\centering
\caption{Data source description.}
\label{tab:data_source}
\begin{adjustbox}{max width=\textwidth}
\begin{tabular}{ >{\arraybackslash}m{4cm} >{\arraybackslash}m{12cm}}
\toprule
\textbf{Source} & \textbf{Description}\\

\midrule

\textbf{Web and Internet content} &  High-quality web crawl datasets containing billions of pages from diverse internet sources, including news articles, blogs, and general web content. These datasets undergo extensive cleaning and deduplication processes to ensure text quality for language model training.\\

\midrule

\textbf{Books and literary works} & Digitized collections of books spanning various genres, languages, and time periods. Sources include public domain texts, open-access libraries, and e-book platforms, providing rich narrative content and diverse writing styles. \\

\midrule

\textbf{Encyclopedias and knowledge bases} & Structured knowledge repositories like Wikipedia and other encyclopedic sources across multiple languages. These provide factual, well-organized information on diverse topics with consistent formatting and citation standards. \\

\midrule

\textbf{Academic and research resources} & Peer-reviewed papers, preprints, theses, and scholarly publications from repositories like arXiv and academic databases. These sources offer technical, specialized content with rigorous methodology and domain expertise. \\

\midrule

\textbf{Social media and forums} & User-generated content from platforms like Reddit and Stack Exchange, capturing conversational language, community discussions, and Q\&A formats that reflect natural human communication patterns. \\

\midrule

\textbf{Integration of existing datasets} & Curated collections that combine and refine multiple existing open-source datasets, leveraging previous data curation efforts to create comprehensive training corpora. \\
\bottomrule
\end{tabular}
\end{adjustbox}
\end{table*}

\begin{table*}
\centering
\caption{Data type description.}
\label{tab:data_type}
\begin{adjustbox}{max width=\textwidth}
\begin{tabular}{ >{\arraybackslash}m{4cm} >{\arraybackslash}m{12cm}}
\toprule
\textbf{Type} & \textbf{Description}\\

\midrule

\textbf{Raw text} & A broad umbrella for any string-serializable content, \eg, natural language plus tables, sequences, code, logs, \etc Used for language modeling or domain pretraining without explicit prompts/answers or paired media. \\

\midrule

\textbf{Text QA} & Text-only question-answer pairs, optionally with supporting passages, supervising reading comprehension or factual reasoning. \\

\midrule

\textbf{Text QA with CoT} & Text QA augmented with explicit multi-step explanations or derivations alongside the final answer. \\

\midrule

\textbf{VQA} & Visual question-answering pairs, where each image is with a question and the corresponding answer. \\

\midrule

\textbf{VQA (multi-image)} & A question grounded on two or more related images, requiring cross-image comparison, temporal alignment, or aggregation. \\

\midrule

\textbf{VQA with CoT} & VQA data augmented with step-by-step rationales or intermediate reasoning traces in addition to the final answer. \\

\midrule

\textbf{Image-text} & Image-text pairs, where the text contains description (\eg, captions, reports) for alignment, captioning, retrieval, or representation learning. \\

\midrule

\textbf{Video-text} & Video-text pairs, where the text contains description (\eg, subtitles, transcripts, narrations) for alignment, captioning, retrieval, or representation learning. \\

\midrule

\textbf{Classification, regression, generation, \etc} & For numeric/matrix/graph records lacking natural-language pairing, annotate by supervised objective. \\

\bottomrule
\end{tabular}
\end{adjustbox}
\end{table*}

\newpage

\onecolumn
\fontsize{4pt}{3pt}\selectfont  %
\begin{landscape}  
\begin{longtable}{m{1.3cm}|m{3cm}|m{2cm}|m{2.5cm}|m{1.5cm}|m{2cm}|l|l|m{2cm}|m{1cm}|m{0.7cm}|m{2cm}|m{1.2cm}|m{1cm}}
    \caption{Summary of pre-training and post-training datasets for scientific LLMs/MLLMs. \textcolor{magenta}{[link]} directs to dataset websites.}
    \label{tab:training_datasets}\\
    \toprule[1pt]
    \textbf{Scientific Domain} & \textbf{Dataset}  &\textbf{Subdomain} & \textbf{Modality} & \textbf{Purpose} & \textbf{Type} & \textbf{Release} & \textbf{Language}& \textbf{Source}& \makecell{\bfseries Annotation\\ \bfseries Pipeline} & \makecell{\bfseries Human \\ \bfseries Annotators} & \textbf{Human Tasks} & \makecell{\bfseries Auto-annotation\\ \bfseries Tools} & \textbf{Size} \\
    \midrule
    \endfirsthead  

    
    \multicolumn{7}{r}{\tablename\ \thetable{} -- continued from previous page}\\
    \toprule[1pt]
    \textbf{Scientific Domain} & \textbf{Dataset}  &\textbf{Domain} & \textbf{Modality} & \textbf{Purpose} & \textbf{Type} & \textbf{Release} & \textbf{Language}& \textbf{Source}& \makecell{\bfseries Annotation \\ \bfseries Pipeline} &\makecell{\bfseries Human \\ \bfseries Annotators} & \textbf{Human Tasks} & \makecell{\bfseries Auto-annotation\\ \bfseries Tools} & \textbf{Size} \\
    \midrule
    
    \endhead
    
    \midrule
    \multicolumn{7}{r}{Continued on next page}\\
    \endfoot

    \bottomrule[1pt]
    \endlastfoot

    \multirow{6}{*}{Life Sciences}

    &MIRAGE~\cite{dongre2025mirage}~\href{https://arxiv.org/abs/2506.20100}{[link]} & Agriculture &  Biological entity photos & SFT & VQA (multi-image) & 2025.06 & EN & Scientific databases & Semi-automated & N/A & Data generation & GPT-4.1 & 37,512 \\
    &CROP~\cite{zhang2024empowering}~\href{https://dl.acm.org/doi/10.5555/3737916.3739585}{[link]} & Agriculture & Academic papers & SFT & Text QA & 2024.09 & EN, ZH & Academic and research resources & Semi-automated & N/A & Data generation & GPT-4 & 211,909 \\

    & ToT‑Biology~\href{https://huggingface.co/datasets/moremilk/ToT-Biology}{[link]} & General Biology &  Biomedical QA & SFT, CoT & Text QA with CoT & 2025.01 & EN & Academic and research resources & N/A & N/A & N/A & N/A & 23,000 \\

    & BioASQ10b-factoid~\cite{krithara2023bioasq}~\href{https://huggingface.co/datasets/legacy107/bioasq10b-factoid}{[link]} & General Biology & Clinical dialogue & SFT & Text QA & 2023.07 & EN & Academic and research resources & Manual & N/A & Data generation and review & N/A & 1.25K \\

    & ReasonMed~\cite{sun2025reasonmed}~\href{https://huggingface.co/datasets/lingshu-medical-mllm/ReasonMed}{[link]}&  Healthcare and Medical Sciences  &  Clinical dialogue & SFT, CoT &  Text QA with CoT & 2025.06 &  EN  &  Comprehensive multi-source integration  &  Automated  &  N/A  &  N/A  &  Qwen-2.5-72B, DeepSeek-R1-Distill-Llama-70B, HuatuoGPT-o1-70B  & 194,925 \\

    & Open-PMC-18M~\cite{baghbanzadeh2025open}~\href{https://huggingface.co/datasets/vector-institute/open-pmc-18m}{[link]} & Healthcare and Medical Sciences & CT, CFP & Pre-training & Image-text & 2025.06 & EN & Academic and research resources & Automated & N/A & N/A & N/A & 25,000,000 \\
\end{longtable}
\end{landscape}  
\twocolumn

\onecolumn
\fontsize{4pt}{3pt}\selectfont  %
\begin{landscape}  
\begin{longtable}{m{1cm}|m{2.2cm}|m{2.2cm}|m{2cm}|m{0.8cm}|m{1.2cm}|l|m{0.6cm}|m{2cm}|m{1cm}|m{1cm}|m{1.7cm}|m{1.2cm}|m{0.8cm}|m{1cm}|m{1.5cm}}
    \caption{Summary of evaluation datasets for scientific LLMs/MLLMs. \textcolor{magenta}{[link]} directs to dataset websites.}
    \label{tab:eval_datasets}\\
    \toprule[1pt]
    \textbf{Scientific Domain} & \textbf{Dataset}  &\textbf{Subdomain} & \textbf{Modality} & \textbf{Level} & \textbf{Type} & \textbf{Release} & \textbf{Language}& \textbf{Source}&  \makecell{\bfseries Annotation \\ \bfseries Pipeline} & \makecell{\bfseries Human \\ \bfseries Annotators} & \textbf{Human Tasks} &  \makecell{\bfseries Auto-annotation\\ \bfseries Tools} & \textbf{Size}  &  \makecell{\bfseries Evaluation \\ \bfseries Type} & \textbf{Metrics}\\
    \midrule
    \endfirsthead
    
    \multicolumn{16}{c}{\tablename\ \thetable{} continued from previous page}\\
    \toprule[1pt]
   \textbf{Scientific Domain} & \textbf{Dataset}  &\textbf{Domain} & \textbf{Modality} & \textbf{Level} & \textbf{Type} & \textbf{Release} & \textbf{Language}& \textbf{Source}& \makecell{\bfseries Annotation \\ \bfseries Pipeline} & \makecell{\bfseries Human \\ \bfseries Annotators} & \textbf{Human Tasks} & \makecell{\bfseries Auto-annotation\\ \bfseries Tools} & \textbf{Size} & \makecell{\bfseries Evaluation \\ \bfseries Type} & \textbf{Metrics}\\
    \midrule
    \endhead
    
    \midrule
    \multicolumn{16}{r}{Continued on next page}\\
    \endfoot
    \bottomrule[1pt]
    \endlastfoot
    
    \multirow{16}{*}{Life Science} 
    
    &SeedBench~\cite{ying-etal-2025-seedbench}~\href{https://arxiv.org/pdf/2505.13220}{[link]} & Agriculture & Breeding literature & Expert & Text QA & 2025.05 & EN, ZH & Academic and research resources & Semi-automated & N/A & Data generation and review & GPT-4 & 2,264 & MCQ, Open-ended & Acc, F1, ROUGE\\

    &AgEval~\cite{arbab2024ageval}~\href{https://arxiv.org/abs/2407.19617}{[link]} & Agriculture & Plant stress phenotyping photos and annotations & Expert & VQA & 2025.01 & EN & Scientific databases & N/A& N/A & N/A & N/A & 1,200 & Classification, Regression & F1, NMAE \\

    &AgXQA~\cite{kpodo2024agxqa}~\href{https://dl.acm.org/doi/10.1016/j.compag.2024.109349}{[link]} & Agriculture & Agricultural extension records & Expert & Text QA & 2024.10 & EN & Academic and research resources & Semi-automated & N/A & N/A & N/A & 2,186 & Open-ended & EM, F1 \\

    &
    Fundus-MMBench~\cite{liu2025constructing}~\href{https://huggingface.co/datasets/MeteorElf/Fundus-MMBench}{[link]} & Healthcare and Medical Sciences & CFP & Expert &  VQA & 2025.07 & EN & Integration of existing datasets & Manual & N/A & Data review & N/A & 620  & MCQ & Acc  \\

    & ReXVQA~\cite{palReXVQALargescaleVisual2025}~\href{https://huggingface.co/datasets/rajpurkarlab/ReXVQA}{[link]} & Healthcare and Medical Sciences & X-ray & N/A & VQA & 2025.06 & EN & Integration of existing datasets & Semi-automated & 3 & Data review & GPT-4o, ClinicalBERT, MedEmbed & 40,557  & MCQ & Acc \\

    & HealthBench~\cite{arora2025healthbench}~\href{https://openai.com/index/healthbench/}{[link]} & Healthcare and Medical Sciences & Clinical dialogue, Medical task requests, Medical record summarization, \etc & Expert & Text QA & 2025.05 & EN & Comprehensive multi-source integration & Semi-automated & 262 & Data generation and review & GPT-o1, GPT-4.1 & 5,000 & Open-ended & Customized rubric criterion \\
\end{longtable}
\end{landscape}  
\twocolumn

\newpage
\onecolumn
\fontsize{4pt}{5pt}\selectfont  %

\begin{landscape}  
\begin{longtable}{m{2cm}|m{2.2cm}|m{2cm}|m{1.2cm}|l|m{0.6cm}|m{2cm}|m{1cm}|m{1cm}|m{1.2cm}|m{1.4cm}|m{0.8cm}|m{1cm}|m{1cm}|m{1.5cm}}
    \caption{Summary of general science evaluation datasets for scientific LLMs/MLLMs. \textcolor{magenta}{[link]} directs to dataset websites.}
    \label{tab:general_datasets}\\
    \toprule[1pt]
    \textbf{Dataset}  &\textbf{Scientific Domain} & \textbf{Modality} &  \textbf{Type} & \textbf{Release} & \textbf{Language}& \textbf{Source}&  \makecell{\bfseries Annotation \\ \bfseries Pipeline} & \makecell{\bfseries Human \\ \bfseries Annotators} & \textbf{Human Tasks} &  \makecell{\bfseries Auto-annotation\\ \bfseries Tools}  & \textbf{Size}  &  \textbf{Level} & \makecell{\bfseries Evaluation \\ \bfseries Type} & \textbf{Metrics}\\
    \midrule
    \endfirsthead
    
    \multicolumn{15}{c}{\tablename\ \thetable{} -- continued from previous page}\\
    \toprule[1pt]
    \textbf{Dataset}  &\textbf{Scientific Domain} & \textbf{Modality} &  \textbf{Type} & \textbf{Release} & \textbf{Language}& \textbf{Source}&  \makecell{\bfseries Annotation \\ \bfseries Pipeline} & \makecell{\bfseries Human \\ \bfseries Annotators} & \textbf{Human Tasks} &  \makecell{\bfseries Auto-annotation\\ \bfseries Tools}  & \textbf{Size}  &  \textbf{Level} & \makecell{\bfseries Evaluation \\ \bfseries Type} & \textbf{Metrics}\\
    \midrule
    \endhead
    
    \multicolumn{15}{r}{Continued on next page}\\
    \endfoot
    \bottomrule[1pt]
    \endlastfoot

    MMMU~\cite{Yue2024MMMU}~\href{https://huggingface.co/datasets/MMMU/MMMU}{[link]} & Science (Biology, Chemistry, Geography, Math, Physics), Health \& Medicine (Basic Medical Science, Clinical Medicine, Diagnostics, Pharmacy, Public Health), Tech \& Engineering (Materials, \etc)  & Scientific VQA, MRI, CT, X-ray, \etc  & VQA & 2023.11 &  EN&  Comprehensive multi-source inte-
    gration &  Semi-automatic & 50 & Data review & Claude, GPT-4, GPT-4V & 11,550 & Expert  & MCQ  & Acc \\

    \midrule
    
    MMMU Pro~\cite{yueMMMUProMoreRobust2025}~\href{https://huggingface.co/datasets/MMMU/MMMU_Pro}{[link]}
    & Science (Biology, Chemistry, Geography, Math, Physics), Health \& Medicine (Basic Medical Science, Clinical Medicine, Diagnostics, Pharmacy, Public Health), Tech \& Engineering (Materials, \etc) & Scientific VQA, MRI, CT, X-ray, \etc  & VQA & 2023.11 &  EN&  Comprehensive multi-source integration &  Semi-automatic & N/A & Data review & Claude, GPT-4, GPT-4V & 5,190 & Expert  & MCQ  & Acc \\

    \midrule
     ScienceQA ~\cite{saikh2022scienceqa}~\href{https://github.com/lupantech/ScienceQA}{[link]} &
     Biology, Earth Science, Physics, Chemistry, Geography  & Scientific query, Scientific instruction, Science textbooks and literature  & VQA & 2022.01 & EN & Books and literary works & Manual & 9+ & Data generation and review & ViT, GPT-2 & 21.2k &  Primary school, Secondary school  & MCQ & Acc \\

    \midrule
    
    SciQA \cite{auer2023sciqa} \href{https://huggingface.co/datasets/orkg/SciQA}{[link]} & Material Science, Chemistry, Life sciences & Scientific query  & Text QA & 2023.05 & EN & Academic and research resources & Manual & N/A & Data generation and review & N/A & 2,565 & Expert & Open-ended & Acc \\
\end{longtable}
\end{landscape}  
\twocolumn

\newpage

\onecolumn
\fontsize{4pt}{3pt}\selectfont  %
\begin{landscape}  
\begin{longtable}{m{1.3cm}|m{3cm}|m{3cm}|m{2cm}|m{2cm}|m{2cm}|m{1cm}|m{1cm}}
    \caption{Summary of scientific large language models. \textcolor{magenta}{[link]} directs to model websites.}
    \label{tab:sci_llms}\\
    \toprule[1pt]
    \textbf{Scientific Domain} & \textbf{Model}  & 
    \textbf{Subdomain}  & 
    \textbf{Parameters} & \textbf{Base Model} & \textbf{Modality Encoder} & \textbf{Release} & \textbf{Open-source} \\
    
    \midrule
    \endfirsthead  

    
    \multicolumn{7}{c}{\tablename\ \thetable{} -- continued from previous page}\\
    \toprule[1pt]
    
    \textbf{Scientific Domain} & \textbf{Models} & \textbf{Domain}  & \textbf{Parameters} & \textbf{Base LLM} & \textbf{Modality encoder} & \textbf{Release} & \textbf{Open-source} \\
    
    \midrule
    
    \endhead

    
    \midrule
    \multicolumn{7}{r}{Continued on next page}\\
    \endfoot

    \bottomrule[1pt]
    \endlastfoot

    \multirow{2}{*}{General-purpose}

    & Galactica~\cite{taylor2022galactica}~\href{https://github.com/paperswithcode/galai}{[link]} &  General Science  & 120B & N/A & N/A & 2022.11 & \ding{51} \\

    & DARWIN~\cite{xie2023darwin}~\href{https://github.com/MasterAI-EAM/Darwin}{[link]} &  General Science  & 7B & LLaMA-7B, Vicuna-7B & N/A & 2023.08 & \ding{51}  \\

    & FORGE~\cite{yin2023forge}~\href{https://github.com/at-aaims/forge}{[link]} &  General Science  & 26B &  GPT-NeoX & N/A & 2023.11 & \ding{51}  \\

    & SciGLM~\cite{zhang2024sciglm}~\href{https://github.com/THUDM/SciGLM}{[link]} & General Science   & 6B / 32B & ChatGLM3 & N/A & 2024.01 & \ding{51}  \\

    & SciDFM~\cite{sun2024scidfm}~\href{https://huggingface.co/OpenDFM/SciDFM-MoE-A5.6B-v1.0}{[link]} & General Science  & 18.2B-A5.6B & N/A & N/A & 2024.09 & \ding{51}  \\

    & OmniScience~\cite{prabhakar2025omniscience}~\href{https://github.com/google-deepmind/alphafold3?tab=readme-ov-file}{[link]} & General Science   &  70B & LLaMA-3.1 & N/A & 2025.03 & \ding{55}  \\

    &    Intern-S1~\cite{bai2025interns1scientificmultimodalfoundation}~\href{https://github.com/InternLM/Intern-S1}{[link]} & General Science & 241B-A28B/8B& Qwen3-235B-A22B, Qwen3-8B & InternViT-6B, InternViT-300M & 2025.08 & \ding{51} \\
\end{longtable}
\end{landscape}  
\twocolumn

{
    \small
    \bibliographystyle{IEEEtran}
    \bibliography{ref}
}

\newpage

 





\end{document}